\documentclass[a4paper,11pt]{article}
\usepackage{jmlr2e}
\usepackage{amsmath,amssymb}
\usepackage{graphicx}
\usepackage{epstopdf}
\usepackage{color}
\usepackage{mathrsfs}
\usepackage{bm}
\usepackage{multirow}
\usepackage{booktabs}
\usepackage{url}
\usepackage[ruled]{algorithm}
\usepackage{algpseudocode}
\usepackage{algpascal}
\usepackage{algc}
\usepackage[left]{lineno}

\newcommand\ASTART{\bigskip\noindent\begin{minipage}[b]{0.5\linewidth}}

\newcommand\AENDSKIP{\end{minipage}\bigskip}
\newcommand\AEND{\end{minipage}}

\newcommand{\Part}[3]{ \frac{ \partial^{#3} #1 }{ \partial #2^{#3} } }%partial derivative
\newcommand{\V}[1]{\bm{#1} } %vector command

\newcommand{\Ave}[1]{\left\langle {#1} \right\rangle} %thermal average 
\newcommand{\gsim}{\ $\raisebox{-.7ex}{$\stackrel{\textstyle >}{\sim}$}$\,\ }

\newcommand{\mR}{\mathbb{R}}

\newcommand{\lb}{\left(}
\newcommand{\rb}{\right)}
\newcommand{\lbb}{\left\{}
\newcommand{\rbb}{\right\}}

\newcommand{\Req}[1]{eq.\ (\ref{eq:#1})}

\newcommand{\NReq}[1]{(\ref{eq:#1})}
\newcommand{\Reqs}[2]{eqs.\ (\ref{eq:#1},\ref{eq:#2})}
\newcommand{\Reqsthree}[3]{eqs.\ (\ref{eq:#1},\ref{eq:#2},\ref{eq:#3})}

\newcommand{\Reqss}[2]{eqs.\ (\ref{eq:#1}-\ref{eq:#2})}

\newcommand{\Rfig}[1]{Fig.\ \ref{fig:#1}}

\newcommand{\Lfig}[1]{\label{fig:#1}}
\newcommand{\Leq}[1]{\label{eq:#1}}
\newcommand{\Rsec}[1]{sec.\ \ref{sec:#1}}
\newcommand{\Lsec}[1]{\label{sec:#1}}
\newcommand{\be}{\begin{eqnarray}}
\newcommand{\ee}{\end{eqnarray}}
\newcommand{\ba}{\begin{array}}
\newcommand{\ea}{\end{array}}
\newcommand{\no}{\nonumber}

\newcommand{\subbe}{\begin{subequations}}
\newcommand{\subee}{\end{subequations}}
\newcommand{\bs}{\backslash}

\DeclareMathOperator*{\argmin}{arg\,min}

\newcommand{\TRE}{\epsilon}
\newcommand{\LOOE}{\epsilon_{\rm LOO}}

\newcommand{\Lcode}[1]{\label{code:#1}}
\newcommand{\Rcode}[1]{Alg.\ \ref{code:#1}}
\newcommand{\Rcodes}[2]{Algs.\ \ref{code:#1},\ref{code:#2}}
\newcommand{\NLL}[1]{q_{#1}}
\newcommand{\GRAD}{\nabla}
\newcommand{\HESS}{\partial^2}
\newcommand{\WC}{\ast}

\begin{document}
\title{Accelerating Cross-Validation in \\ Multinomial Logistic Regression with $\ell_1$-Regularization}
\author{\name Tomoyuki Obuchi \email obuchi@c.titech.ac.jp 
       \AND \name Yoshiyuki Kabashima \email kaba@c.titech.ac.jp \\
       \addr Department of Mathematical and Computing Science\\
       Tokyo Institute of Technology\\
       2-12-1, Ookayama, Meguro-ku, Tokyo, Japan
%       \AND
%       \addr Department of Mathematical and Computing Science\\
%       Tokyo Institute of Technology\\
%       Tokyo, 152-8552, Japan
       }
\ShortHeadings{Accelerating CV in Multinomial Logistic Regression}{Obuchi and Kabashima}

\editor{}
%\linenumbers
\maketitle

%%%%%%%%%%%%%%%%%%%%%%%%%%%%%%%%%%%%%%%%%%%%%%%%%%%%%
%%%%%%%%%%%%%%%%%%%%%%%%%%%%%%%%%%%%%%%%%%%%%%%%%%%%%
%%%%%%%%%%%%%%%%%%%%%%%%%%%%%%%%%%%%%%%%%%%%%%%%%%%%%
\begin{abstract}
We develop an approximate formula for evaluating a cross-validation estimator of predictive likelihood for multinomial logistic regression regularized by an $\ell_1$-norm. This allows us to avoid repeated optimizations required for literally conducting cross-validation; hence, the computational time can be significantly reduced. The formula is derived through a perturbative approach employing the largeness of the data size and the model dimensionality. An extension to the elastic net regularization is also addressed. The usefulness of the approximate formula is demonstrated on simulated data and the ISOLET dataset from the UCI machine learning repository. MATLAB and python codes implementing the approximate formula are distributed in~\citep{obuchi:acv_mlr_matlab,obuchi:acv_mlr_python}. 
\end{abstract}


%%%%%%%%%%%%%%%%%%%%%%%%%%%%%%%%%%%%%%%%%%%%%%%%%%%%%
%%%%%%%%%%%%%%%%%%%%%%%%%%%%%%%%%%%%%%%%%%%%%%%%%%%%%
%%%%%%%%%%%%%%%%%%%%%%%%%%%%%%%%%%%%%%%%%%%%%%%%%%%%%
\section{Introduction} \Lsec{Introduction}
Multinomial classification is a ubiquitous task. There are several ways to treat this task, such as the naive Bayesian methods, neural networks, decision trees, and hierarchical classification schemes~\citep{Book:Hastie:09}. Among them, in this paper, we focus on multinomial logistic regression (MLR), which is simple but powerful enough to be used in many present day applications. 

Let us denote each feature vector by $\V{x}_{\mu}\in \mR^{N}$ and its class by $y_{\mu}\in \{1,\cdots,L\}$, where $\mu=1\cdots,M$ denotes the index of given data. The MLR uses a linear structural model with parameters $\{ \V{w}_{a} \in \mR^{N} \}_{a=1}^{L}$ and computes a class-$a$ bias as an overlap:
\be
u_{\mu a}=\V{x}_{\mu}^{\top} \V{w}_{a}.
\Leq{overlap}
\ee
A probability such that the feature vector $\V{x}_{\mu}$ belongs to the class $a$ is computed through a softmax function $\phi$ as:
\be
\phi\lb a \Bigr| \lbb u_{\mu b} \rbb_{b=1}^{L} \rb
=\frac{e^{u_{\mu a}} }{\sum_{b=1}^{L}e^{u_{\mu b}}}.
\ee
These define the MLR. 

The maximum likelihood estimation is usually employed to train the MLR, though the learning result tends to be inefficient when the data size is not sufficiently larger than the model dimensionality or noises in relevant levels are present. A common technique to overcome this difficulty is to introduce a penalty or regularization. In this paper, we use an $\ell_1$-regularization, which induces a sparse classifier as a learning result and is accepted to be effective. Given $M$ data points $D^{M}\equiv \{(\V{x}_\mu,y_{\mu})\}_{\mu=1}^{M}$ , the $\ell_1$-regularized estimator is defined by the following optimization problem:
\be
&&
\{ \hat{\V{w}}_a(\lambda)\}_a 
=
\argmin_{\{ \V{w}_{a} \}_a}
\lbb  \mathcal{H} \lb \lbb \V{w}_{a} \rbb_{a=1}^{L} \Bigr| D^M,\lambda \rb  \rbb,
\\ &&
 \mathcal{H} \lb \lbb \V{w}_{a} \rbb_{a=1}^{L} \Bigr| D^M,\lambda \rb 
\equiv 
\sum_{\mu=1}^{M}
\NLL{\mu}\lb \lbb \V{w}_{a} \rbb_{a=1}^{L} \rb
+\lambda \sum_{a=1}^{L}||\V{w}_{a}||_1,
\Leq{w-full}
\\ &&
\NLL{\mu}\lb \lbb \V{w}_{a} \rbb_{a=1}^{L} \rb
=
-\ln \phi\lb y_{\mu} \Bigr| \lbb u_{\mu a}=\V{x}^{\top}_{\mu}\V{w}_a \rbb_{a=1}^{L} \rb,
\Leq{NLL}
\ee
where we denote the negative log-likelihood as $\NLL{\mu}$ and define a regularized cost function or Hamiltonian $\mathcal{H}$.

The introduction of regularization causes another problem of model selection or hyper-parameter estimation with respect to $\lambda$. A versatile framework providing a reasonable estimate is cross-validation (CV), but it has a disadvantage in terms of the computational cost. The literal CV requires repeated optimizations which can be a serious computational burden when the data size and the model dimensionality are large. The purpose of this paper is to resolve this problem by inventing an efficient approximation of CV. 

Our technique is based on a perturbative expansion employing the largeness of the data size and the model dimensionality. Similar techniques were also developed for the Bayesian learning of simple perceptron and committee machine~\citep{opper1996mean,opper1997mean}, for Gaussian process and support vector machine~\citep{opper2000gaussian1,opper2000gaussian2,vapnik2000bounds}, for linear regression with the $\ell_1$-regularization~\citep{obuchi2016cross,rad2018scalable,wang2018approximate} and with the two-dimensional total variation~\citep{obuchi2017accelerating}. Actually, this perturbative approach is fairly general and can be applied to a wide class of generalized linear models with simple convex regularizations. For example in the present MLR case, it is easy to extend our result to the case where both the $\ell_1$- and $\ell_2$-regularizations exist~\citep[elastic net,][]{zou2005regularization}, which is used in a common implementation~\citep{friedman2010regularization}. The derivation of our approximate formula below is, however, conducted on the case of the $\ell_1$-regularization only, for simplicity. The extension to the elastic net case is stated after the derivation.

The rest of the paper is organized as follows. In \Rsec{Formulation}, we state our formulation and how to derive the approximate formula. In \Rsec{Numerical}, we compare our approximation result with that of the literally conducted CV on simulated data and on the ISOLET dataset from UCI machine learning repository~\citep{Lichman:13}. The accuracy and the computational time of our approximate formula are reported in comparison with the literal CV. The limitation of a simplified version of the approximation is also examined. The last section is devoted to the conclusion.

%%%%%%%%%%%%%%%%%%%%%%%%%%%%%%%%%%%%%%%%%%%%%%%%%%%%%
%%%%%%%%%%%%%%%%%%%%%%%%%%%%%%%%%%%%%%%%%%%%%%%%%%%%%
%%%%%%%%%%%%%%%%%%%%%%%%%%%%%%%%%%%%%%%%%%%%%%%%%%%%%
\section{Formulation} \Lsec{Formulation}
In the maximum likelihood estimation framework, it is natural to employ a predictive likelihood as a criterion for model selection~\citep{bjornstad1990predictive,ando2010predictive}. We require a good estimator of the predictive likelihood, and the CV provides a simple realization of it. Particularly in this paper, we consider an estimator based on the leave-one-out (LOO) CV. The LOO solution is described by
\be
&&
\{ \hat{\V{w}}_a^{\bs \mu}(\lambda)\}_a =
\argmin_{\{ \V{w}_{a} \}_a}
\lbb
\mathcal{H}^{\bs \mu}\lb \lbb \V{w}_{a} \rbb_{a=1}^{L} \Bigr| D^M,\lambda \rb
\rbb,
\Leq{w-LOO}
\\ &&
\mathcal{H}^{\bs \mu}\lb \lbb \V{w}_{a} \rbb_{a=1}^{L} \Bigr| D^M,\lambda \rb
\equiv
\mathcal{H} \lb \lbb \V{w}_{a} \rbb_{a=1}^{L} \Bigr| D^M,\lambda \rb
- \NLL{\mu}\lb \lbb \V{w}_{a} \rbb_{a=1}^{L} \rb.
\ee
Denoting the overlap of $\V{x}_\mu$ with the LOO solution as $\hat{u}^{\bs \mu}_{\mu a}=\V{x}_{\mu}^{\top}\hat{\V{w}}_a^{\bs \mu}$, as well as that with the full solution $\hat{u}_{\mu a}=\V{x}_{\mu}^{\top}\hat{\V{w}}_a$, we can define the LOO estimator (LOOE) of the predictive negative log-likelihood as: 
\be
\LOOE(\lambda)
=
\frac{1}{M}\sum_{\mu=1}^{M}\NLL{\mu}\lb \lbb \hat{\V{w}}^{\bs \mu}_{a} \rbb_{a=1}^{L} \rb
=
-\frac{1}{M}\sum_{\mu=1}^{M}\ln \phi\lb y_{\mu} \Bigr|\{ \hat{u}_{\mu a}^{\bs \mu} \}_{a=1}^{L}\rb.
\Leq{LOOE}
\ee
In the following, the predictive negative log-likelihood is simply called prediction error. The minimum of the LOOE determines the optimal value of $\lambda$ though its evaluation requires us to solve \Req{w-LOO} $M$ times, which is computationally demanding. 
%%%%%%%%%%%%%%%%%%%%%%%%%%%%%%%%%%%%%%%%%%%%%%%%%%%%%
%%%%%%%%%%%%%%%%%%%%%%%%%%%%%%%%%%%%%%%%%%%%%%%%%%%%%
\subsubsection{Notations} \Lsec{Notations}
Here, we fix the notations for a better flow of the derivation shown below. By summarizing the class index, we introduce a vector notation of the overlap as $\V{u}_{\mu}=(u_{\mu a})_a\in \mR^{L}$ and an extended vector representation of the weight vectors $\{ \V{w}_a \}_a$ as $\V{W}=( \V{w}_{a} )_a \in \mR^{LN}$. The $m$th component of $\V{W}$ can thus be decomposed into two parts as $m=(m_c,m_f)$ where $m_c\in \{1,\cdots, L\} $ denotes the class index and $m_f\in \{1,\cdots,N\}$ represents the component index of the feature vector. Namely we write $W_{m}=w_{m_c m_f}$. Correspondingly, we leverage a matrix $X^{\mu}\in \mR^{L\times LN}$ to define a repetition representation of the feature vector $\V{x}_{\mu}$: Each component is defined as:
\be
X^{\mu}_{a m}\equiv \delta_{a m_c}x_{\mu m_f}.
\Leq{X^mu}
\ee
This yields simple and convenient relations:
\be
\V{u}_{\mu}=X^{\mu}\V{W},~X^{\mu}=\lb \Part{\V{u}_{\mu}}{\V{W}}{}\rb^{\rm \top}.
\ee
Further, the class-$a$ probability of $\mu$th data at the full solution $\hat{\V{W}}=(\hat{\V{w}}_a)_a$ is denoted by:
\be
p_{a|\mu}=\phi(a|\{\hat{u}_{\mu b} \}_b)=\frac{e^{\hat{u}_{\mu a}}}{\sum_{b=1}^{L}e^{\hat{u}_{\mu b}}}
\ee
These notations express the gradient and the Hessian of $q_{\mu}$ at the full solution as:
\be
&&
\GRAD \NLL{\mu}(\hat{\V{W}})\equiv 
\left. \Part{\NLL{\mu}}{\V{W}}{}\right|_{\V{W}=\hat{\V{W}}}
=
\Part{\V{u}_{\mu}}{\V{W}}{} 
\left.
\Part{}{\V{u}_{\mu}}{}\NLL{\mu}
\right|_{\V{u}_{\mu}=\hat{\V{u}}_{\mu}}
=\lb X^{\mu}\rb^{\top}\V{b}^{\mu},
\Leq{NLLgradient}
\\ &&  
\HESS \NLL{\mu}(\hat{\V{W}})\equiv 
\left.
\frac{\partial^2 \NLL{\mu}}{\partial\V{W} \partial\V{W}' }
\right|_{\V{W}=\V{W}'=\hat{\V{W}}}
\no \\ &&
=\Part{\V{u}_{\mu}}{\V{W}}{} 
\lb
\left.
\frac{\partial^2 \NLL{\mu}}{\partial\V{u}_{\mu} \partial\V{u}'_{\mu} }
\right|_{\V{u}_{\mu}=\V{u}'_{\mu}=\hat{\V{u}}_{\mu}}
\rb
 \lb \Part{\V{u}'_{\mu}}{\V{W}'}{} \rb^{\top}
=
\lb X^{\mu}\rb^{\top} F^{\mu} X^{\mu},
\Leq{NLLHessian}
\ee
where 
\be
&&
\V{b}^{\mu}\equiv (p_{1|\mu}-\delta_{1y_{\mu}},p_{2|\mu}-\delta_{2y_{\mu}},\cdots,p_{L|\mu}-\delta_{Ly_{\mu}})^{\top},
\Leq{b^mu}
\\ &&
F^{\mu}_{ab}\equiv \delta_{ab}p_{a|\mu}-p_{a|\mu}p_{b|\mu}.
\Leq{F^mu}
\ee
In addition, we denote the cost function Hessians at the respective solutions as: 
\be
&&
G \equiv  \HESS \mathcal{H}(\hat{\V{W}}) 
=\sum_{\mu} \lb \HESS \NLL{\mu}(\hat{\V{W}}) \rb,
\Leq{G}
\\ &&
G^{\bs \mu}
\equiv \HESS \mathcal{H}^{\bs \mu}(\hat{\V{W}}^{\bs \mu})
=\sum_{\nu(\neq \mu)} \lb \HESS \NLL{\nu}(\hat{\V{W}}^{\bs \mu}) \rb.
\Leq{G-LOO}
\ee
Finally, we introduce the symbol $A(\V{W}) \equiv \{ m | W_{m} \neq 0 \}$ representing the index set of the active components of $\V{W}$ and $\hat{A} \equiv A(\hat{\V{W}})$. Given $\hat{\V{W}}$, we denote the active components of a vector $\V{Y}\in \mR^{LN}$ by the subscript as $\V{Y}_{\hat{A}}$. A similar notation is used for any matrix and the symbol $\WC$ is assumed to represent all of the components in the corresponding dimension. 

%%%%%%%%%%%%%%%%%%%%%%%%%%%%%%%%%%%%%%%%%%%%%%%%%%%%%
%%%%%%%%%%%%%%%%%%%%%%%%%%%%%%%%%%%%%%%%%%%%%%%%%%%%%
\subsection{Approximate formula} \Lsec{Approximate formula}
For a simple derivation, it is important to consider that the $\V{w}$-dependence of $\phi$ appears only in the overlap $u=\V{x}^{\top}\V{w}$. Hence, it is sufficient to provide the relation between $\hat{u}_{\mu a}$ and $\hat{u}^{\bs \mu}_{\mu a}$ in order to derive the approximate formula. 

A crucial assumption to derive the formula is that the active set is ``common'' between the full and LOO solutions, $\hat{\V{W}}=(\hat{\V{w}}_a)_a$ and $\hat{\V{W} }^{\bs \mu}=(\hat{\V{w}}^{\bs \mu}_a)_a$; namely $\hat{A}=\hat{A}^{\bs \mu}\equiv A(\hat{\V{W}}^{\bs \mu})$. Although this assumption is literally not true, we numerically confirmed that this approximately holds. In other words, the change of the active set is small enough compared to the size of the active set itself when considering the LOO operation when $N$ and $M$ are large. Moreover, in a related problem of an $\ell_1$-regularized linear regression, the so-called LASSO, it has been shown that the contribution of the active set change vanishes in a limit $N, M\to \infty$ keeping $\alpha=M/N=O(1)$~\citep{obuchi2016cross}. It is expected that the same holds in the present problem. Hence, we adopt this assumption in the following definition. Note that this idea of the active set constantness can be found in preceding analyses of support vector machine~\citep{opper2000gaussian2,vapnik2000bounds}. 

Once the active set $\hat{A}$ is assumed to be known and unchanged by the LOO operation, it is easy to determine the active components of the full and LOO solutions $\hat{\V{W}}$ and $\hat{\V{W}}^{\bs \mu}$. The vanishing condition of the gradient of the cost function is the determining equation:
\be
&&
\lb \GRAD \mathcal{H} \rb_{\hat{A}}=0 \Rightarrow \hat{\V{W}}_{\hat{A}}, 
\Leq{g}
\\ &&
\lb \GRAD \mathcal{H}^{\bs \mu} \rb_{\hat{A}}
=\lb \GRAD \mathcal{H} \rb_{\hat{A}}-\lb \GRAD q_{\mu} \rb_{\hat{A}}=0 
\Rightarrow \hat{\V{W}}^{\bs \mu}_{\hat{A}}.
\Leq{g-LOO}
\ee
The difference between the gradients is only $\GRAD \NLL{\mu}$, and hence the difference between $\hat{\V{W}}$ and $\hat{\V{W}}^{\bs \mu}$ is expected to be small. Denoting the difference as $\V{d}^\mu=\hat{\V{W}}-\hat{\V{W}}^{\bs \mu}$ and expanding \Req{g-LOO} with respect to $\V{d}^{\mu}$ up to the first order, we obtain an equation determining $\V{d}^{\mu}$:
\be
\V{d}^{\mu}_{\hat{A}}=-\lb G^{\bs \mu}_{\hat{A}\hat{A}} \rb^{-1}
\lb \GRAD \NLL{\mu}(\hat{\V{W}}) \rb_{\hat{A}}.
\Leq{d^mu}
\ee
Inserting this and \Req{NLLgradient} into the definition $\V{d}^\mu=\hat{\V{W}}-\hat{\V{W}}^{\bs \mu}$ and multiplying $X^{\mu}$ from left, we obtain:
\be
&&
\hat{\V{u}}^{\bs \mu}_{\mu}
\approx 
\hat{\V{u}}_{\mu}+C_{\mu}^{\bs \mu} \V{b}^{\mu},
\Leq{u-LOO-pre}
\\ &&
C_{\mu}^{\bs \mu}\equiv X^{\mu}_{*\hat{A}} \lb G^{\bs \mu}_{\hat{A}\hat{A}} \rb^{-1} \lb X^{\mu}_{*\hat{A}} \rb^{\top}.
\Leq{C_mu-LOO}
\ee
This equation implies that the matrix inversion operation is necessary for each $\mu$, which still requires a significant computational cost. To avoid this, we employ an approximation and the Woodbury matrix inversion formula in conjunction with \Reqsthree{NLLHessian}{G}{G-LOO}. The result is:
\be
&&
\lb G^{\bs \mu} \rb^{-1}\equiv 
\lb \HESS \mathcal{H}^{\bs \mu}(\hat{\V{W}}^{\bs \mu}) \rb^{-1}\approx
\lb \HESS \mathcal{H}^{\bs \mu}(\hat{\V{W}}) \rb^{-1}=\lb G-  \lb X^{\mu}\rb^{\top} F^{\mu} X^{\mu} \rb^{-1}
\no \\ && 
=
G^{-1}-G^{-1}\lb X^{\mu}\rb^{\top} \lb  -F^{\mu}+X^{\mu}G^{-1}\lb X^{\mu}\rb^{\top} \rb^{-1}X^{\mu}G^{-1}.
\ee
Inserting this into \Req{u-LOO-pre} and simplifying several factors, we obtain: 
\be
\hat{\V{u}}^{\bs \mu}_{\mu}\approx \hat{\V{u}}_{\mu}+C_{\mu}\lb I_{L}-F^{\mu} C_{\mu}\rb^{-1}\V{b}^{\mu},
\Leq{u-LOO}
\ee
where
\be
C_{\mu}=X^{\mu}_{\WC \hat{A}}\lb G_{ \hat{A} \hat{A}} \rb^{-1} \lb X^{\mu}_{\WC \hat{A} }\rb^{\top}.
\Leq{C_mu}
\ee
Now, all of the variables on the righthand side of \Req{u-LOO} can be computed from the full solution $\hat{\V{W}}$ only, which enables us to estimate the LOOE by leveraging a one-time optimization using all of the data $D^{M}$, while avoiding repeated optimizations. 

We should mention the computational cost of this approximation: it is mainly scaled as $O(ML^2|\hat{A}| +ML|\hat{A}|^2 +|\hat{A}|^3 )$. The first two terms come from the construction of $G_{\hat{A}\hat{A}}$ and $C_{\mu}$, and the last one is derived from the inverse of $G$. If $|\hat{A}|$ is proportional to the feature dimensionality $N$, this computational cost is of the third order with respect to the system dimensionality $N$ and $M$. This is admittedly not cheap and the computational cost for the $k$-fold literal CV with a moderate value of $k$ becomes smaller than that for our approximation in a large dimensionality limit. We, however, stress that there actually exists a wide range of $N$ and $M$ values in which our approximation outperforms the literal CV in terms of the computational time, as later demonstrated in \Rsec{Numerical}. Moreover, for treating much larger systems, we invent a further simplified approximation based on the above approximate formula. The computational cost of this simplified version is scaled only linearly with respect to the system parameters $N$ and $M$. Its derivation is in \Rsec{Further simplified} and the precision comparison to the original approximation is in \Rsec{Numerical}.

Another sensitive issue is present in computing $\lb G_{ \hat{A} \hat{A}} \rb^{-1}$. Occasionally the cost function Hessian $G$ has zero eigenvalues and is not invertible. We handle this problem in the next subsection.

%%%%%%%%%%%%%%%%%%%%%%%%%%%%%%%%%%%%%%%%%%%%%%%%%%%%%
\subsubsection{Handling zero modes}\Lsec{Handling zero modes}
In the MLR, there is an intrinsic symmetry such that the model is invariant under the addition of any constant vector to the weight vectors of all classes: 
\be
\V{w}_{a}\to \V{w}_{a}+\V{v}~(\forall{a}).
\ee
In this sense, the weight vectors defining the same model are ``degenerated'' and our MLR is singular. For finite $\lambda$, this is not harmful because the regularization term resolves this singularity and selects an optimal one $\{ \hat{\V{w}}_{a} \}_a$ with the smallest value of $||\V{w}_{a}||_1$ among the degenerated vectors. However, this does not mean that the associated Hessian is non-singular. The regularization term does not provide any direct contribution to the Hessian and as a result, the Hessian tends to have some zero modes.  This prevents taking the inverse Hessian $G^{-1}$ in \Req{C_mu}. How can we overcome this?

One possibility is to fix the weights of one certain class at constant values when solving the optimization problem \NReq{w-full}. This is termed ``gauge fixing'' in physics, and one convenient gauge in the present problem will be the zero gauge in which the weights in a chosen class are fixed at zeros. This is actually found in some earlier implementations~\citep{krishnapuram2005sparse,schmidt2010graphical} and is preferable for our approximate formula because it removes the harmful zero modes of the Hessian from the beginning. However, some other implementations which are currently well accepted do not employ such gauge fixing~\citep{friedman2010regularization}, and moreover even with gauge fixing very small eigenvalues sometimes accidentally emerge in the Hessian. Hence, for user convenience, we require another way of avoiding this problem.

Another possibility is to remove the zero modes by hand. By construction, the zero modes are associated to the model invariance. This implies that those zero modes are irrelevant and may be removed. In fact, we are only interested in the perturbations which truly change the model, and the modes which maintain the model unchanged are unnecessary. According to this consideration, we replace $G^{-1}$ in \Req{C_mu} with the zero-mode-removed inverse Hessian $\overline{G}^{-1}$. The computation of $\overline{G}^{-1}$ is straightforward: we perform the eigenvalue decomposition of $G_{\hat{A}\hat{A}}$ and obtain the eigenvalues $\{ d_{i}\}_{i=1}^{|\hat{A}|}$ and eigenvectors $\{ \V{v}_{i}\}_{i=1}^{|\hat{A}|}$, which allows us to represent 
\be
G_{\hat{A}\hat{A}}
=\sum_{i}d_i\V{v}_i\V{v}_i^{\top}
=\sum_{i\in S^{+}}d_i\V{v}_i\V{v}_i^{\top},
\ee
where $S^+$ denotes the index set of the modes with finite eigenvalues. Then, $\overline{G}^{-1}$ is defined as:
\be
\overline{G}^{-1}_{\hat{A}\hat{A}}\equiv \sum_{i\in S^{+}}d_i^{-1}\V{v}_i\V{v}_i^{\top}.
\Leq{Ginv-zmr}
\ee
Finally, we replace $G^{-1}$ by $\overline{G}^{-1}$ in \Req{C_mu}, and obtain: 
\be
C_{\mu}=X^{\mu}_{\WC \hat{A}}\overline{G}^{-1}_{ \hat{A} \hat{A}} \lb X^{\mu}_{\WC \hat{A} }\rb^{\top}.
\Leq{C^mu2}
\ee
By using this instead of \Req{C_mu}, the problem caused by the zero modes can be avoided.

%%%%%%%%%%%%%%%%%%%%%%%%%%%%%%%%%%%%%%%%%%%%%%%%%%%%%
\subsubsection{Extension to the mixed regularization case}
Let us briefly state how we can generalize the present result to the case of the mixed regularizations of the $\ell_1$- and $\ell_2$-terms \citep[elastic net,][]{zou2005regularization}. The problem to be solved can be defined as follows:
\be
\{ \hat{\V{w}}_a(\lambda_1,\lambda_2)\}_a &=&
\argmin_{\{ \V{w}_{a} \}_a}\lbb 
\sum_{\mu=1}^{M}
\NLL{\mu}\lb \lbb \V{w}_{a} \rbb_{a=1}^{L} \rb
+\lambda_1 \sum_{a=1}^{L}||\V{w}_{a}||_1
+\frac{\lambda_2}{2} \sum_{a=1}^{L}||\V{w}_{a}||_2^2
\rbb.
\Leq{w-mixed}
\ee 
where $||\cdot ||_2$ denotes the $\ell_2$ norm. Following the derivation in \Rsec{Approximate formula}, we realize that the derivation is essentially the same, and the difference only appears in the cost function Hessian:
\be
&&
G_{\rm mxd} =\sum_{\mu} \lb \HESS \NLL{\mu}(\hat{\V{W}}) \rb+\lambda _2 I_{NL},~
\Leq{G-mixed}
\ee
where $I_{K}$ is the identity matrix of size $K$. As a result, we can compute the LOO solution by leveraging the same equation as \Req{u-LOO} by replacing the definition of $C_{\mu}$, \Req{C_mu}, with: 
\be 
C_{\mu}=X_{*\hat{A}}^{\mu} \lb \lb G_{\rm mxd} \rb_{\hat{A}\hat{A}} \rb^{-1} \lb X_{*\hat{A}}^{\mu} \rb^{\top}.
\ee
Thanks to the $\ell_2$ term, the zero mode removal is not needed since the eigenvalues are lifted up by $\lambda_2$.

%%%%%%%%%%%%%%%%%%%%%%%%%%%%%%%%%%%%%%%%%%%%%%%%%%%%%
\subsubsection{Binomial case}
The binomial case $L=2$ is particularly interesting in several applications and thus we write down the specific formula for this case. 

In the binomial case, it is fairly common to express the class $y$ as a binary $y=0, 1$ and to use the following logit function:
\be
\phi_{\rm logit}(y_\mu|u_\mu)=\frac{\delta_{y_{\mu}1}+\delta_{y_{\mu}0}e^{-u_{\mu}} }{1+e^{-u_{\mu}}},
\ee
where 
\be
u_{\mu}=\V{x}_\mu^{\top}\V{w}.
\ee
If we identify $y=0$ in this case as $y=1$ in the two-class MLR case, this is nothing but the two-class MLR with a zero gauge $\V{w}_1=\V{0}$. Hence, there is no harmful zero mode in the Hessian and we can straightforwardly apply our approximate formula. The explicit form in this case is: 
\be
\hat{u}^{\bs \mu}_{\mu}\approx \hat{u}_{\mu}+\frac{c^{\mu}}{1-\Part{\NLL{\mu}}{u_{\mu}}{2}c^{\mu}}\Part{\NLL{\mu}}{u_\mu}{},
\Leq{u-LOO-logit}
\ee
where $\NLL{\mu}=-\ln \phi_{\rm logit}(y_\mu|u_\mu)$ and 
\be
&&
\Part{\NLL{\mu}}{u_{\mu}}{}=\delta_{y_{\mu}0}-\frac{e^{-u_{\mu}}}{1+e^{-u_{\mu}}},
\\ &&
\Part{\NLL{\mu}}{u_{\mu}}{2}=\frac{e^{-u_{\mu}}}{\lb 1+e^{-u_{\mu}} \rb^2},
\Leq{diff-logit}
\\ &&
G_{\hat{A}\hat{A}}=\sum_{\mu=1}^{M} \Part{\NLL{\mu}}{u_{\mu}}{2}
\lb \V{x}_{\mu}\V{x}_{\mu}^{\top} \rb_{\hat{A}\hat{A}},
\Leq{G-logit}
\\ &&
c^{\mu}=\lb \V{x}_{\mu}^{\top} \rb_{\hat{A}}\lb G_{\hat{A}\hat{A}} \rb^{-1} \lb  \V{x}_{\mu} \rb_{\hat{A}},
\ee
and $\hat{A}=\{i|\hat{w}_{i}\neq 0 \}$ is the active set of the full solution, as before. 

Note that this approximation can be easily generalized to arbitrary differentiable output functions by replacing the logit function $\phi_{\rm logit}$. Readers are thus encouraged to implement approximate CVs in a variety of different problems.

%%%%%%%%%%%%%%%%%%%%%%%%%%%%%%%%%%%%%%%%%%%%%%%%%%%%%
%%%%%%%%%%%%%%%%%%%%%%%%%%%%%%%%%%%%%%%%%%%%%%%%%%%%%
\subsection{Further simplified approximation}\Lsec{Further simplified}
As mentioned above, the computational cost of our approximation is $O(M L^2 |\hat{A}|+M L |\hat{A}|^2+|\hat{A}|^3 )$ and should be reduced for treating larger systems. For this, we derive a further simplified approximation based on the invented approximate formula above. We call this a self-averaging (SA) approximation according to physics terminology.

The basic idea for simplifying our approximate formula is to assume that correlations between $W_m$ and $W_n$ are sufficiently weak. The meaning of ``correlation'' is not evident here, but as seen in \Rsec{The SA approximation} the Hessian $G$ can be connected to a (rescaled) covariance $\chi$ between $W_{m}$ and $W_n$ in a statistical mechanical formulation introducing a probability distribution of $\V{W}$. Our weak correlation assumption requires that the correlation between different feature components is negligibly small; $\chi_{mn}(\equiv (1/\beta){\rm cov}(W_m,W_n))=\chi_{(m_f,m_c),(n_f,n_c)}=\delta_{m_f n_f}(\chi_{m_f})_{m_c n_c}$, where $m_c,n_c (=1,\cdots,L)$ are the class indices and $m_f,n_f(=1,\cdots,N)$ are the feature component indices defined thus far, and $\beta$ is the rescaling factor. In this way, the Hessian is assumed to be expressed in a rather restricted form:
\be
\lb G^{\bs \mu} \rb^{-1}_{mn}\approx 
\left\{
\begin{array}{cc}
\lb \chi_{m_f} \rb_{m_cn_c}\delta_{m_f n_f},  & (m,n\in \hat{A})       \\
 0, &   ({\rm otherwise})
\end{array}
\right.,
\Leq{SAkey}
\ee
Namely, the SA Hessian is allowed to take finite values if and only if the two indices share the same feature vector component. The dependence on the data index $\mu$ is also assumed to be negligible, implying that strong heterogeneity among feature vectors is assumed to be absent.

To proceed with the computation, we require a closed equation to determine the $L\times L$ matrix $\chi_{i}$ for $i=1,\cdots,N$. Its derivation is rather technical and is deferred to \Rsec{The SA approximation}. The result is: 
\be
\lb \chi_i \rb_{\hat{A}_i \hat{A}_i }
=\lb 
\lambda_2I_{|\hat{A}_i|}+
 \sigma_{x}^{2}
 \sum_{\nu=1}^{M}  
 \lb 
  \lb I_{L}+F^{\nu}C_{\rm SA}  \rb^{-1}F^{\nu}  
 \rb_{\hat{A}_i \hat{A}_i }
\rb^{-1},
\Leq{chi-SA}
\ee
where $\sigma_x^2=\sum_{\mu}\sum_{i}x_{\mu i}^2/(NM)$ and $\hat{A}_i=\{a |\hat{w}_{a i} \neq 0\}$ is the set of active class variables at the feature component $i$; the other components of $\chi_i$ related to inactive variables are zeros. The SA approximation of $C^{\bs \mu}_{\mu}$, $C_{\rm SA}\in \mR^{L\times L}$, is defined by:
\be
C_{\rm SA}=\sigma_{x}^2\sum_{i=1}^{N}\chi_{i}.
\Leq{C_SA}
\ee
Using the solution of \Reqs{chi-SA}{C_SA}, the approximate formula is now simply expressed as:
\be
\hat{\V{u}}^{\bs \mu}_{\mu}\approx \hat{\V{u}}_{\mu}+C_{\rm SA} \V{b}^{\mu}.
\Leq{u-LOO-SA}
\ee
Note that there is no factor like $\lb I_{L}-F^{\mu} C_{\mu}\rb^{-1}$ in contrast to \Req{u-LOO}, because we directly approximate $C^{\bs \mu}_{\mu}$ in \Req{u-LOO-pre}. 

When solving \Reqs{chi-SA}{C_SA}, the inverse at the right-hand side of \Req{chi-SA} becomes occasionally ill-defined again due to the presence of zero modes. In such cases, we should remove the zero modes as \Req{Ginv-zmr}. Putting $R=\lambda_2I_{L}+
 \sigma_{x}^{2}\sum_{\mu=1}^{M}  \lb   \lb I_{L}+F^{\mu}C_{\rm SA}  \rb^{-1}F^{\mu} \rb$ and performing the eigenvalue decomposition, we define its zero-mode-removed inverse $\overline{R}^{-1}$ as:
\be
R_{\hat{A}_i\hat{A}_i}
=\sum_{j}d_{j}\V{v}_j\V{v}_j^{\top}
=\sum_{j\in S^{+}}d_{j}\V{v}_j\V{v}_j^{\top}
\Rightarrow
\overline{R}^{-1}_{\hat{A}_i\hat{A}_i}=\sum_{j\in S^+}d^{-1}_{j}\V{v}_j\V{v}_j^{\top},
\Leq{Rinv-zmr}
\ee
where $S^+$ is the index set of the modes with finite eigenvalues. This requires a $O(L^3)$ computational cost at a maximum. Leveraging this approach, a naive way to solve \Reqs{chi-SA}{C_SA} is a recursive substitution. If this converges in a constant time, irrespectively of the system parameters $N, M$ and $L$, then the computational cost of the SA approximation is scaled as $O(NL^3+ML^3)$. This is linear in the feature dimensionality $N$ and the data size $M$ and hence, its advantage is significant. 

%%%%%%%%%%%%%%%%%%%%%%%%%%%%%%%%%%%%%%%%%%%%%%%%%%%%%
%%%%%%%%%%%%%%%%%%%%%%%%%%%%%%%%%%%%%%%%%%%%%%%%%%%%%
\subsection{Summary of procedures}\Lsec{Summary of}
Here, we summarize the two versions of the approximation derived thus far as algorithmic procedures. We call the first version, based on \Req{u-LOO}, the approximate CV or ACV, and call the second one, using \Req{u-LOO-SA}, the self-averaging approximate CV or SAACV. The procedures of ACV and SAACV are given in \Rcode{ACV} and \Rcode{SAACV}, respectively; they are written for the case of the mixed regularization \NReq{w-mixed}.
%%%%%%%%%%%%%%%%%%%%%%%%%%%%%%%
\alglanguage{pseudocode}
\begin{algorithm}[htbp]
\caption{Approximate CV of the MLR}\Lcode{ACV}
\begin{algorithmic}[1]
\Procedure{ACV}{$\hat{\V{W}}(\lambda_1,\lambda_2),D^M,\lambda_2$}
	\State Compute the active set $\hat{A}$ from $\hat{\V{W}}$
	\State Compute $\{ \hat{\V{u}}_{\mu},X^{\mu},\V{b}^{\mu},F^{\mu} \}_{\mu}$ 
		by \Reqs{overlap}{X^mu}, \NReq{b^mu} and \NReq{F^mu}
	\State $G_{\hat{A}\hat{A}} \gets \sum_{\mu=1}^{M}\lb X^{\mu}\rb^{\top} F^{\mu} X^{\mu} 
	+\lambda_2 I_{|\hat{A}|}$
		\Comment{$O(ML|\hat{A}|^2 +ML^2|\hat{A}|)$}
    \If{$\lambda_2$ is large enough} 		\Comment{$O(|\hat{A}|^3)$}
      \State $\overline{G}^{-1}_{\hat{A}\hat{A}}=\lb G_{\hat{A}\hat{A}} \rb^{-1}$
      \Else
      \State Compute $\overline{G}^{-1}_{\hat{A}\hat{A}}$ by \Req{Ginv-zmr}
      \EndIf	
	\For{$\mu=1,\cdots,M$} 
		\Comment{$O(ML|\hat{A}|^2 +ML^2|\hat{A}|+ML^3)$}
	\State $C_{\mu}\gets 
		X^{\mu}_{*\hat{A}}\overline{G}^{-1}_{\hat{A}\hat{A}} \lb X^{\mu}_{*\hat{A}}\rb^{\top}$
	\State $\hat{\V{u}}^{\bs \mu}_{\mu}\gets 
		\hat{\V{u}}_{\mu}+C_{\mu}\lb I_{L}-F^{\mu} C_{\mu}\rb^{-1}\V{b}^{\mu}$	
	\EndFor
	\State Compute $\LOOE$ from $\{ \V{u}_{\mu}^{\bs \mu} \}_{\mu}$ by \Req{LOOE} 
	\State \Return $\LOOE$
\EndProcedure
\end{algorithmic}
\end{algorithm}
%%%%%%%%%%%%%%%%%%%%%%%%%%%%%%%
%%%%%%%%%%%%%%%%%%%%%%%%%%%%%%%
\begin{algorithm}[htbp]
\caption{Self-averaging approximate CV of the MLR }\Lcode{SAACV}
\begin{algorithmic}[1]
\Procedure{SAACV}{$\hat{\V{W}}(\lambda_1,\lambda_2),D^M,\lambda_2$}
	\State Compute the active sets $\{ \hat{A}_i \}_{i=1}^{N}$ from $\hat{\V{W}}$
	\State Compute $\{ \V{u}_{\mu},X^{\mu},\V{b}_{\mu},F^{\mu} \}_{\mu}$ 
		by \Reqs{overlap}{X^mu}, \NReq{b^mu} and \NReq{F^mu}
	%%%
	\State $t \gets 0$	\Comment{Start initialization}
	\For{$i=1,\cdots,N$}
		\State $\lb \chi_i^{\bs \mu} \rb^{(t)} \gets 0$,
		\State $\lb \chi_i^{\bs \mu} \rb^{(t)}_{\hat{A}_i\hat{A}_i} \gets \sigma_{x}^{-2}$,
	\EndFor 		
	\State $\Delta \gets 100$	\Comment{End initialization}
	%%%
	\While{$\Delta>\theta$}	\Comment{Compute $C_{\rm SA}$ by recursion}
		\State $C_{\rm SA}^{(t+1)}\gets \sigma_{x}^2\sum_{i=1}^{N} \lb \chi_i^{\bs \mu} \rb^{(t)}$ 
		\State $R \gets \sigma_{x}^2\sum_{\mu=1}^{M}\lb I_L+F^{\mu}C_{\rm SA}^{(t+1)} \rb^{-1}F^{\mu}
			+\lambda_2 I_{L}$
\Comment{$O(ML^3)$}
		\State $ \Delta\gets 0 $
		\For{$i=1,\cdots,N$} \Comment{$O(NL^3)$}
		 \If{$\lambda_2$ is large enough
		\State $\overline{R}_{\hat{A}_i\hat{A}_i}^{-1}=\lb R_{\hat{A}_i\hat{A}_i} \rb^{-1}$
		\Else
		\State Compute $\overline{R}_{\hat{A}_i\hat{A}_i}^{-1}$ by \Req{Rinv-zmr} from $R$
		\EndIf	
		}
			\State $\lb \chi_i^{\bs \mu} \rb^{(t+1)}_{\hat{A}_i \hat{A}_i}
			\gets  \overline{R}_{\hat{A}_i\hat{A}_i}^{-1}$
			\State $\Delta\gets \Delta+\Bigl| \Bigl| \lb \chi_i^{\bs \mu} \rb^{(t+1)}_{\hat{A}_i \hat{A}_i}
			-\lb \chi_i^{\bs \mu} \rb^{(t)}_{\hat{A}_i \hat{A}_i} \Bigr|\Bigr|_{\rm F} $
		\EndFor 
		\State $\Delta\gets \Delta/N$
		\State $t \gets t+1$ 
	\EndWhile
	%%%
	\For{$\mu=1,\cdots,M$}  %\Comment{$O(ML^2)$}
		\State $\V{u}_{\mu}^{\bs \mu}\gets \V{u}_{\mu}+C_{\rm SA}^{(t)}\V{b}^{\mu} $
	\EndFor
	\State Compute $\LOOE$ from $\{ \V{u}_{\mu}^{\bs \mu} \}_{\mu}$ by \Req{LOOE} 
	\State \Return $\LOOE$
\EndProcedure
\end{algorithmic}
\end{algorithm}
%%%%%%%%%%%%%%%%%%%%%%%%%%%%%%%
Comments are added for specifying the time consuming parts in the entire procedures. In \Rcode{SAACV}, we describe an actual implementation for solving $C_{\rm SA}$ by recursion, which is not fully specified in \Rsec{Further simplified}. The symbol $||\cdot||_{\rm F}$ denotes the Frobenius norm and we set the threshold $\theta$ judging the convergence as $\theta=10^{-6}$ in typical situations. We also set as $10^{-6}$ the threshold judging if $\lambda_2$ is large or not.

%%%%%%%%%%%%%%%%%%%%%%%%%%%%%%%%%%%%%%%%%%%%%%%%%%%%%
%%%%%%%%%%%%%%%%%%%%%%%%%%%%%%%%%%%%%%%%%%%%%%%%%%%%%
%%%%%%%%%%%%%%%%%%%%%%%%%%%%%%%%%%%%%%%%%%%%%%%%%%%%%
\section{Numerical experiments}\Lsec{Numerical}
In this section, we examine the precision and actual computational time of ACV and SAACV in numerical experiments. Both simulated and actual datasets~\citep[from UCI machine learning repository, ][]{Lichman:13} are used. 

For examination, we compute the errors also by literally conducting $k$-fold CV with some $k$s, and compare it to the result of our approximate formula. In principle, we should compare our approximate result with that of the LOO CV ($k=M$) because our formula approximates it.  However for large $M$, the literal LOO CV requires huge computational burdens despite that the result is empirically not much different from that of the $k$-hold CV with moderate $k$s. Hence in some of the following experiments with large $M$, we use the $10$-hold CV instead of the LOO CV. Further, to directly check the approximation accuracy, we also compute the normalized error difference defined as
\be
\frac{\LOOE^{\mathrm{approximate}}-\epsilon_{\rm CV}^{\mathrm{literal}} }{\epsilon_{\rm CV}^{\mathrm{literal}}},
\Leq{NED}
\ee
where $\epsilon_{\rm CV}^{\mathrm{literal}}$ denotes the literal CV estimator of the prediction error while $\LOOE^{\mathrm{approximate}}$ is the approximated LOOE.
Moreover, as a reference, we compute the negative log-likelihood of the full solution $\{\hat{\V{w}}_a\}_{a=1}^{L}$ as:
\be
\TRE=
\frac{1}{M}\sum_{\mu=1}^{M}\NLL{\mu}\lb \lbb \hat{\V{w}}_{a} \rbb_{a=1}^{L} \rb,
\Leq{TRE}
\ee
and call it the training error, hereafter. The training error is expected to be a monotonic increasing function with respect to $\lambda$, while the prediction one is supposed to be non-monotonic. 

In all of the experiments, we used a single CPU of Intel(R) Xeon(R) E5-2630 v3 2.4GHz. To solve the optimization problems in \Reqs{w-full}{w-LOO}, we employed {\it Glmnet}~\citep{friedman2010regularization} which is implemented as a {\it MEX} subroutine in MATLAB\textsuperscript{\textregistered}. The two approximations were implemented as raw codes in MATLAB. This is not the most optimized approach, because as seen in \Rcodes{ACV}{SAACV} our approximate formula uses a number of {\it for} and {\it while} loops which are slow in MATLAB, and hence the comparison is not necessarily fair. However, even in this comparison there is a significant difference in the computational time between the literal CV and our approximations, as shown below. 

In Glmnet, the corresponding optimization problem is parameterized as follows:
\be
\{ \hat{\V{w}}_a(\tilde{\lambda},\eta)\}_a
=
\argmin_{\{ \V{w}_{a} \}_a}\lbb 
\frac{1}{M}\sum_{\mu=1}^{M}
\NLL{\mu}\lb \lbb \V{w}_{a} \rbb_{a=1}^{L} \rb
+\tilde{\lambda}\lb  \eta \sum_{a=1}^{L}||\V{w}_{a}||_1
+\frac{(1-\eta)}{2} \sum_{a=1}^{L}||\V{w}_{a}||_2^2
\rb
\rbb.
\Leq{w-glmnet}
\ee
In the following experiments, we present the results based on this parameterization. We basically prefer $\eta=1$ in which the $\ell_2$ term is absent, because the main contribution of the present paper is to overcome technical difficulties stemming from the $\ell_1$ term. However, Glmnet or its employing algorithm occasionally loses its stability in some uncontrolled manner without the $\ell_2$ term. Hence, in the following experiments we adaptively choose the value of $\eta$.\footnote{When employing our distributed codes implementing the approximate formula~\citep{obuchi:acv_mlr_matlab,obuchi:acv_mlr_python} in conjunction with Glmnet, the parameters $\lambda_1$ and $\lambda_2$ are read as $\lambda_1=M\tilde{\lambda}\eta$ and $\lambda_2=M\tilde{\lambda}(1-\eta)$.}  

A sensitive point which should be noted is the convergence problem of the algorithm for solving the present optimization problem. In Glmnet, a specialized version of coordinate descent methods is employed, and it requires a threshold $\delta$ to judge the algorithm convergence. Unless explicitly mentioned, we set this as $\delta=10^{-8}$ being tighter than the default value. This is necessary since we treat problems of rather large sizes. A looser choice for $\delta$ rather strongly affects the literal CV result, while it does not change the full solution or the training error as much. As a result, our approximations employing only the full solution are rather robust against the choice of $\delta$ compared to the literal CV. This is also demonstrated below.

%%%%%%%%%%%%%%%%%%%%%%%%%%%%%%%%%%%%%%%%%%%%%%%%%%%%%
%%%%%%%%%%%%%%%%%%%%%%%%%%%%%%%%%%%%%%%%%%%%%%%%%%%%%
\subsection{On simulated dataset}\Lsec{On simulated}
Let us start by testing with the simulated data. Suppose each ``true'' feature vector $\V{w}_{0a} $ is independently identically drawn (i.i.d.) from the following Bernoulli-Gaussian prior:
\be
\V{w}_{0a} \sim \prod_{i=1}^{N}\lbb (1-\rho_0)\delta(w_{0ai})+\rho_0 \mathcal{N}(0,1/\rho_0)\rbb,
\ee
where $\mathcal{N}(\mu,\sigma^2)$ denotes a Gaussian distribution whose mean and variance are $\mu$ and $\sigma^2$, respectively. The resultant feature vector $\V{v}_a$ becomes $N\rho_0(\equiv K_0)$-sparse and its norm becomes $\sqrt{N}$ on average. Then, we choose a class $y_{\mu}$ from $\{1,\cdots,L\}$ uniformly and randomly, and generate an observed feature vector $\V{x}_{\mu}$ by leveraging the following linear process: 
\be
\V{x}_{\mu}=\frac{\V{w}_{0y_{\mu}}}{\sqrt{N}}+\V{\xi},
\ee
where $\V{\xi}$ is an observation noise each component of which is i.i.d. from a Gaussian $\mathcal{N}(0,\sigma_N^2)$. 

For convenience, we introduce the ratio of the data size to the feature dimensionality, $\alpha=M/N$, and now obtain five parameters $\{ N,L,\alpha,\rho_0,\sigma_{\xi}^2 \}$ characterizing the experimental setup. It is rather heavy to obtain the dependence of all parameters and below, and hence we mainly focus on the dependence on $L$, $\sigma_{\xi}^2$, and $N$. Other parameters are set to be $\alpha=2$ and $\rho_0=0.5$.

%%%%%%%%%%%%%%%%%%%%%%%%%%%%%%%%%%%%%%%%%%%%%%%%%%%%%
\subsubsection{Result}\Lsec{Result}
Let us summarize the result on simulated data.

\Rfig{simulate-L} shows the plots of the prediction and training errors against $\tilde{\lambda}$ for $L=4,8,16$ at $N=200$ and $\sigma_{\xi}^2=0.01$. 
%%%%%%%%%%%%%%%%%%%%%
\begin{figure}[htbp]
\begin{center}
 \includegraphics[width=0.32\columnwidth]{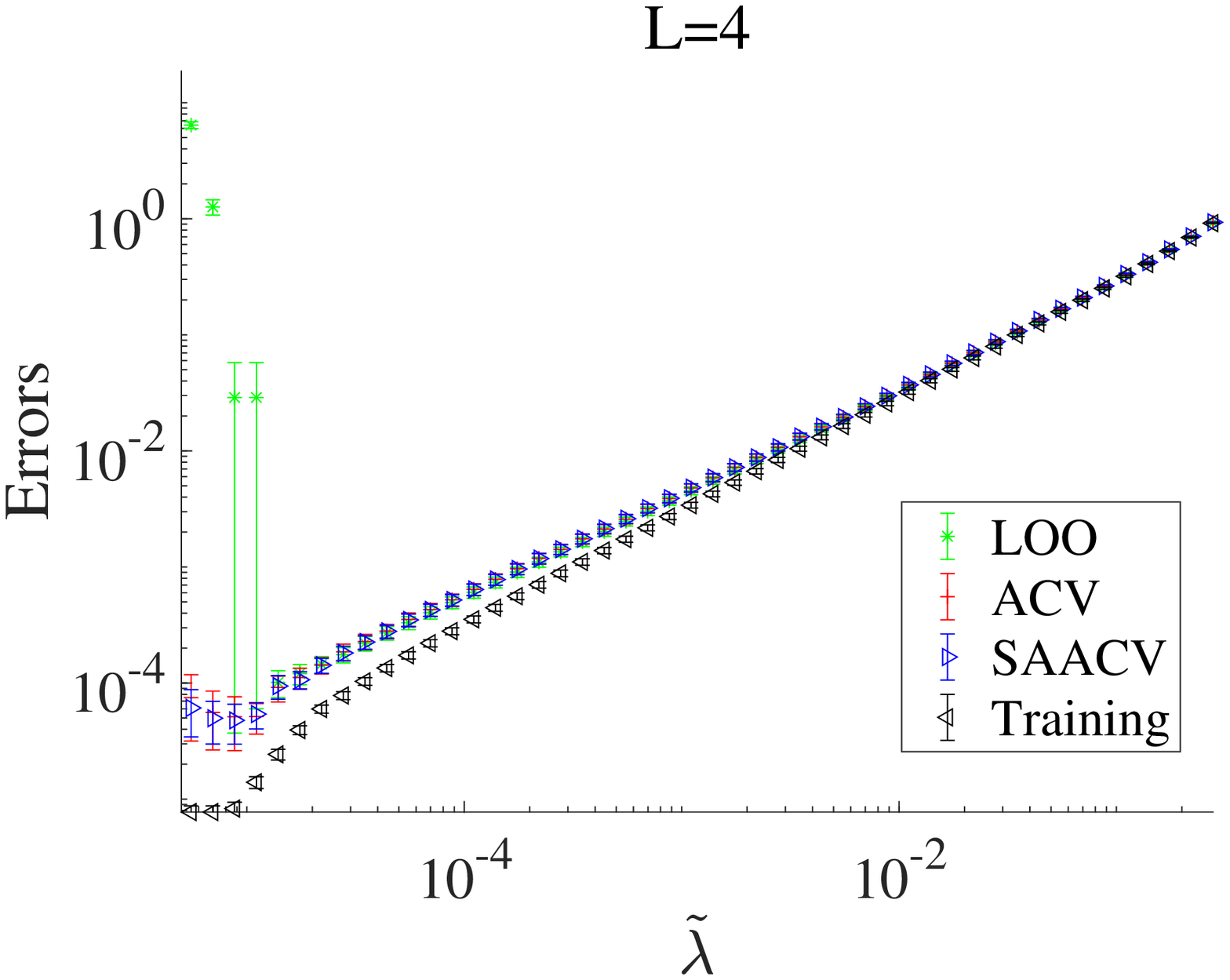}
 \includegraphics[width=0.32\columnwidth]{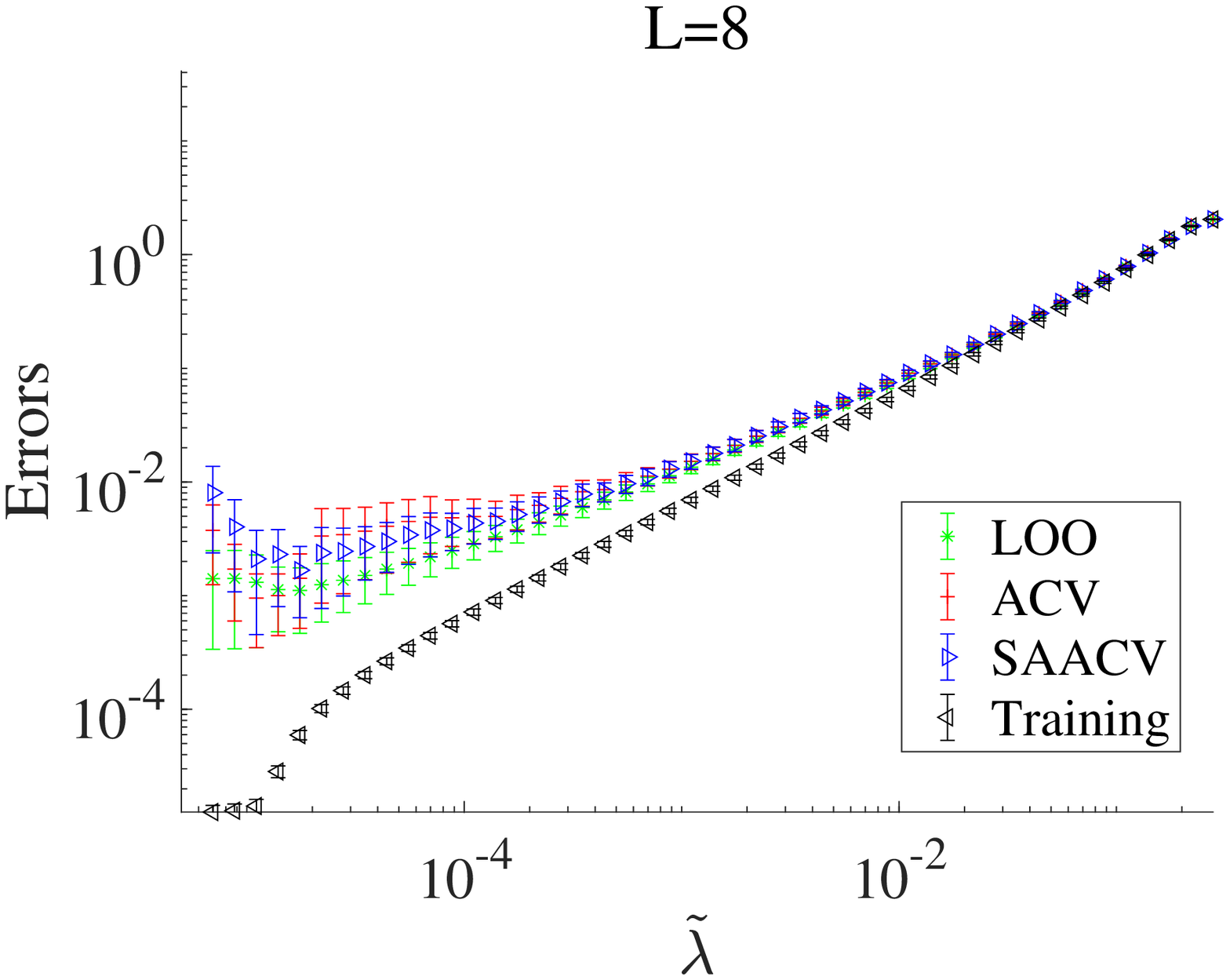}
 \includegraphics[width=0.32\columnwidth]{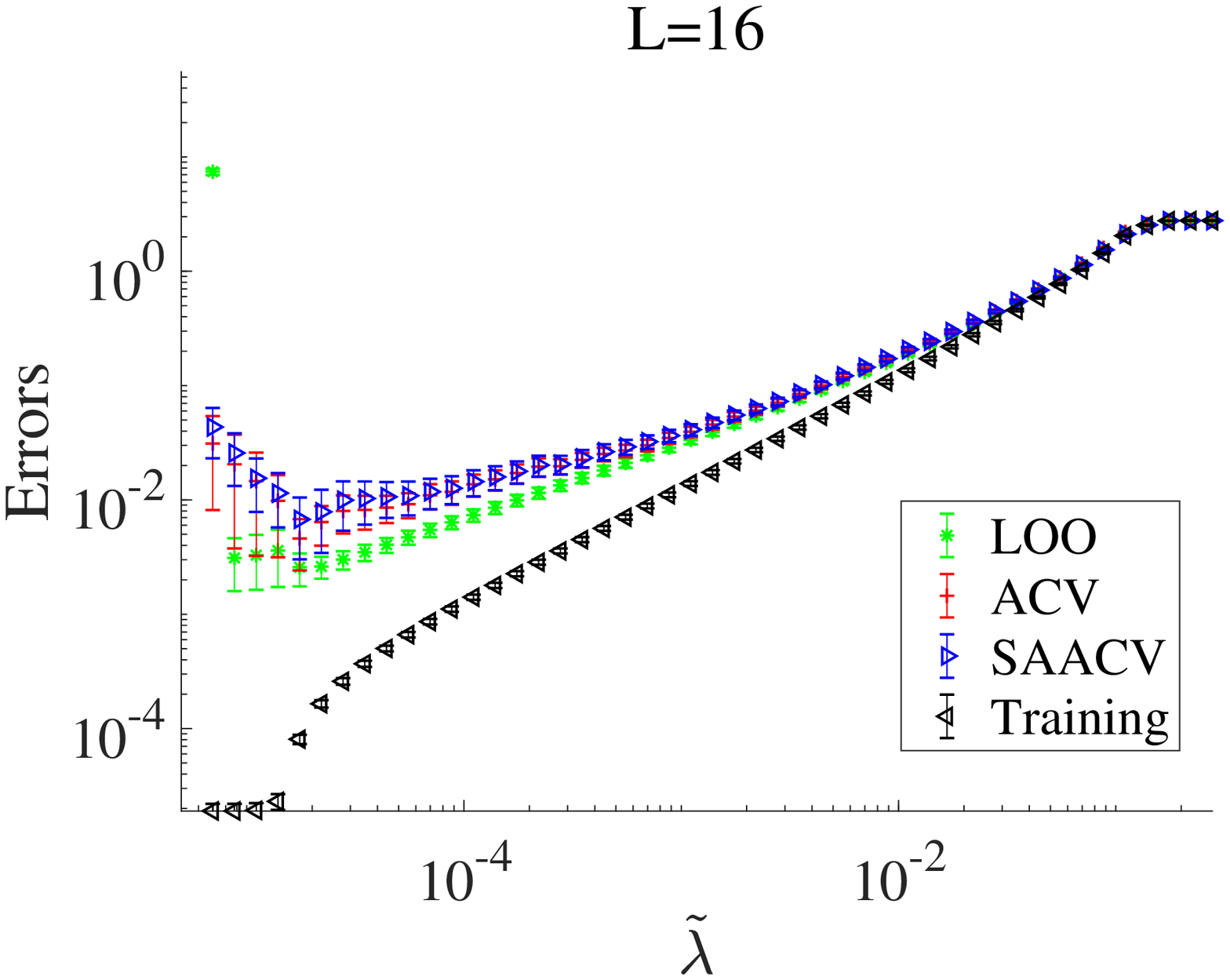}
 \includegraphics[width=0.32\columnwidth]{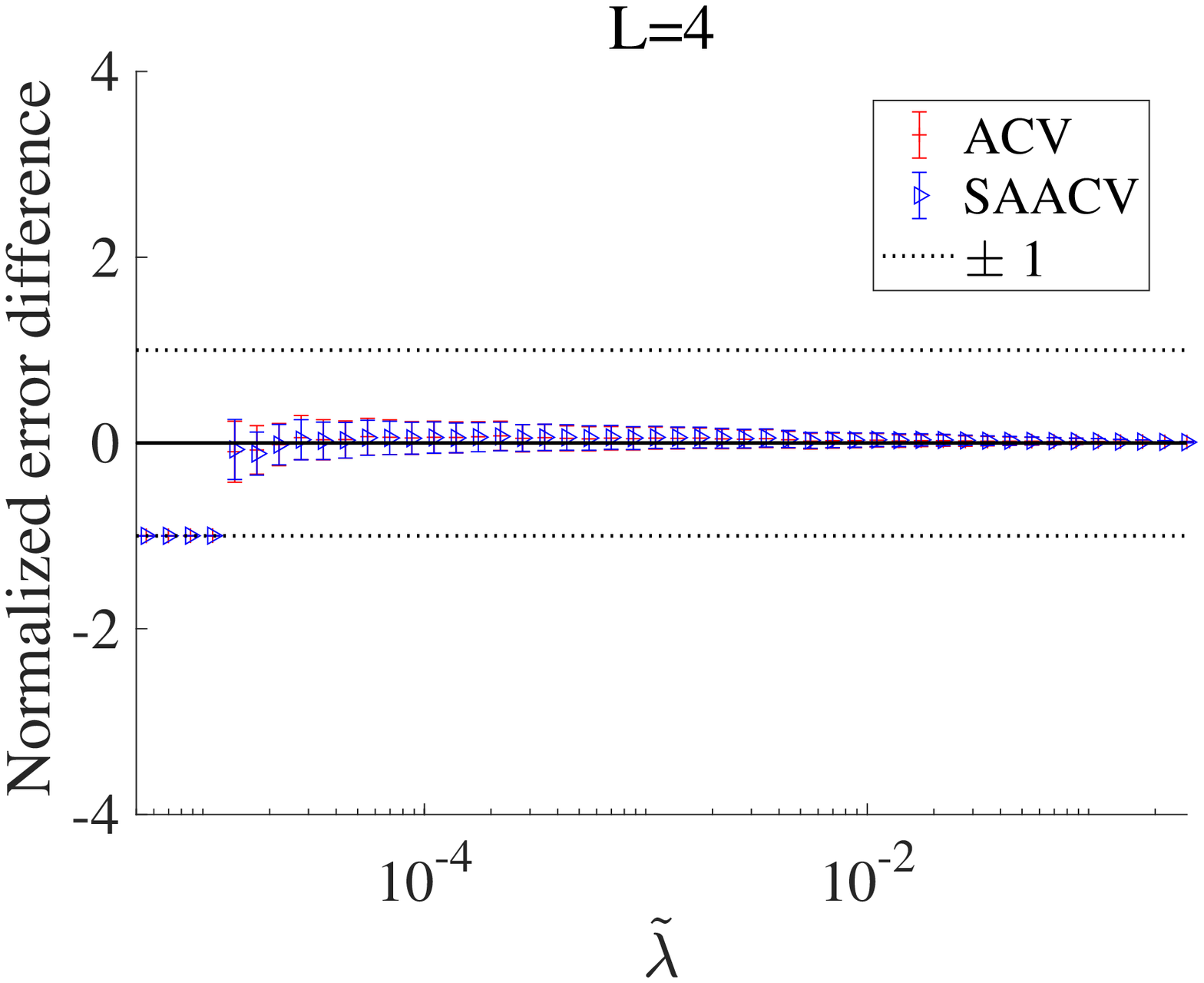}
 \includegraphics[width=0.32\columnwidth]{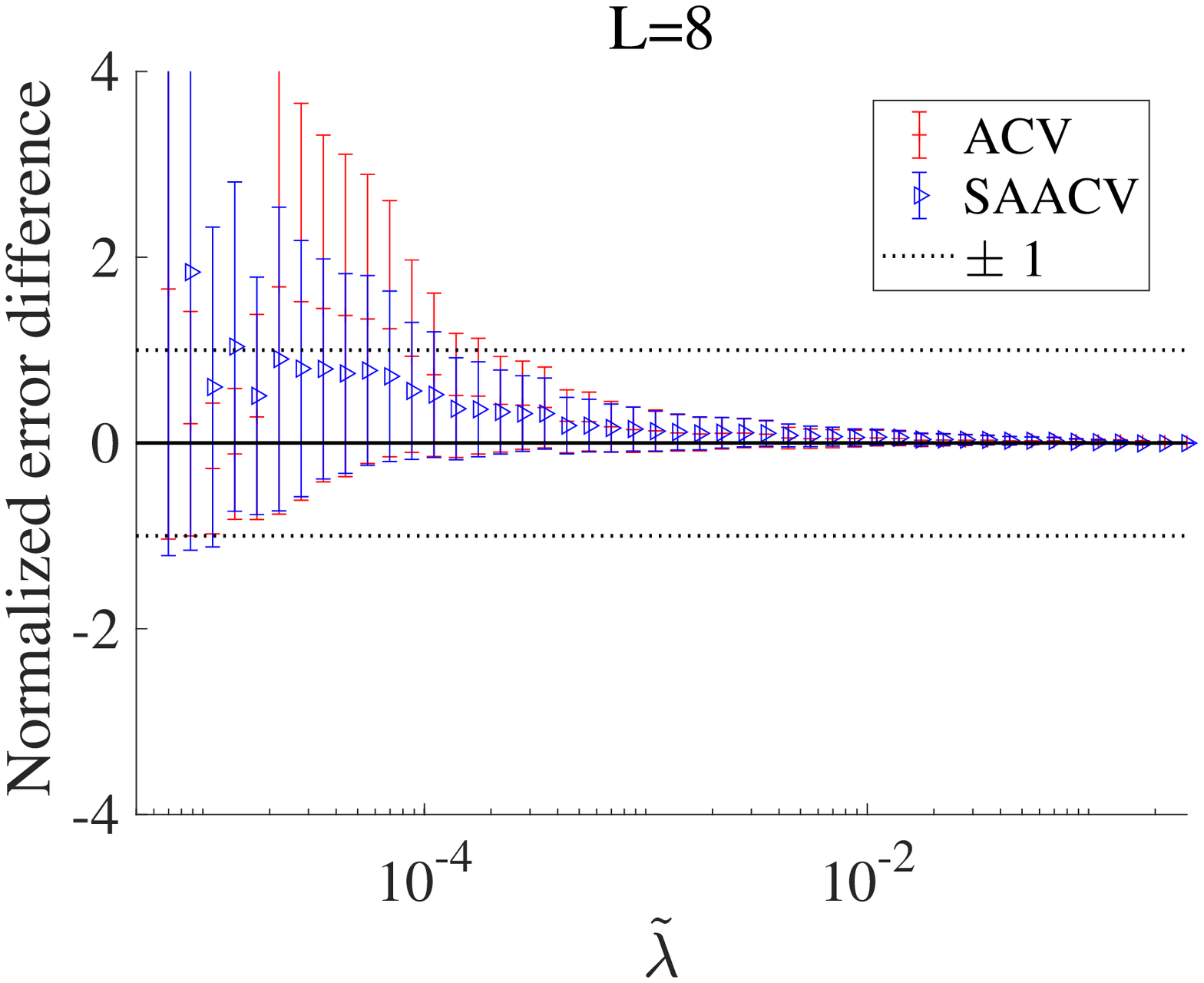}
 \includegraphics[width=0.32\columnwidth]{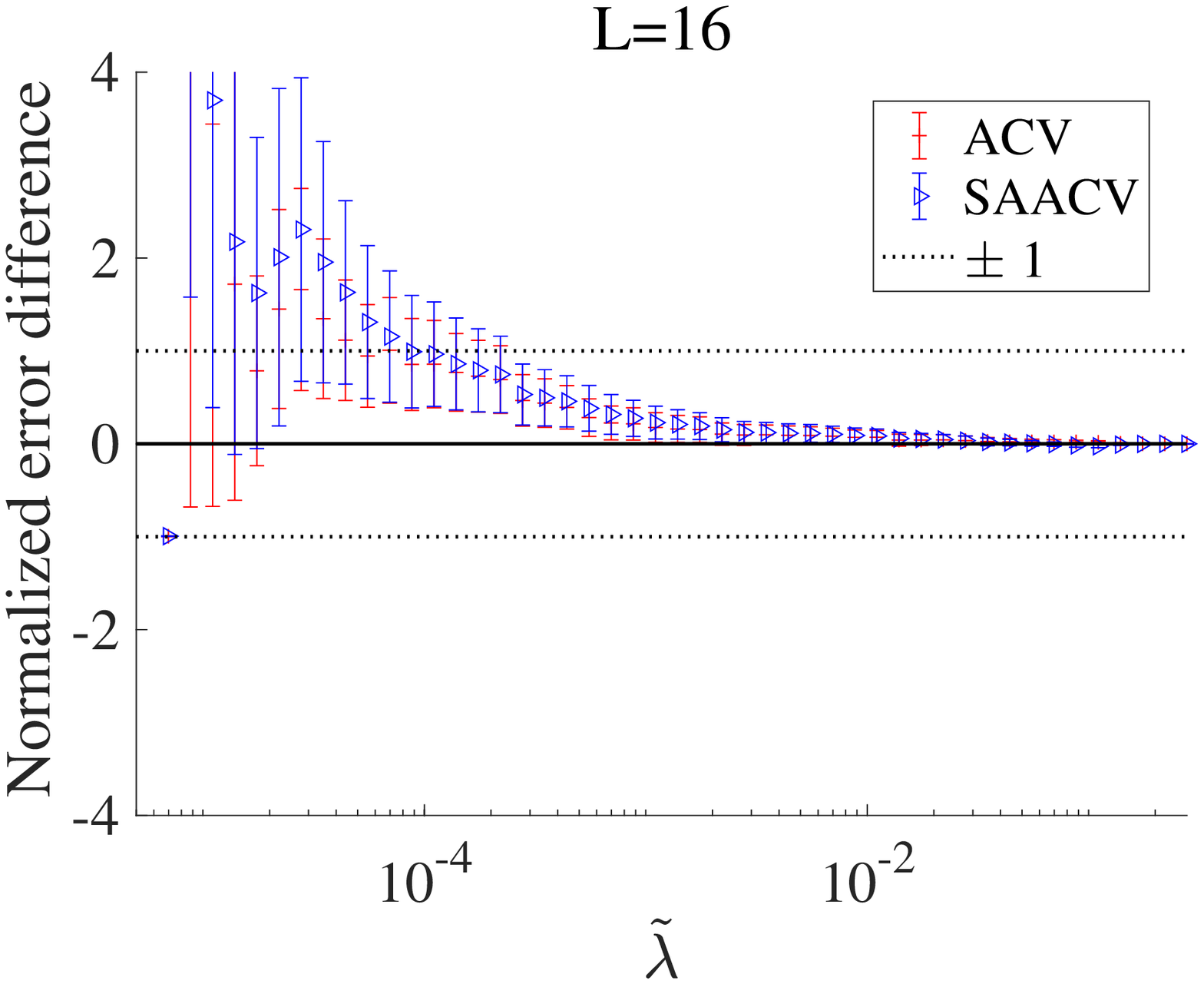}
\caption{(Upper) Log-log plots of the errors against $\tilde{\lambda}$ for several values of the class number $L$. Other parameters are fixed at $N=200$, $\sigma_{\xi}^2=0.01$, $\alpha=2$ and $\rho_0=0.5$. The approximation results are consistent with the literal LOO CV results, except at small $\lambda$s, which is presumably due to a numerical instability occurring in the literal CV at small $\lambda$s. Here, $\eta=0.9$. (Lower) The normalized error difference \NReq{NED} plotted against $\tilde{\lambda}$. The parameters of each panel are them of the corresponding upper one. The horizontal dotted lines denote $\pm 1$ and drawn for comparison. For small $\tilde{\lambda}$s the difference is not negligible, but the literal CV itself is not stable in that region and hence the error difference is not reliable. 
}
\Lfig{simulate-L}
\end{center}
\end{figure}
%%%%%%%%%%%%%%%%%%%%%
This demonstrates that both approximations provide consistent results with the literal LOO CV, except at small $\tilde{\lambda}$s. This inconsistency at small $\tilde{\lambda}$s is considered to be due to a numerical instability occurring in the literal CV. Actually, for small $\tilde{\lambda}$s, we have observed that certain small changes in the data induce large differences in the literal CV result. This example demonstrates that our approximations provide robust curves even in such situations. Note that as $L$ grows the number of estimated parameters $\{\V{w}_a\}_{a=1}^{L}$ increases while the data size $M=\alpha N=400$ is fixed, meaning that the problem becomes more and more underdetermined with the growth of $L$. Hence, \Rfig{simulate-L} demonstrates that the developed approximations work irrespectively of how much the problem is underdetermined.

\Rfig{simulate-noise} exhibits the $\sigma_{\xi}^2$-dependence of the errors and the approximation results for $L=8$ and $N=200$. 
%%%%%%%%%%%%%%%%%%%%%
\begin{figure}[htbp]
\begin{center}
 \includegraphics[width=0.32\columnwidth]{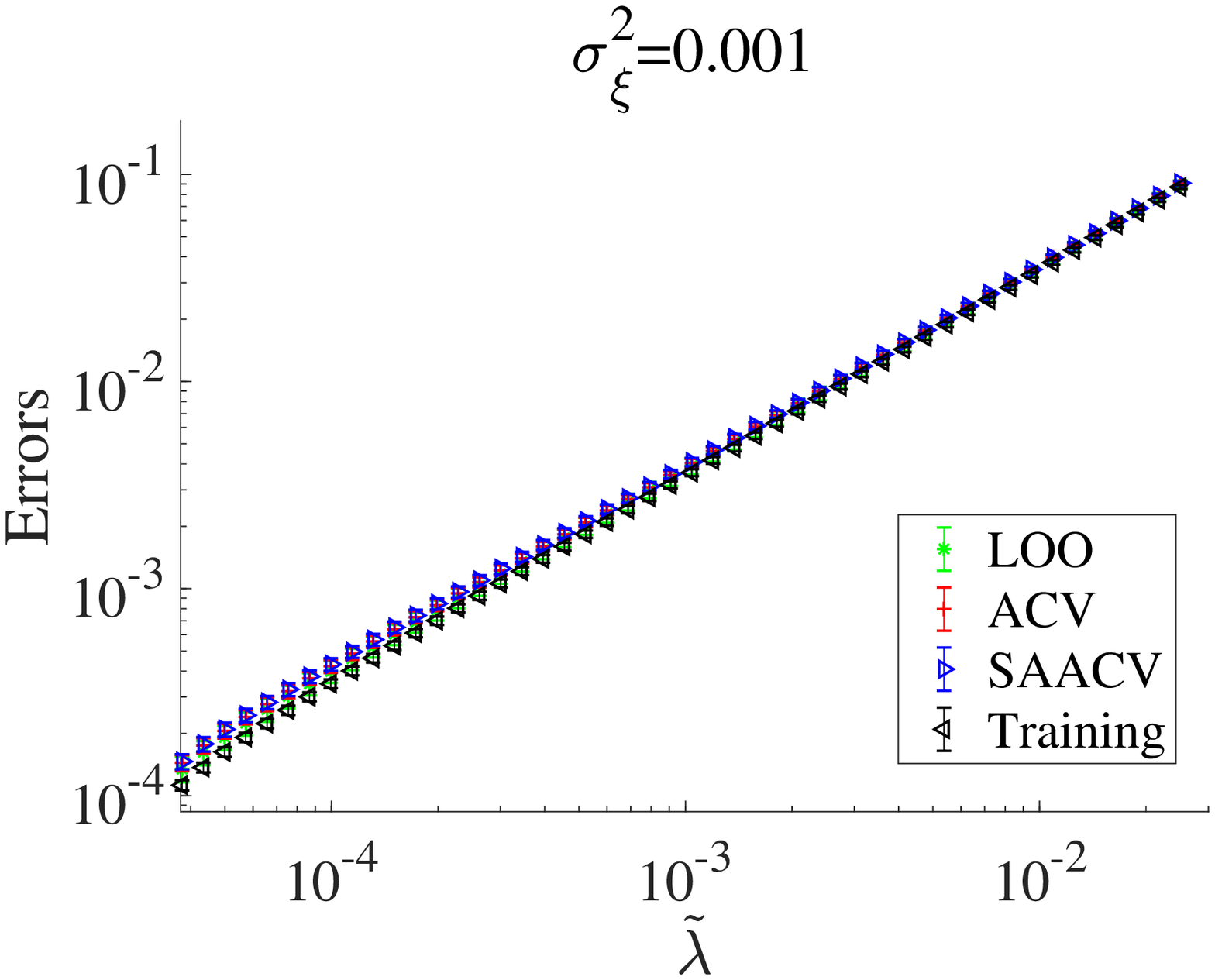}
 \includegraphics[width=0.32\columnwidth]{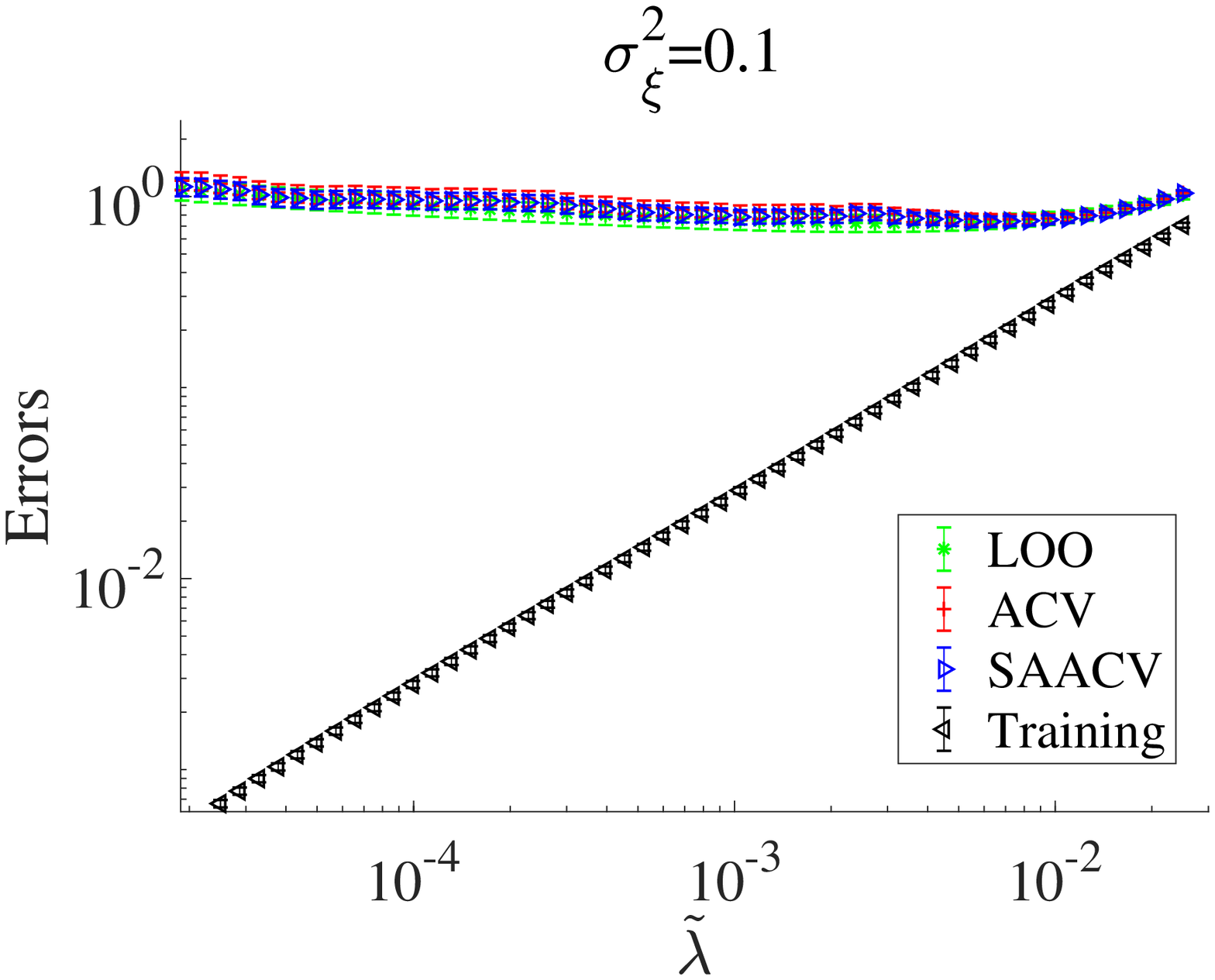}
 \includegraphics[width=0.32\columnwidth]{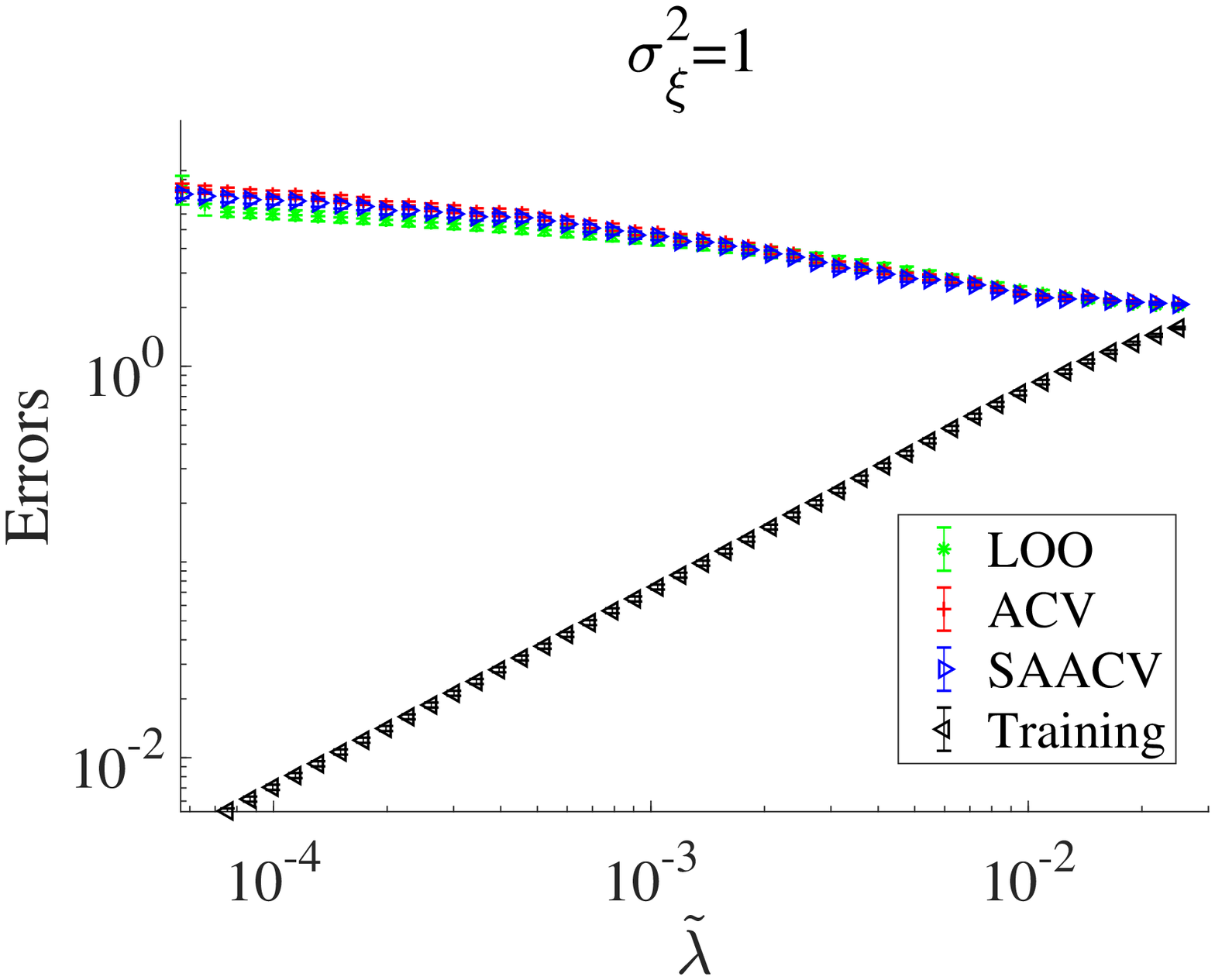}
 \includegraphics[width=0.32\columnwidth]{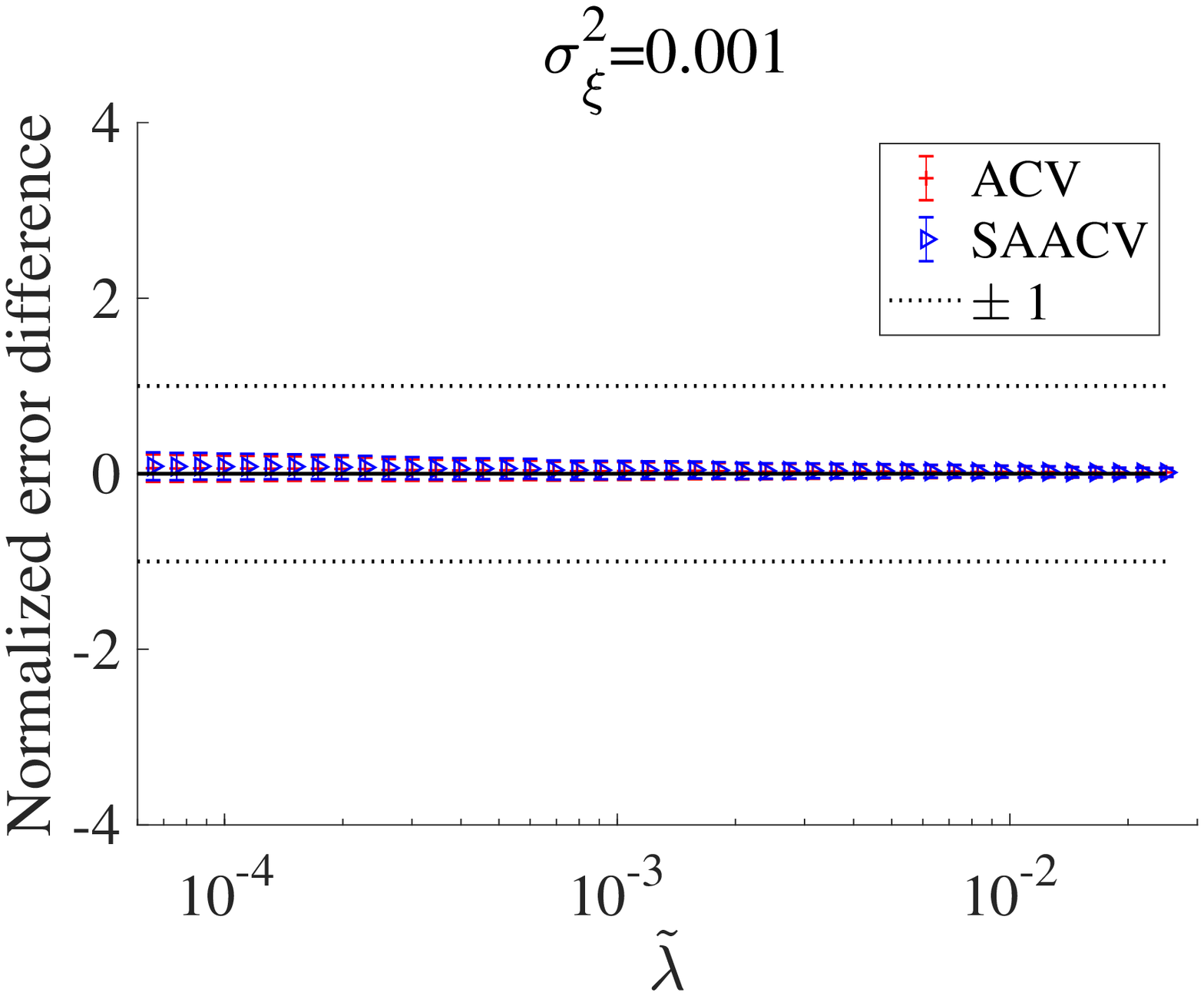}
 \includegraphics[width=0.32\columnwidth]{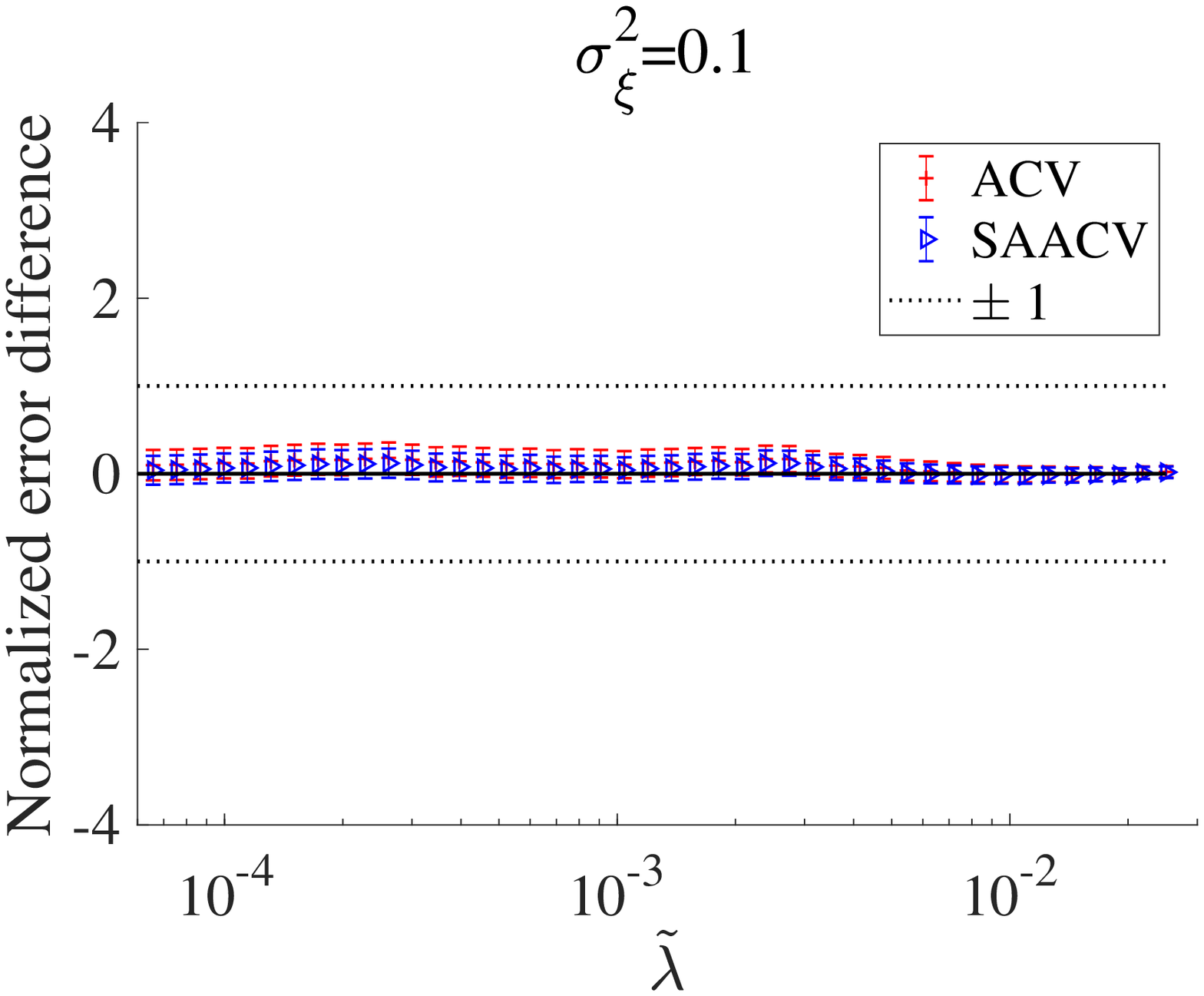}
 \includegraphics[width=0.32\columnwidth]{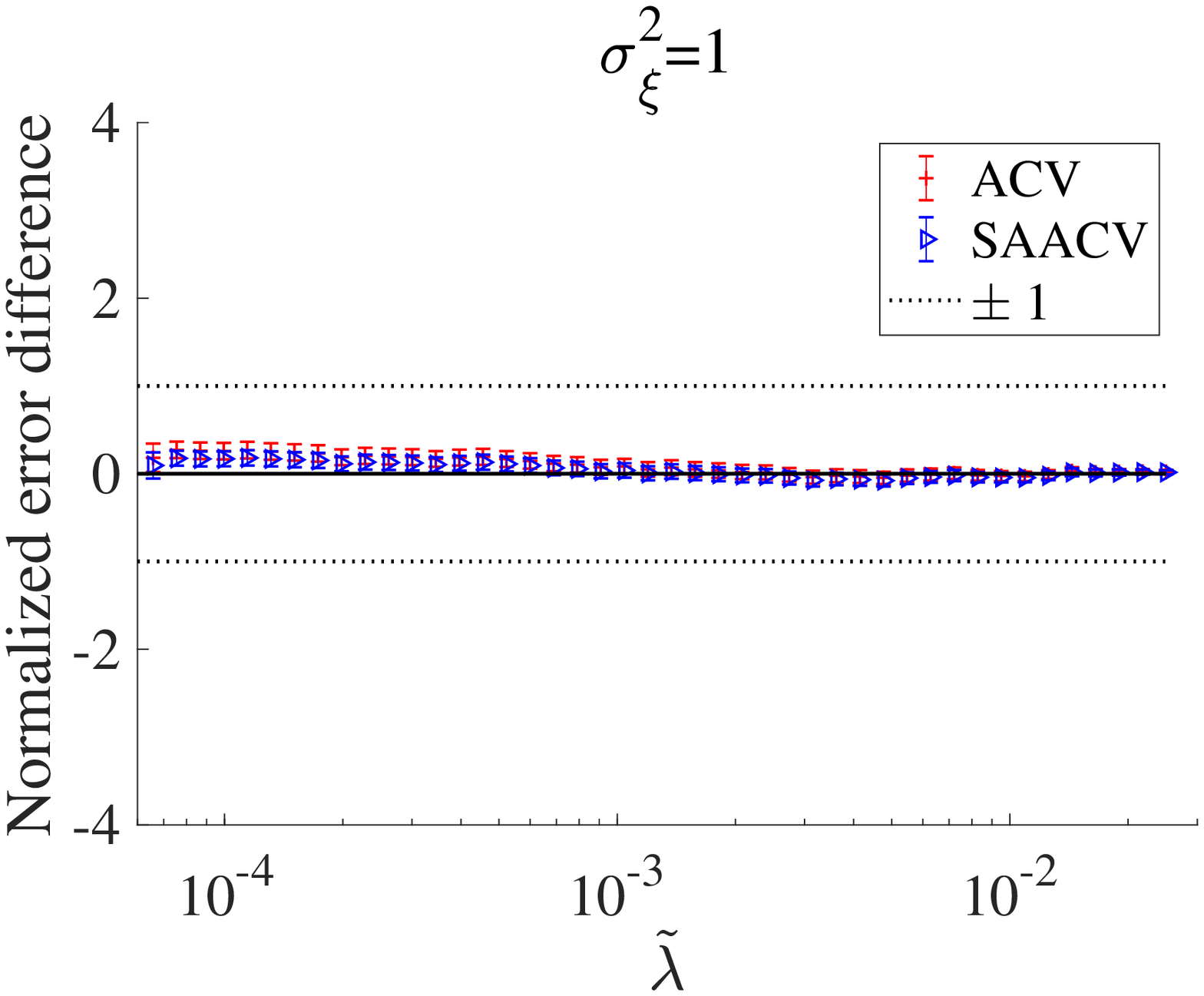}
\caption{(Upper) Log-log plots of the errors against $\tilde{\lambda}$ for several noise strengths. Other parameters are fixed at $N=200$, $L=8$, $\alpha=2$ and $\rho_0=0.5$. The approximation results are consistent with the literal LOO CV, irrespectively of the noise strength. The convergence threshold $\delta$ is set to be $\delta=10^{-9}$ for the case $\sigma_{\xi}^2=1$.  Here, $\eta=1$. (Lower) The normalized error difference \NReq{NED} plotted against $\tilde{\lambda}$. The parameters of each panel are them of the corresponding upper one. In the whole region the difference is negligibly small.}
\Lfig{simulate-noise}
\end{center}
\end{figure}
%%%%%%%%%%%%%%%%%%%%%
For the very weak noise case ($\sigma_{\xi}^2=0.001$, left), the difference between the predictive and training errors is negligible and hence all four curves are not discriminable. For the moderate ($\sigma_{\xi}^2=0.1$, middle) and large ($\sigma_{\xi}^2=1$, right) noise cases, the training curve is very different from the predictive ones. The approximation curves are again consistent with the literal LOO one. 

\Rfig{simulate-size} demonstrates how the approximation accuracy changes as the system size $N$ grows. 
%%%%%%%%%%%%%%%%%%%%%
\begin{figure}[htbp]
\begin{center}
 \includegraphics[width=0.45\columnwidth]{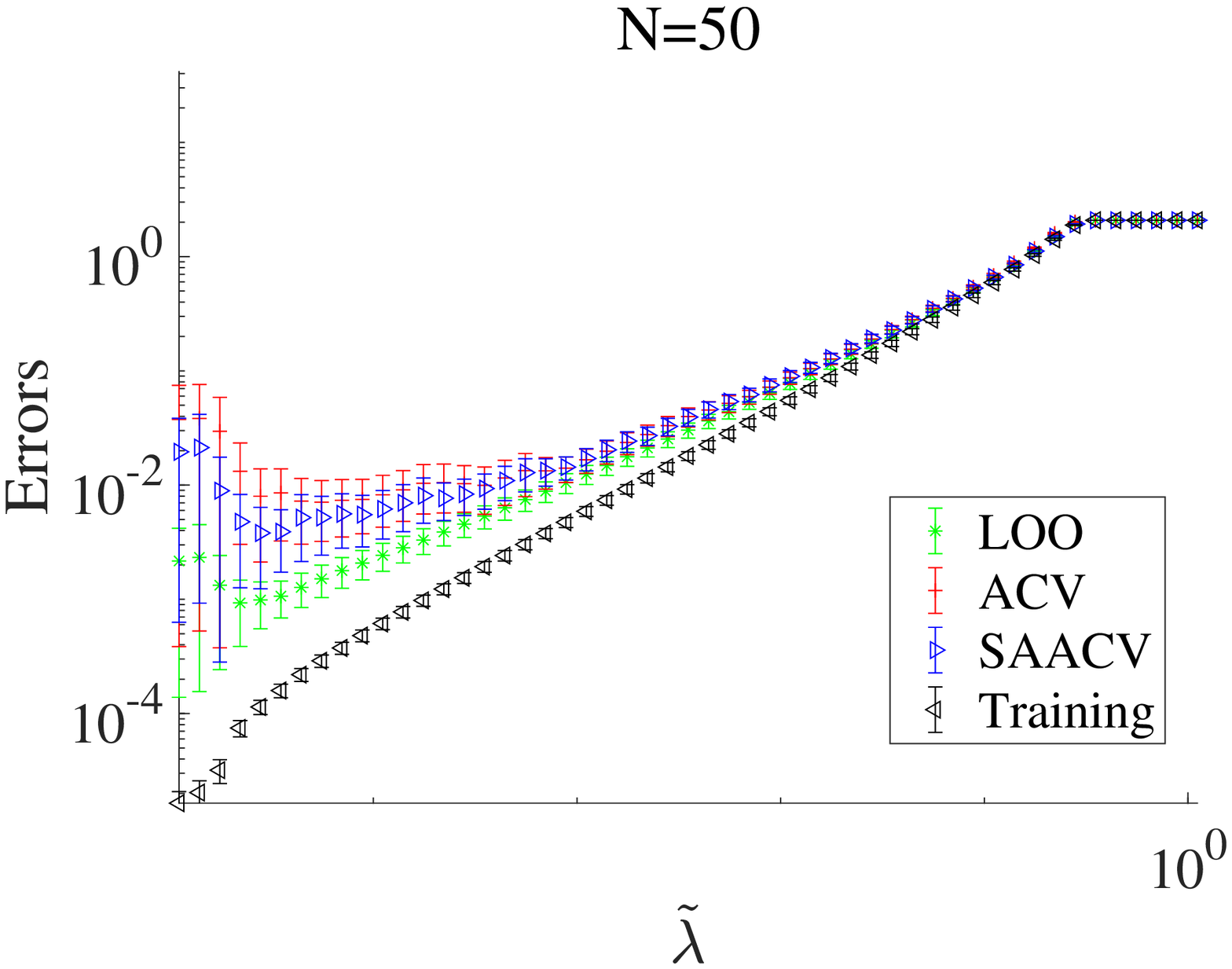}
 \includegraphics[width=0.45\columnwidth]{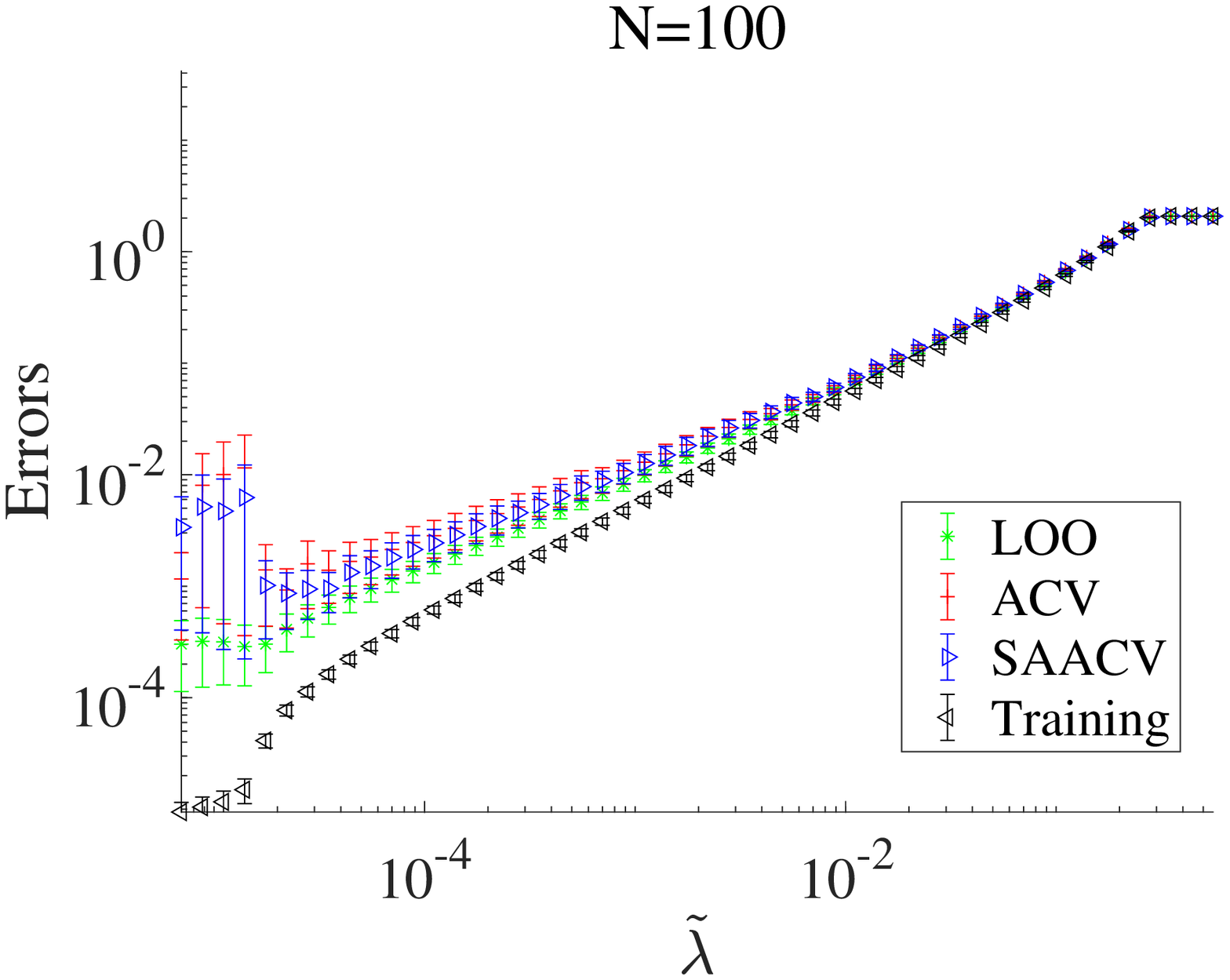}
 \includegraphics[width=0.45\columnwidth]{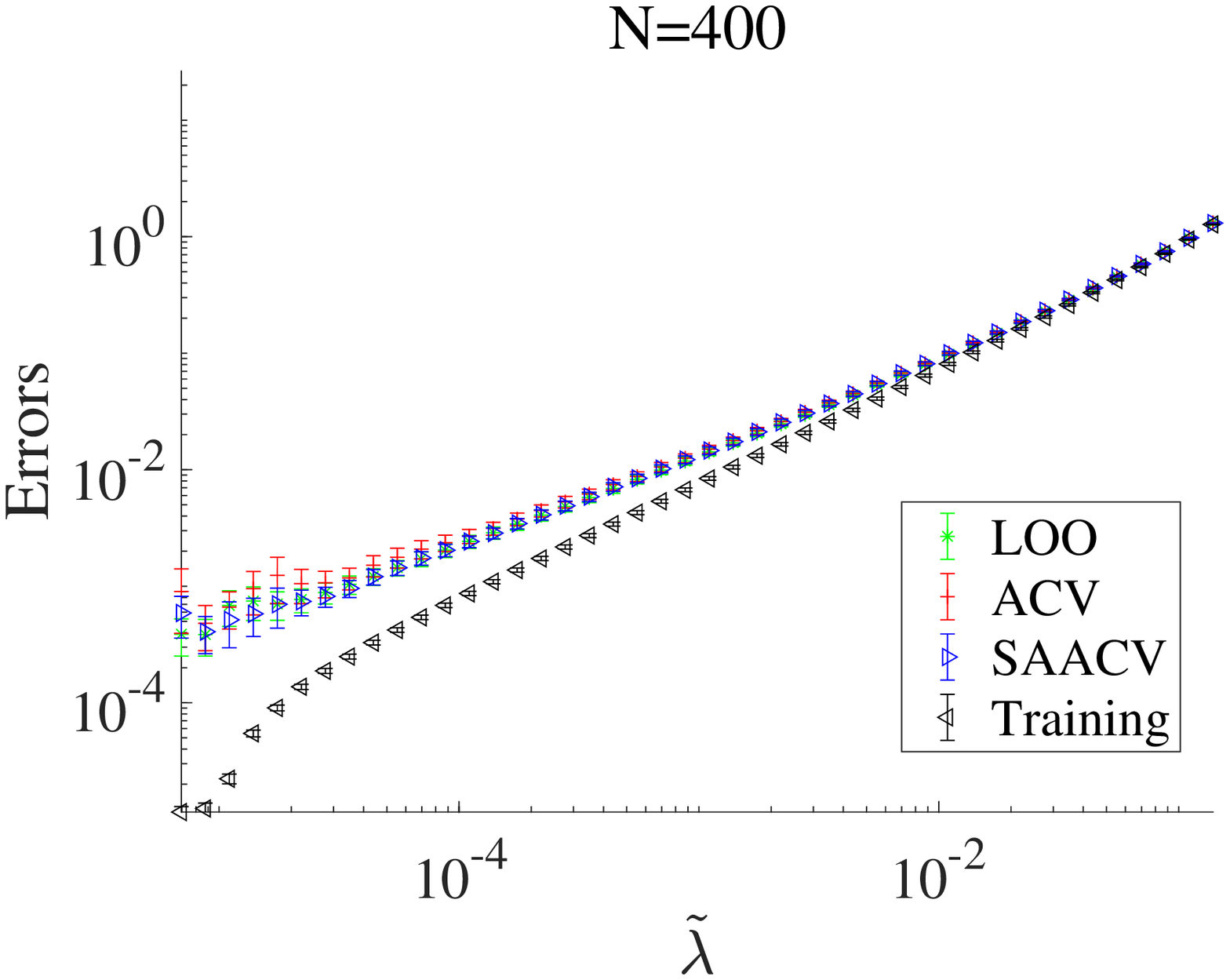}
 \includegraphics[width=0.45\columnwidth]{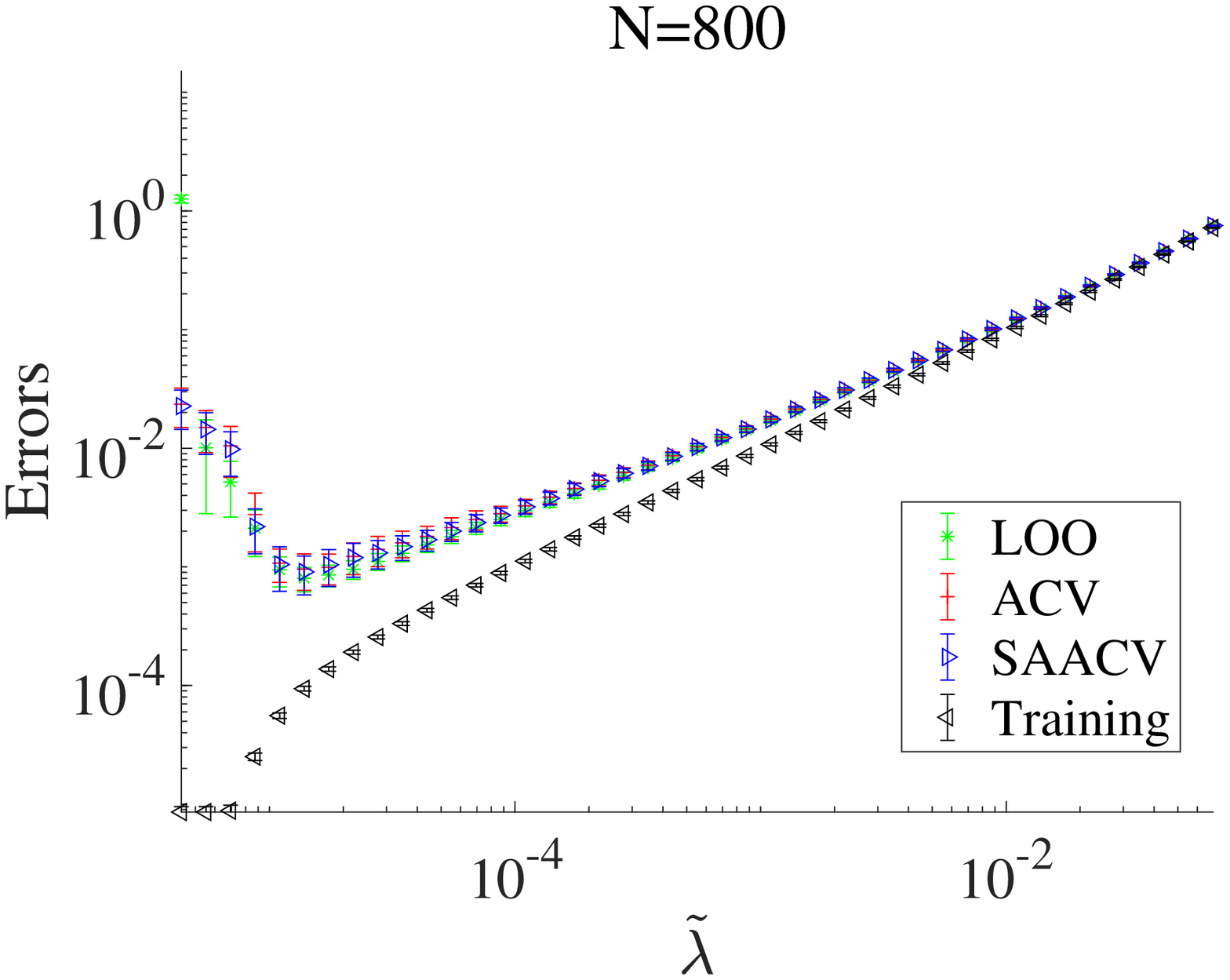}
\caption{Log-log plots of the errors against $\tilde{\lambda}$ for several values of feature dimensionality $N$. Other parameters are fixed at $L=8$, $\sigma_{\xi}^2=0.01$, $\alpha=2$ and $\rho_0=0.5$. Here, $\eta=0.9$.}
\Lfig{simulate-size}
\end{center}
\end{figure}
%%%%%%%%%%%%%%%%%%%%%
For small sizes $N=50,100$, a discriminable difference exists between the results of the approximations and the literal LOO CV, as well as the difference between the results of the two approximations. This is expected, because our derivation relies on the largeness of $N$ and $M$. For large systems $N=400, 800$, the difference among the two approximations and the literal CV is much smaller. Considering this example in conjunction with the middle panel of \Rfig{simulate-L}, we can recognize that our approximate formula becomes fairly precise for $N\geq 200$ in this parameter set. The normalized error difference corresponding to \Rfig{simulate-size} is shown in \Rfig{diff-simulate-size}. 
%%%%%%%%%%%%%%%%%%%%%
\begin{figure}[htbp]
\begin{center}
 \includegraphics[width=0.45\columnwidth]{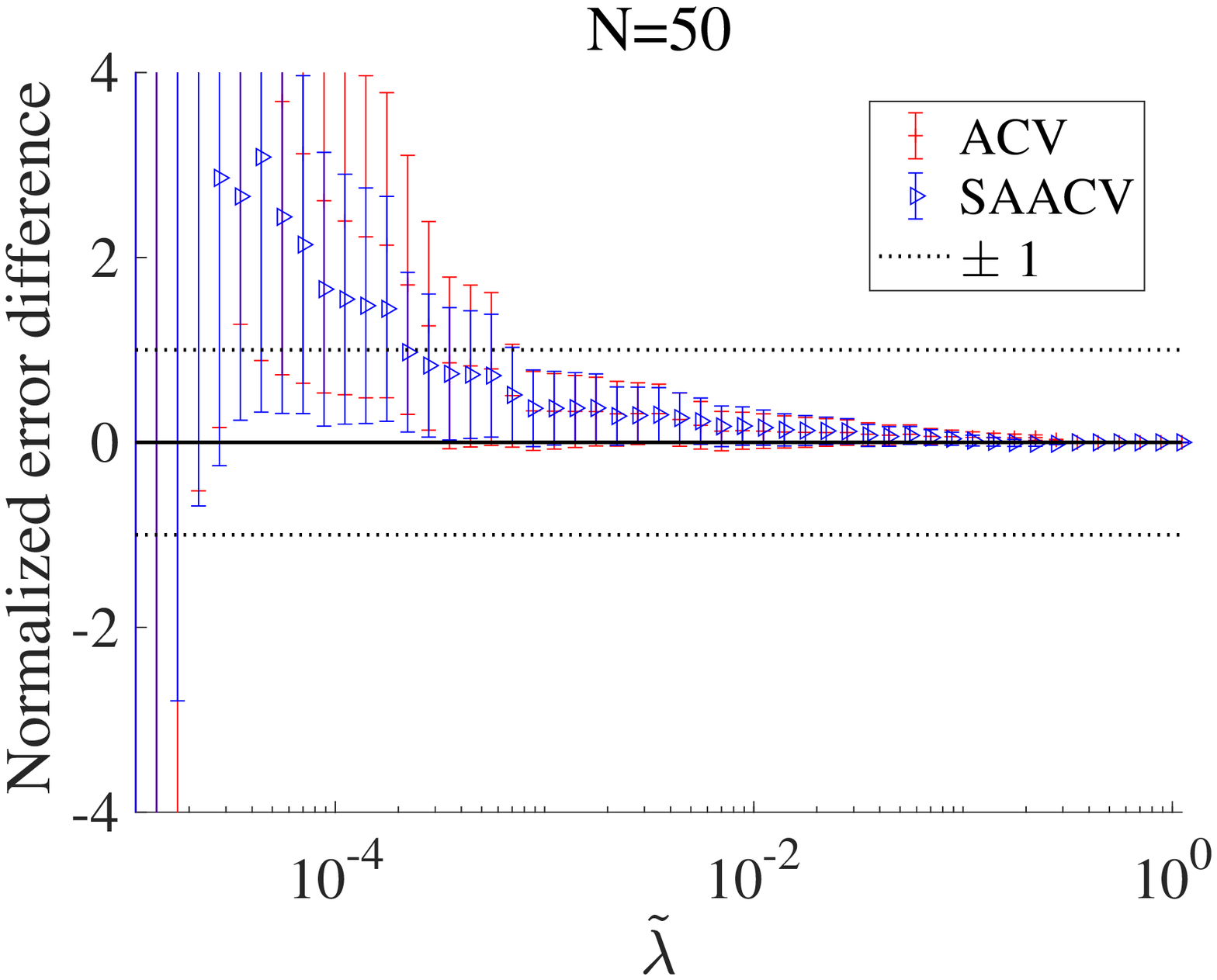}
 \includegraphics[width=0.45\columnwidth]{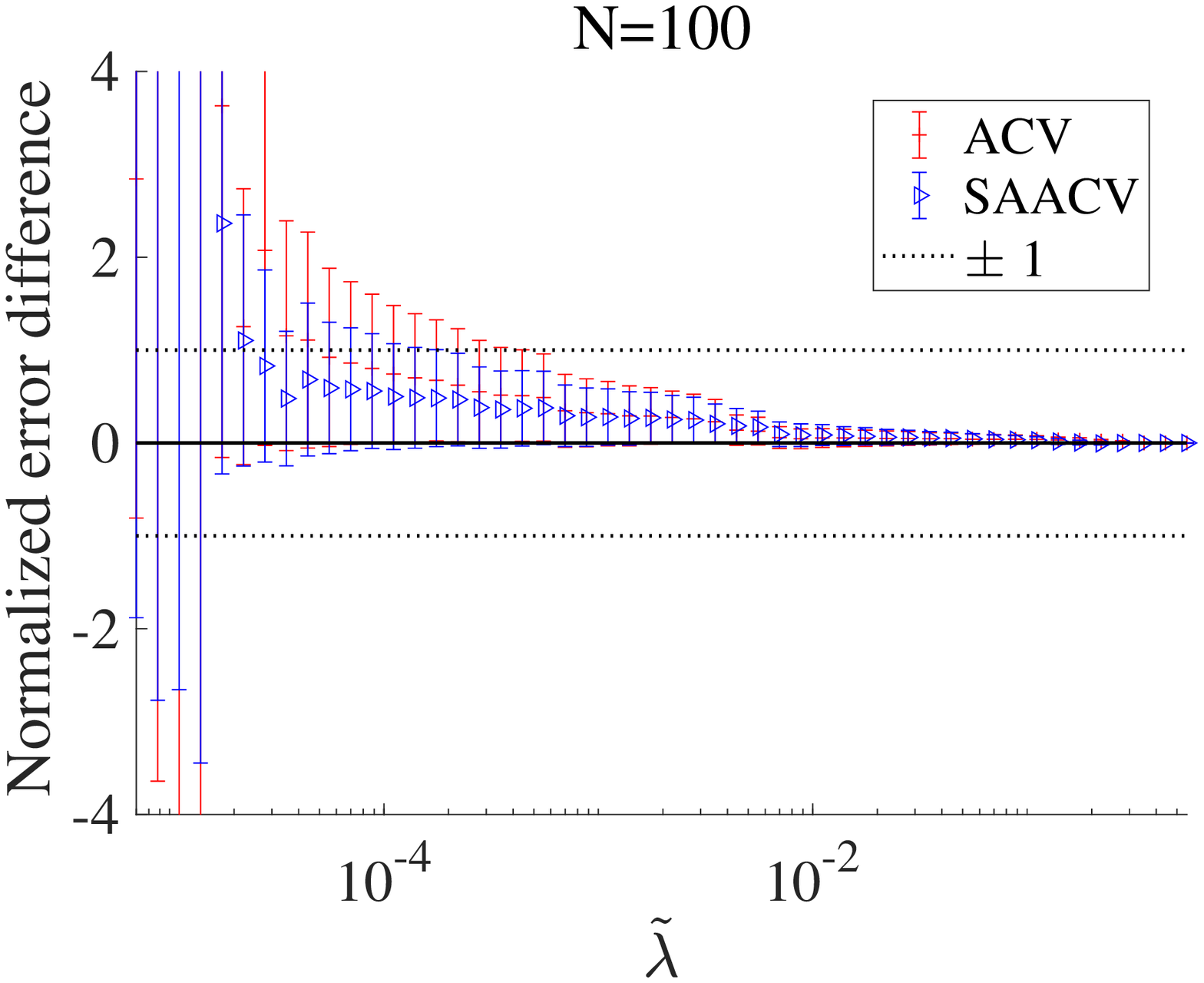}
 \includegraphics[width=0.45\columnwidth]{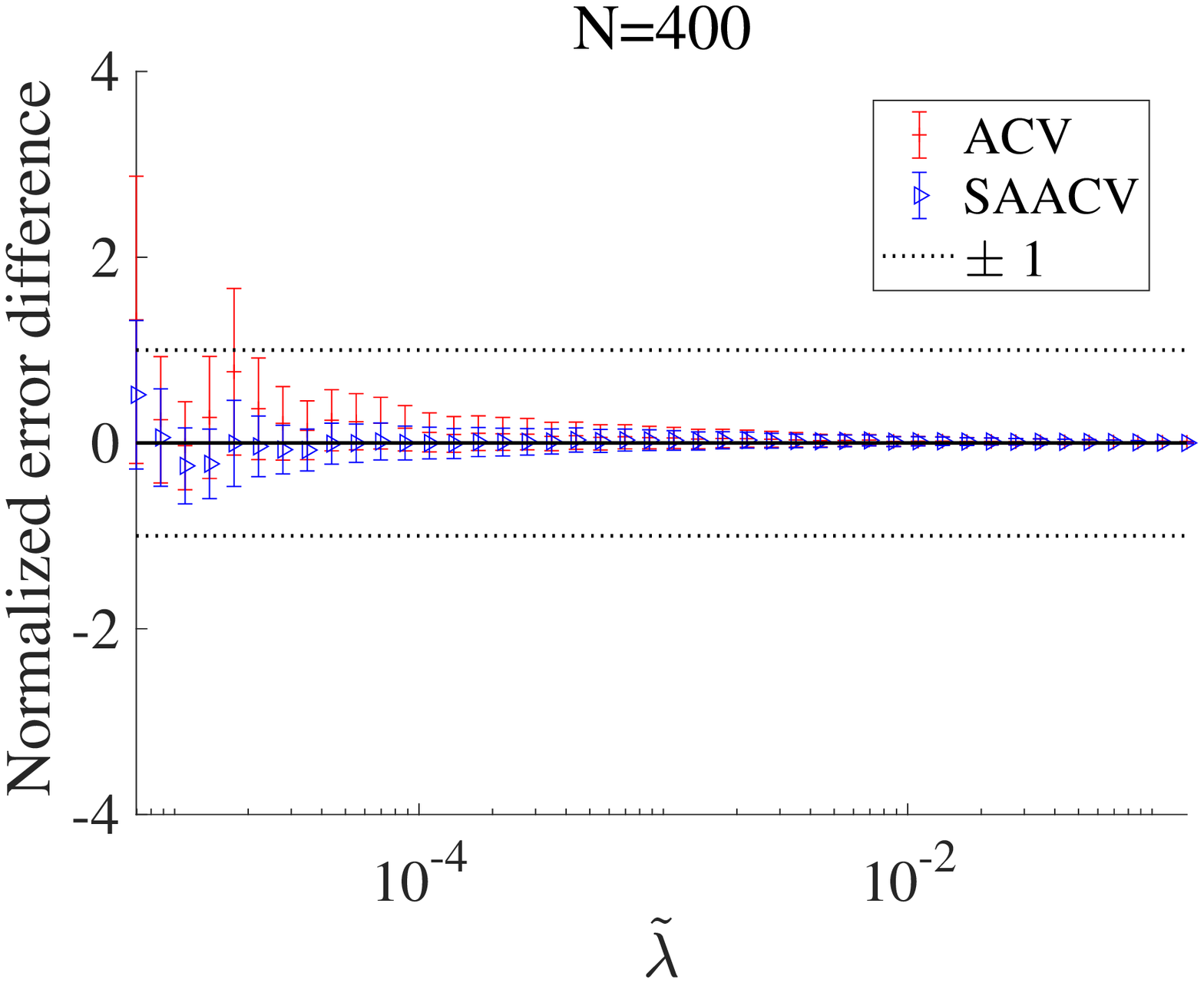}
 \includegraphics[width=0.45\columnwidth]{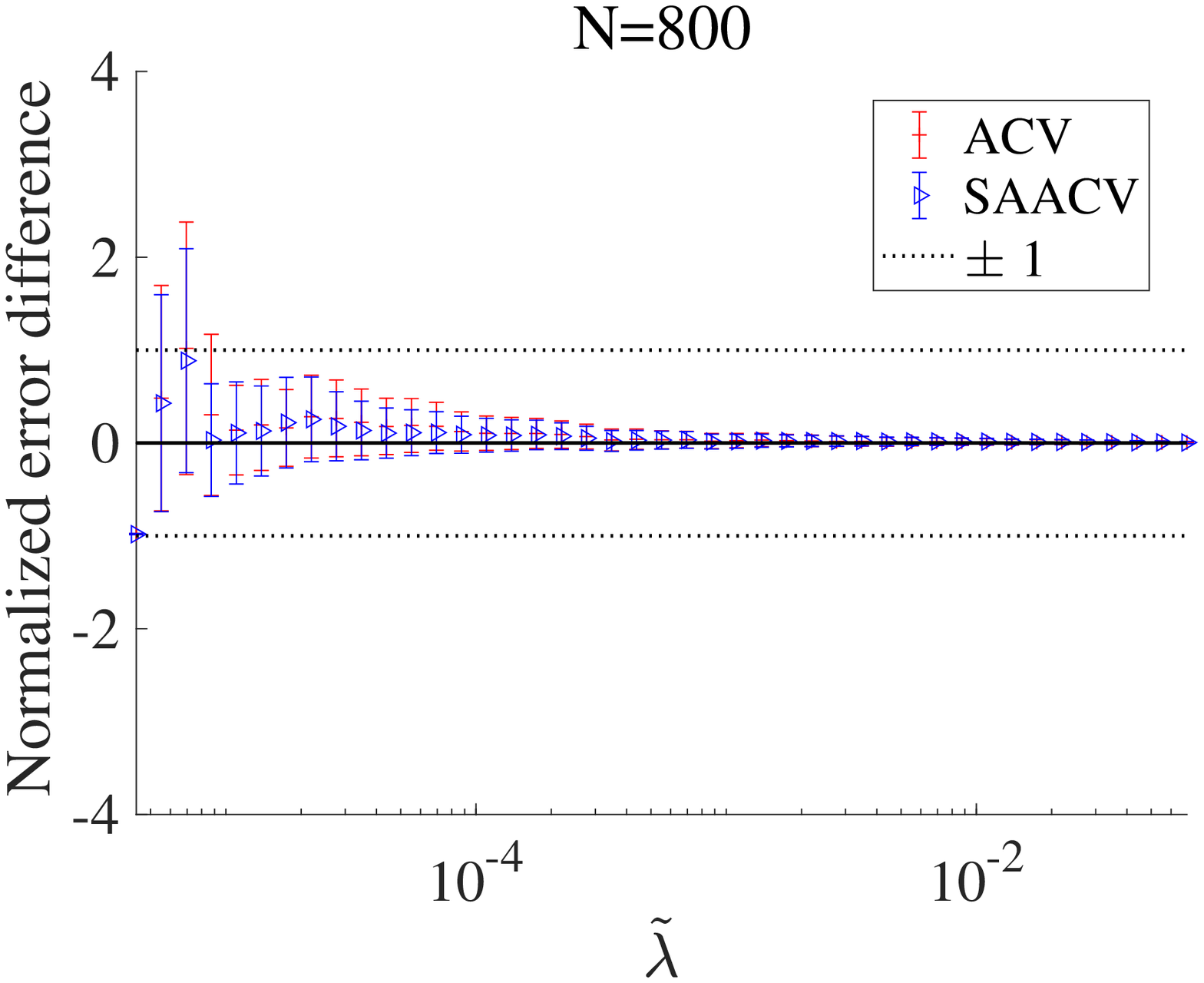}
\caption{The plot of the normalized error difference corresponding to \Rfig{simulate-size}. The difference tends to be smaller as the system size increases. }
\Lfig{diff-simulate-size}
\end{center}
\end{figure}
%%%%%%%%%%%%%%%%%%%%%
We can observe that the difference tends to be smaller as the system size increases, which is expected because the perturbation employed in our approximate formula is justified in the large $N,M$ limit.

Finally, let us consider the actual computational time to evaluate $\{ \hat{\V{w}}_{a} \}_a$ and the approximate LOOEs, and observe its system size dependence. The left panel of \Rfig{simulate-time} provides the plot of the actual computational time against the system size. Here, the number of examined points of $\tilde{\lambda}$ to obtain a solution path is different from size to size, and hence the plotted time is given as the whole computational time to obtain the solution path divided by the number of $\tilde{\lambda}$s points.
%%%%%%%%%%%%%%%%%%%%%
\begin{figure}[htbp]
\begin{center}
 \includegraphics[width=0.45\columnwidth]{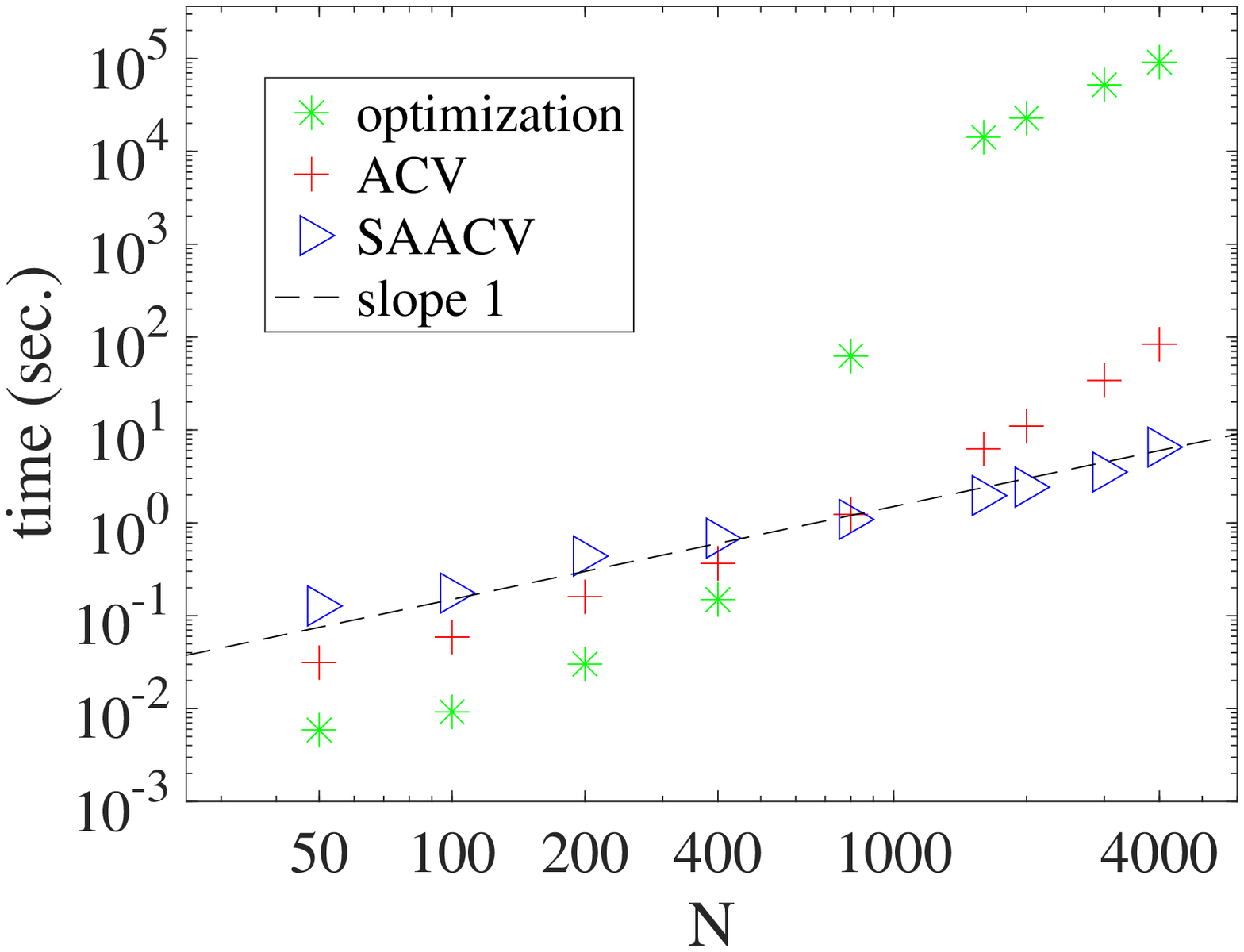}
 \includegraphics[width=0.45\columnwidth]{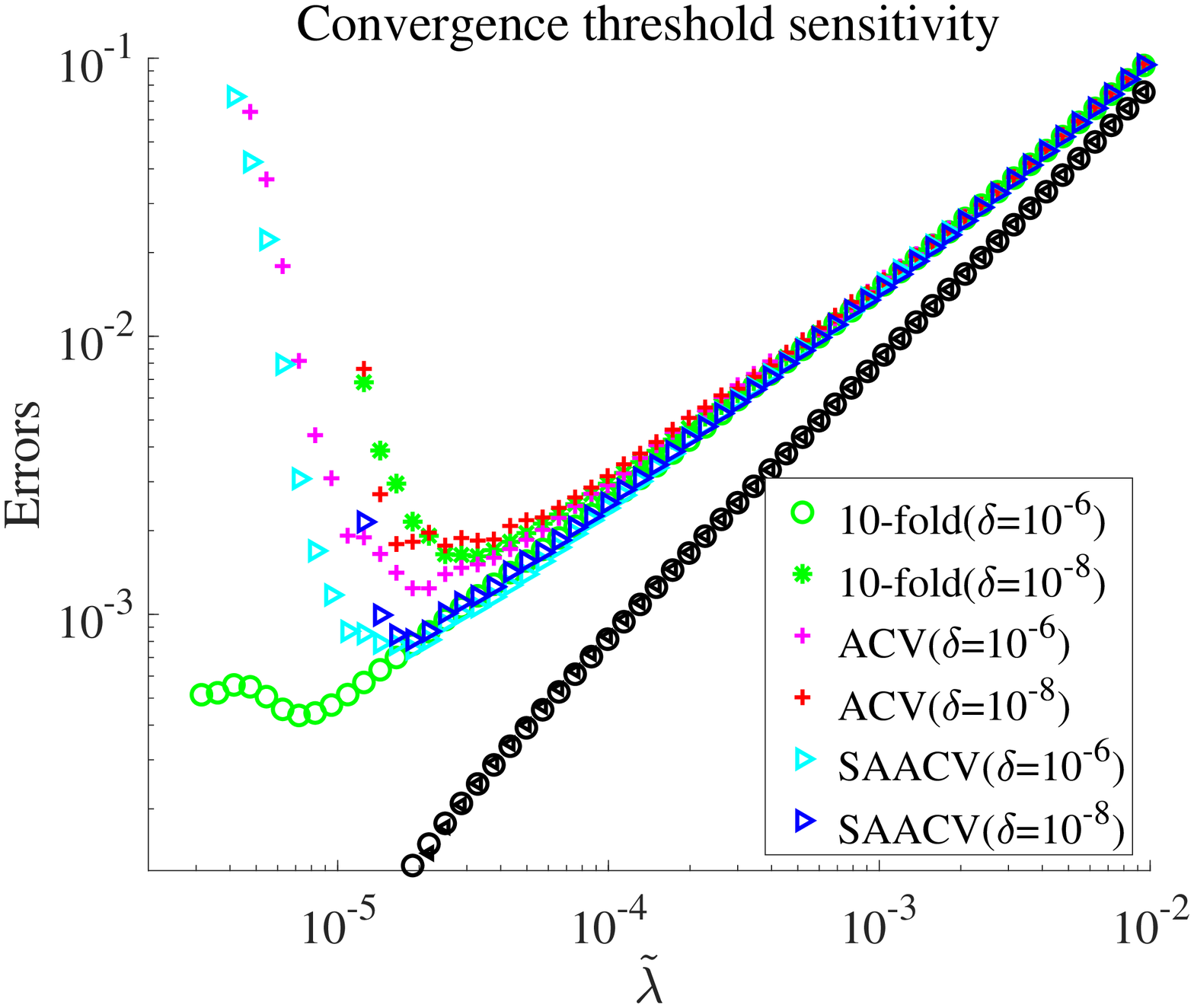}
\caption{(Left) Actual computational time spent to find the solution of \Req{w-full} and that for ACV and SAACV, plotted against the feature dimensionality $N$ in a double logarithmic scale. Note that the computational time for the $k$-fold CV is about $k$ times larger than that for finding the solution of \Req{w-full}, represented by the green asterisks. Parameters are fixed at $L=8$, $\sigma_{\xi}^2=0.01$, $\alpha=2$ and $\rho_0=0.5$. Here, $\eta=1$. (Right) The errors are obtained for the two convergence thresholds $\delta=10^{-6}$ and $\delta=10^{-8}$. Error bars are omitted for visibility. For the tighter case $\delta=10^{-8}$, the minimum value of $\tilde{\lambda}$ in the examined range is larger than that of the case $\delta=10^{-6}$, though the systematic difference with the results of the literal LOO CV is already clear. The training errors of these two different $\delta$, represented by black circles and left-pointing triangles, are completely overlapping. The system parameters are $N=400$, $L=8$, $\sigma_{\xi}^2=0.01$, $\alpha=2$ and $\rho_0=0.5$. Here, $\eta=1$.}
\Lfig{simulate-time}
\end{center}
\end{figure}
%%%%%%%%%%%%%%%%%%%%%
The left panel of \Rfig{simulate-time} clearly displays the advantage and disadvantage of the developed approximations. For small sizes, the computational time for optimization to obtain $\{ \hat{\V{w}}_a \}_a$ is shorter than the time to compute the approximate LOOEs, and hence the literal CV is better.  However, for larger systems, the optimization cost increases rapidly and for $N\gsim 400$ the approximate CV is better. For $N\gsim 800$, the ACV cost exceeds that of SAACV. The SAACV cost behaves linearly as a function of $N$ (see the black dashed line), and hence for larger systems of $N\gsim 800$ SAACV can be a very powerful tool. As a related issue, we mention the convergence problem of the algorithm. In the right panel of \Rfig{simulate-time}, we compare the errors at two different values of the convergence threshold $\delta$. An important observation is that a significant difference exists in the literal CV results while other curves do not show a strong change. This implies that our approximate formula is rather robust and can be used with a rather loose convergence threshold or conversely, we can use the systematic deviation between the literal CV and our approximations as an indicator to verify the tightness of the convergence threshold. This is beneficial, especially when treating large models, for which the convergence check is a common annoying task. 

%%%%%%%%%%%%%%%%%%%%%%%%%%%%%%%%%%%%%%%%%%%%%%%%%%%%%
%%%%%%%%%%%%%%%%%%%%%%%%%%%%%%%%%%%%%%%%%%%%%%%%%%%%%
\subsection{On real-world dataset}\Lsec{On real-world}
Next, we test the approximate formula on a real-world dataset. As shown above, our approximations become more precise if the model dimensionality and data size are large. Hence, we chose the ISOLET dataset which is a relatively large problem among classification tasks collected in the UCI machine learning repository~\citep{Lichman:13}. The feature dimensionality, the data size, and the class number are $N=617$, $M=6238$, and $L=26$, respectively. Here we apply the $10$-fold CV, instead of the LOO CV because of the computational reason, and our approximations to this dataset. The result is given in \Rfig{ISOLET}. 
%%%%%%%%%%%%%%%%%%%%%
\begin{figure}[htbp]
\begin{center}
 \includegraphics[width=0.45\columnwidth]{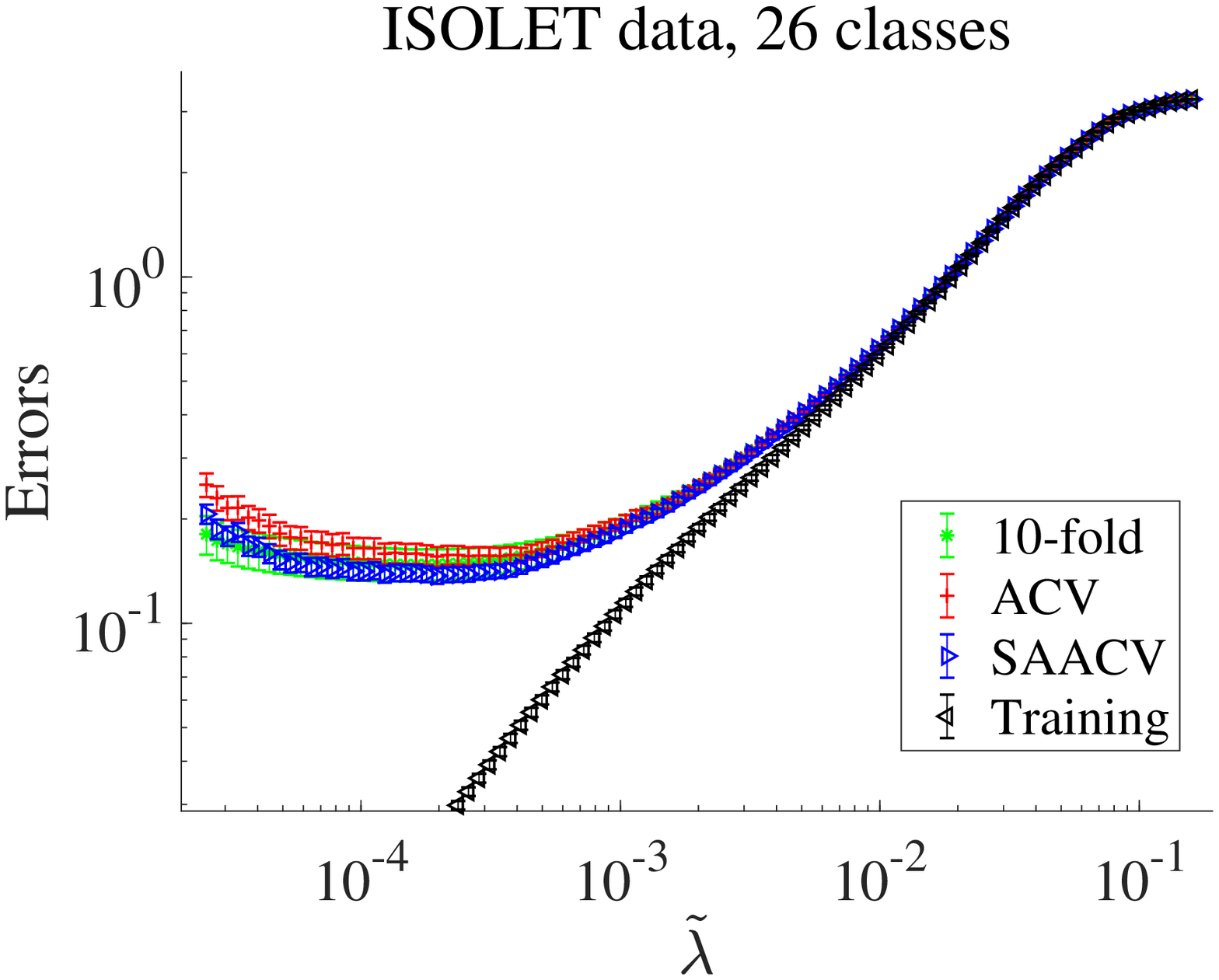}
 \includegraphics[width=0.45\columnwidth]{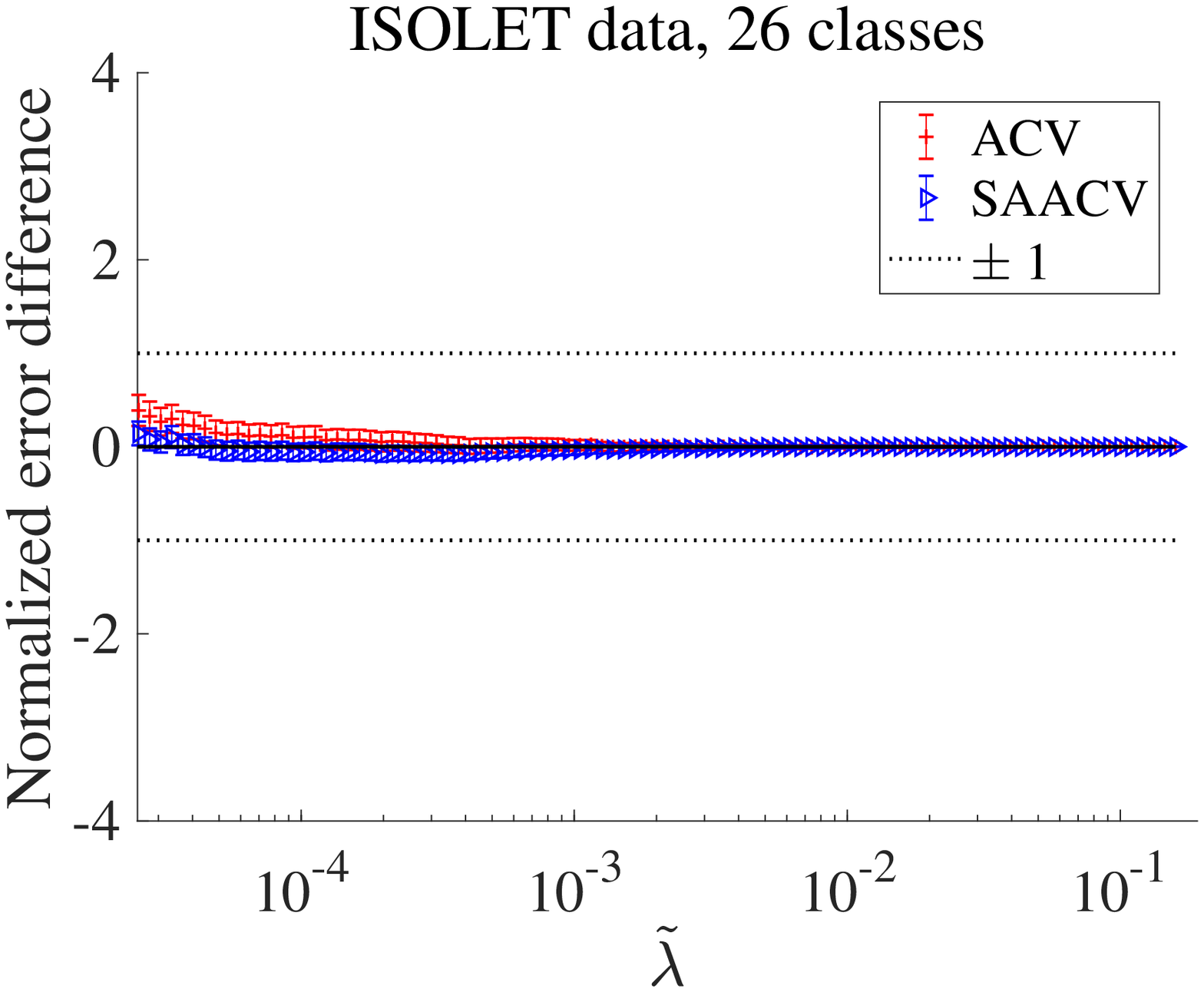}
\caption{Approximate CV performance on the ISOLET data of $L=26$ classes. The errors are shown in the left panel and the normalized error differences between the approximations and the 10-fold CV are in the right panel. At the estimated minimums of the prediction error, the accuracy rate for correctly classifying the test data is about $0.86$ while the probability of recovering the training data is about $0.98$, commonly among the literal CV and the two approximations. At the minimum value of $\tilde{\lambda}$, the leftmost point in the figure, the accuracy rates are different among the three different methods, and are $0.83, 0.78,$ and $0.81$ for the literal CV, ACV and SAACV, respectively. Here, $\eta=1$.
 }
\Lfig{ISOLET}
\end{center}
\end{figure}
%%%%%%%%%%%%%%%%%%%%%
The results of the approximations and of the 10-fold CV demonstrate a fairly good agreement, proving the actual effectiveness of the developed approximations. In an experiment, the actual computational time to obtain the result of the full simulation, of the 10-fold CV, of ACV, and of SAACV were $785$, $7825$,  $5173$, and $689$ seconds, respectively. The system parameters $\{ N,M,L\} $ are rather large in this problem and thus the advantage of ACV is not large, while the efficiency of SAACV stands out in such situations. 

%%%%%%%%%%%%%%%%%%%%%%%%%%%%%%%%%%%%%%%%%%%%%%%%%%%%%
%%%%%%%%%%%%%%%%%%%%%%%%%%%%%%%%%%%%%%%%%%%%%%%%%%%%%
\subsection{When does SAACV fail?}\Lsec{When does}
Two major factors neglected in SAACV are the correlations among feature components and the heterogeneity among feature vectors. If these factors are strong, the approximation accuracy of SAACV is expected to be degraded. In this section, we examine this point.

First, to test the impact of correlations in feature components, we add further constraints to the simulated data treated in \Rsec{On simulated} and examine the approximation performance on the situation. Two cases are treated: the first is the case where the true feature vectors  $\{ \V{w}_{0a} \}$ have common components among all the classes. The result of this case is shown in the left panels in \Rfig{SAACV_fail-correlated}. Here, the fraction of the common components to the non-zero components is $r_{\rm common}=0.9$ and thus the overlap between feature vectors of different classes is rather large. The other is the case where the noise vector has strong correlations among the components. The result of this case is presented in the right panels in \Rfig{SAACV_fail-correlated}, in which the noise strength is $\sigma_{\xi}^2=1$ and the correlation coefficient of any pair of noise components is ${\rm Corr}(\xi_i,\xi_j)=0.9$; hence the noise and the correlation are rather large. 
%%%%%%%%%%%%%%%%%%%%%
\begin{figure}[htbp]
\begin{center}
 \includegraphics[width=0.45\columnwidth]{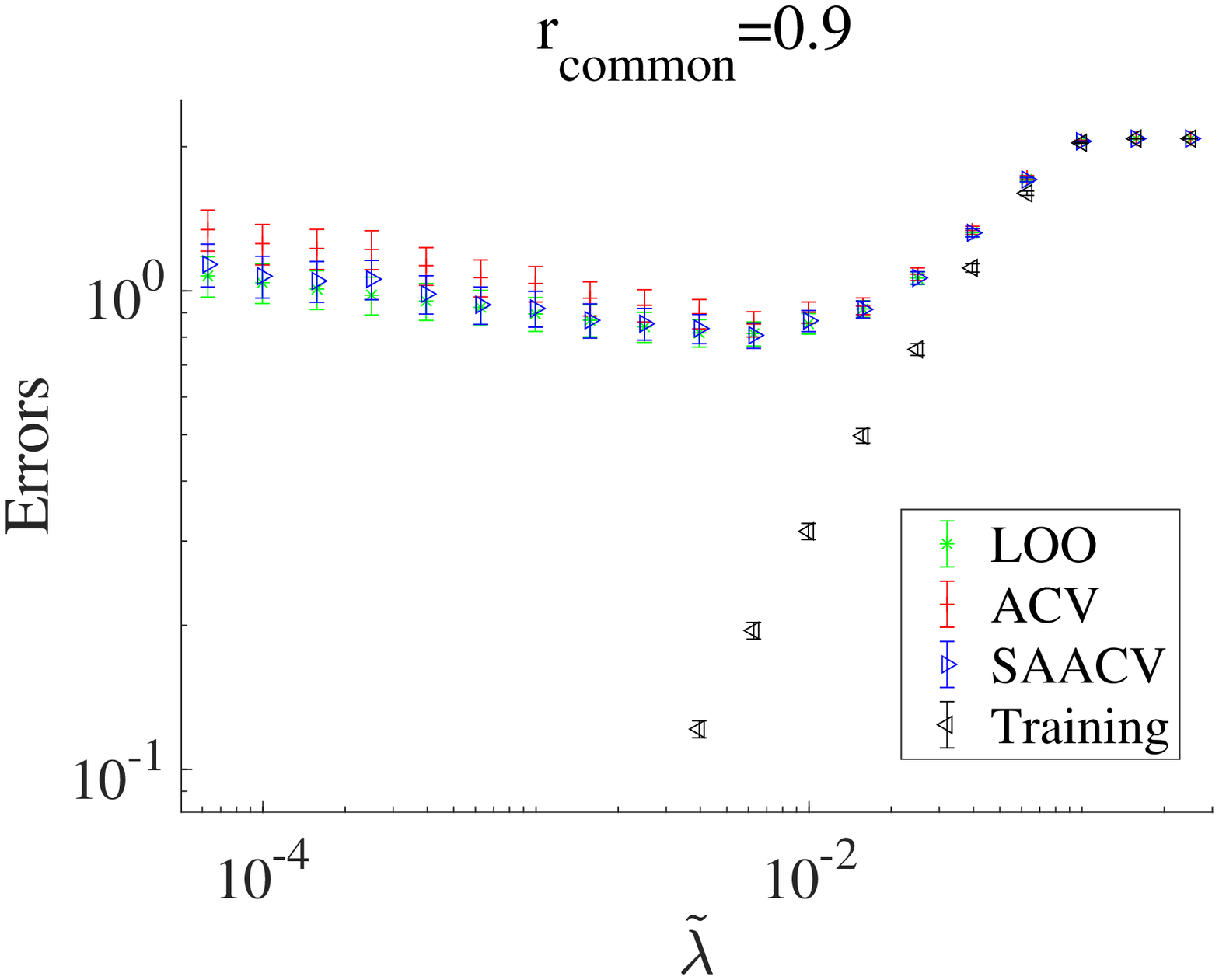}
 \includegraphics[width=0.45\columnwidth]{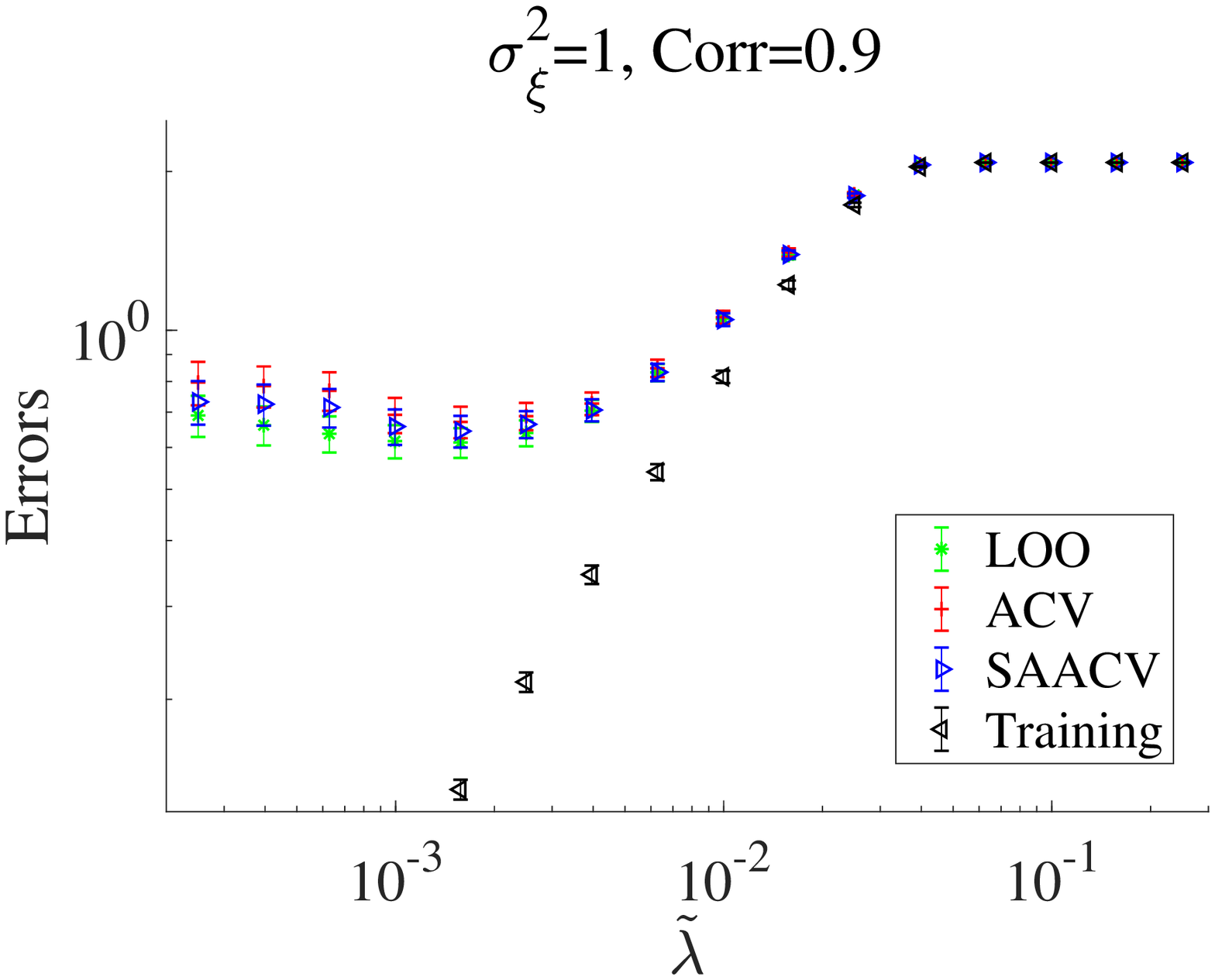}
 \includegraphics[width=0.45\columnwidth]{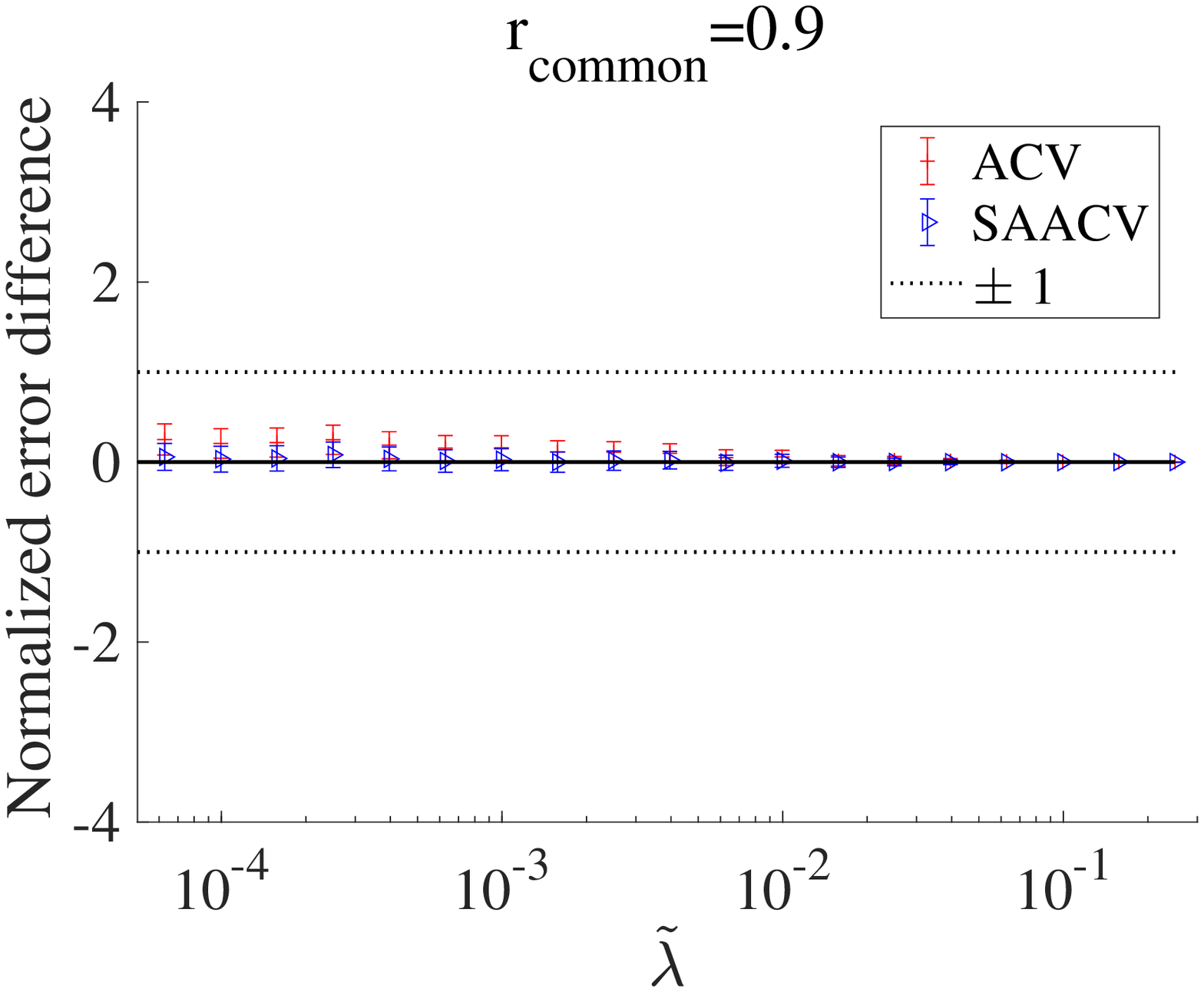}
 \includegraphics[width=0.45\columnwidth]{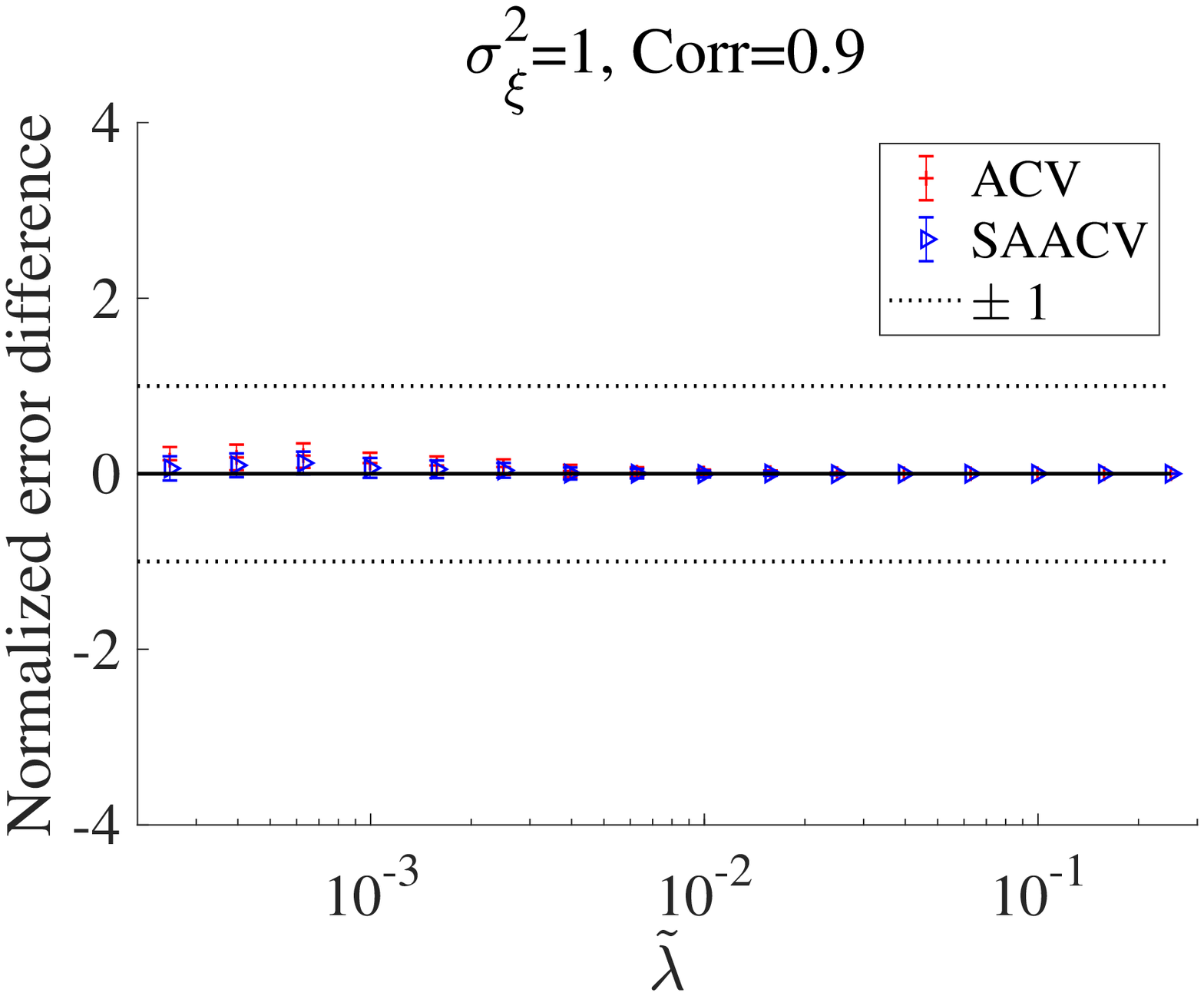}
\caption{(Upper) Log-log plots of the errors against $\tilde{\lambda}$ for correlated feature vectors. The left panel is for the case with common components in true feature vectors while the right one is of the correlated noise case. Parameters $(N,L,\alpha,\rho_0,\eta)=(200,8,2,0.5,1)$  are common in both the cases, while the noise strengths and convergence thresholds are different: $(\sigma_{\xi}^2,\delta)=(0.1,10^{-8})$ (left) and $(\sigma_{\xi}^2,\delta)=(1,10^{-9})$ (right). (Lower) The normalized error difference \NReq{NED} plotted against $\tilde{\lambda}$. The parameters of each panel are them of the corresponding upper one. }
\Lfig{SAACV_fail-correlated}
\end{center}
\end{figure}
%%%%%%%%%%%%%%%%%%%%%
For both the cases, the performance of the approximate formula is fairly good, implying that SAACV is likely to perform well even when components of the feature vectors are correlated. Similar findings were actually obtained in the case of linear models~\citep{obuchi2016cross}. This is a preferable observation because it implies that the applicable limit of SAACV can be extended to a wider class of feature vectors than that is assumed in our present derivation in which the weakness of the correlations is assumed, as seen in \Rsec{The SA approximation}. These also imply that there possibly exists another approximation formula taking into account the correlations but being similar to SAACV. A promising framework to derive such a formula might be the adaptive TAP method~\citep{opper2001adaptive,opper2001tractable,opper2005expectation}. The adaptive TAP method itself requires a larger computational cost than that of SAACV but it is possible to reduce the computational cost up to the linear scaling with respect to $N$ and $M$ by employing an additional simplifying approximation~\citep{kabashima2014signal,DBLP:journals/corr/abs-1801-05411}. This is, however, rather technical and we leave it as a future work.

Second, to examine the effect of the heterogeneity among feature vectors, we introduce an amplifying factor $\Omega$ to control the norm of feature vectors. In particular, we multiply the factor $\Omega$ to the feature vectors of some chosen classes, as $\V{x_{\mu}}\to \Omega \V{x_{\mu}}$. Here, we use the simulated data identical to that for the center panels in \Rfig{simulate-size} of the parameters $(N,L,\alpha,\rho_0,\sigma_\xi^2,\eta)=(200,8,2,0.5,0.1,1)$, except that the amplifying factor $\Omega=100$ is applied to the latter four classes $y=5,6,7,8$. The approximation performance on this dataset is shown in \Rfig{SAACV_fail-amp}.
%%%%%%%%%%%%%%%%%%%%%
\begin{figure}[htbp]
\begin{center}
 \includegraphics[width=0.45\columnwidth]{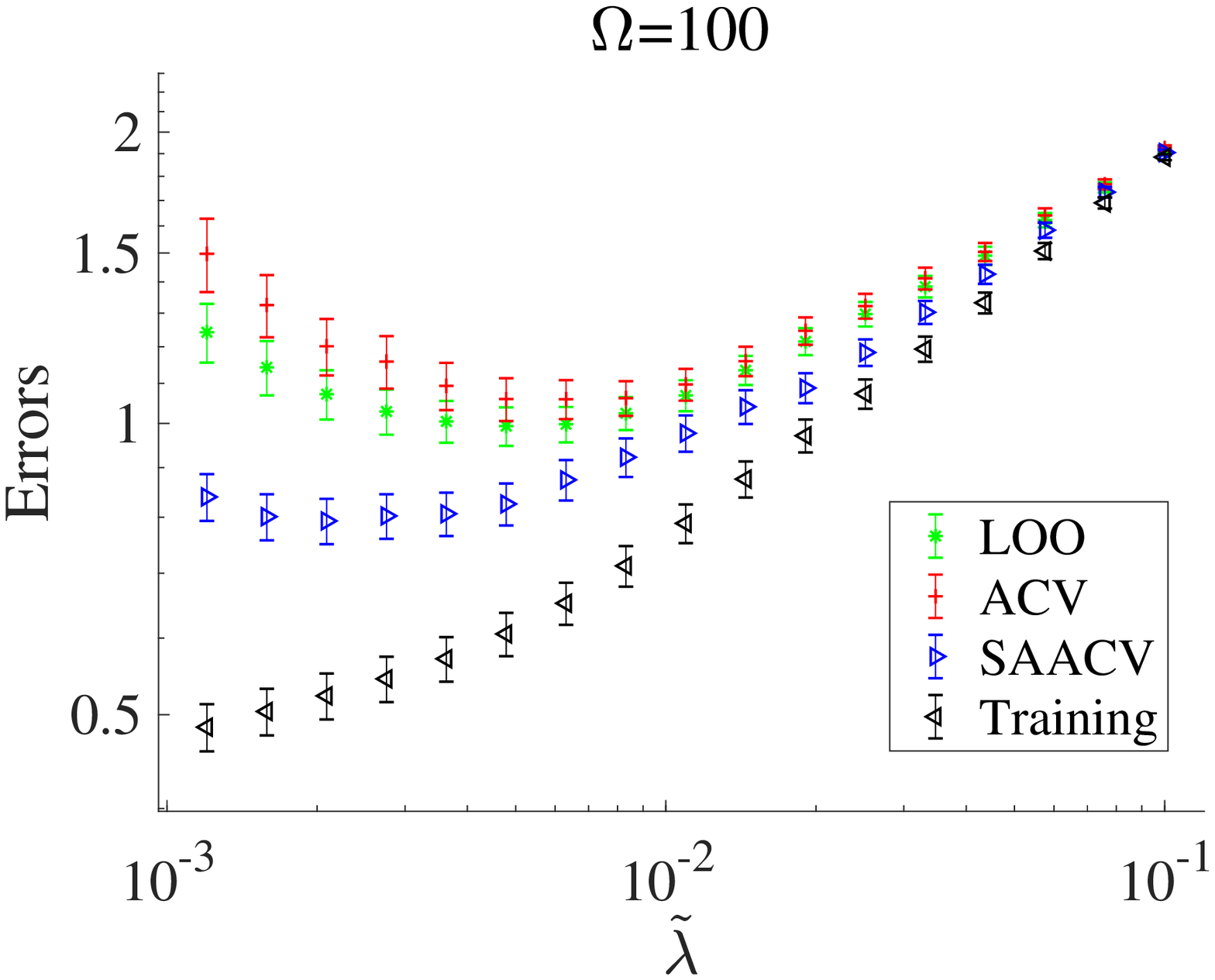}
 \includegraphics[width=0.45\columnwidth]{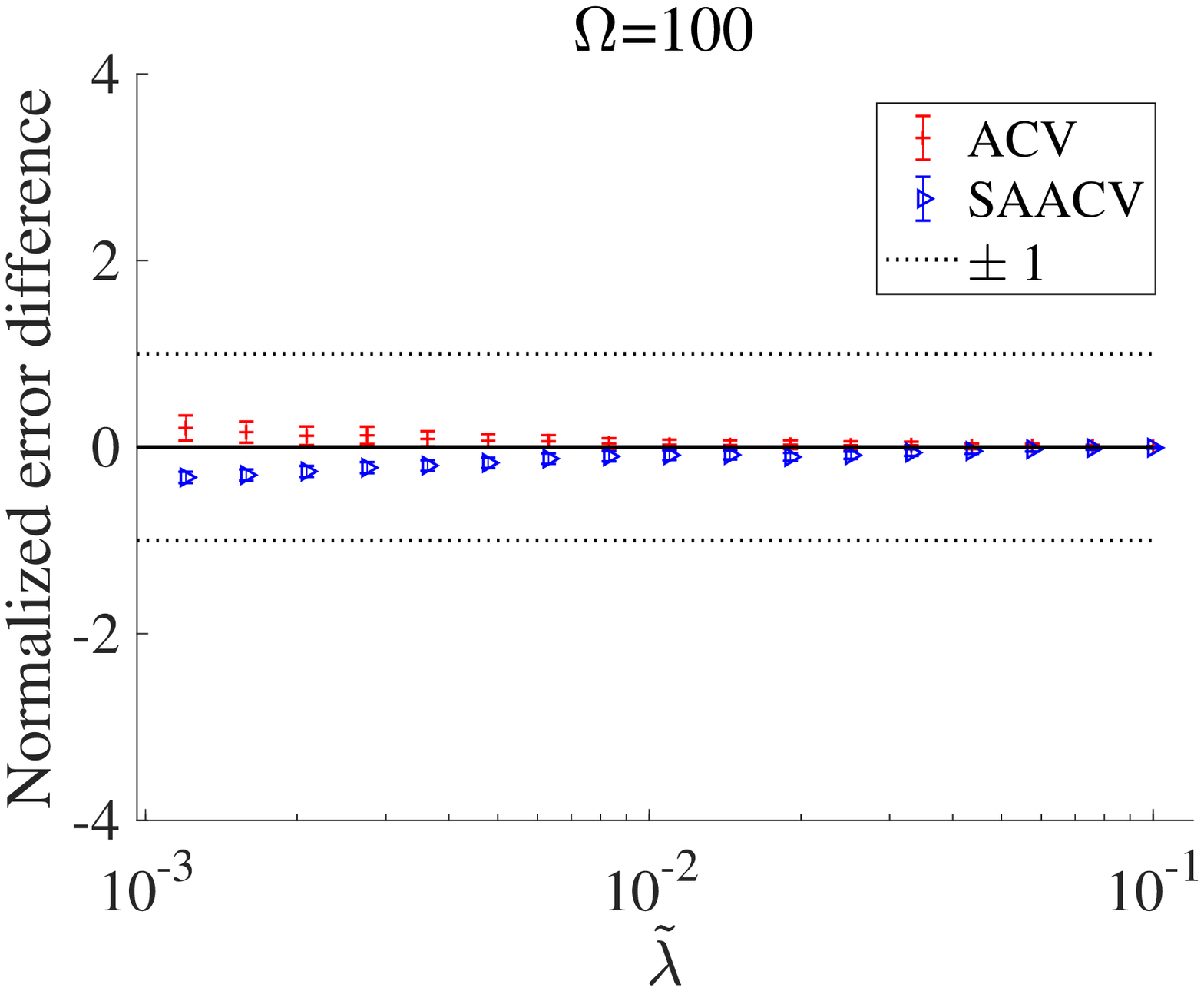}
\caption{(Left) Log-log plots of the errors against $\tilde{\lambda}$ with strong heterogeneity in feature vectors. The same dataset as that of the center panels in \Rfig{simulate-size} is used but the feature vectors for the classes $y_{\mu}=5,6,7,8$ are amplified as $\V{x_{\mu}}\to \Omega \V{x_{\mu}}$ by the factor $\Omega=100$. The ACV result is consistent with the LOO CV one while that of SAACV is not. (Right) The normalized error difference corresponding to the left panel. }
\Lfig{SAACV_fail-amp}
\end{center}
\end{figure}
%%%%%%%%%%%%%%%%%%%%%
We also examine the impact of the same heterogeneity on a real-world dataset in \Rfig{SAACV_fail-mnist}. Here, we treat the well-known MNIST data of handwritten digits~\citep{lecun1998gradient}. For simplicity, we only use the data of two digits: $0$ and $1$. As a preprocessing, feature components with small variances are removed and only the $N=350$ components of the largest variances are retained; the original size of the feature vector is $784=28\times28$ and thus almost the half of the components are discarded. Then, the usual standardization procedure is conducted. Further, we apply the amplifying factor $\Omega=10$ to the class of $1$ (right panels), while the case without the amplification (or $\Omega=1$, left panels) is also examined for comparison. 
%%%%%%%%%%%%%%%%%%%%%
\begin{figure}[htbp]
\begin{center}
 \includegraphics[width=0.45\columnwidth]{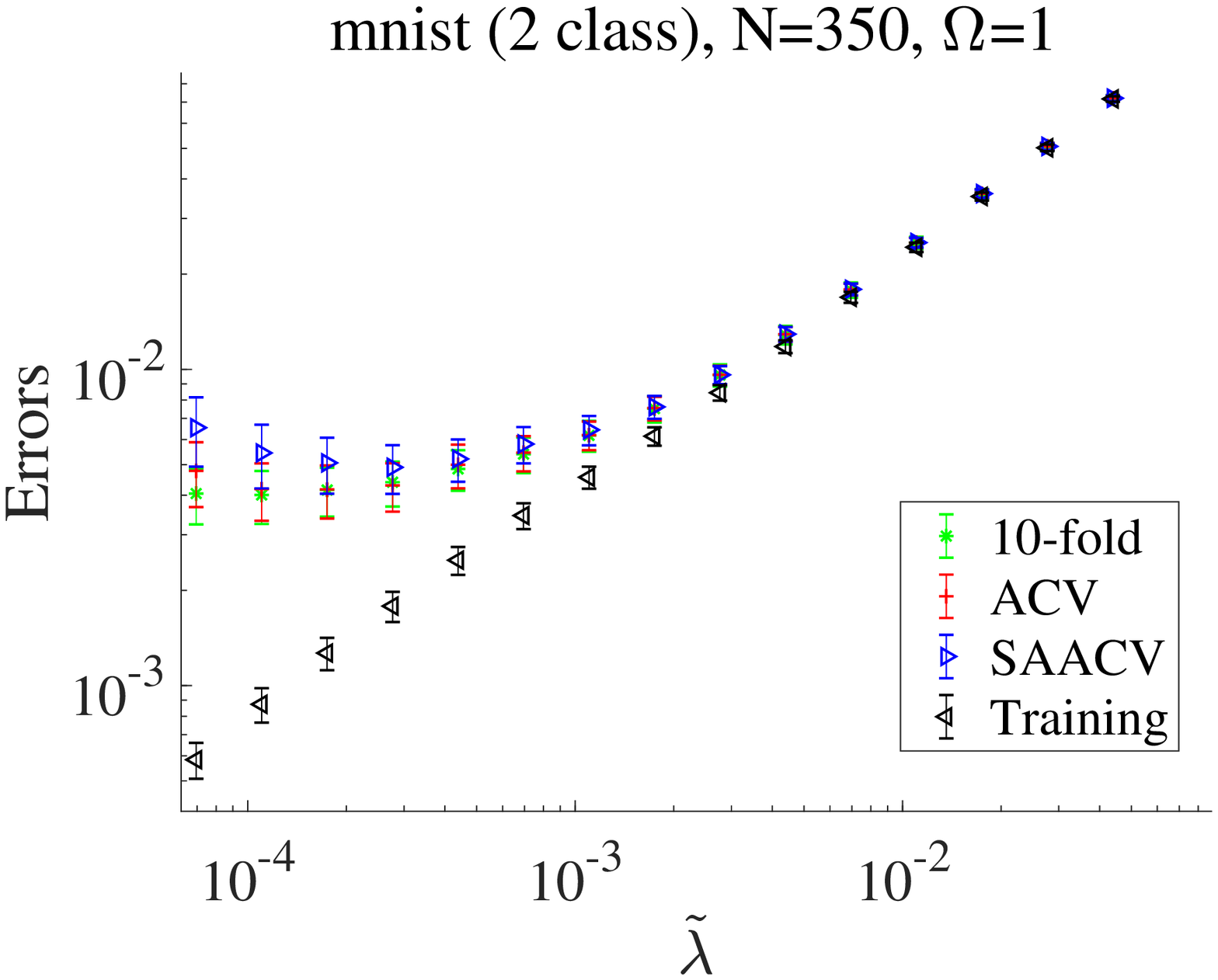}
 \includegraphics[width=0.45\columnwidth]{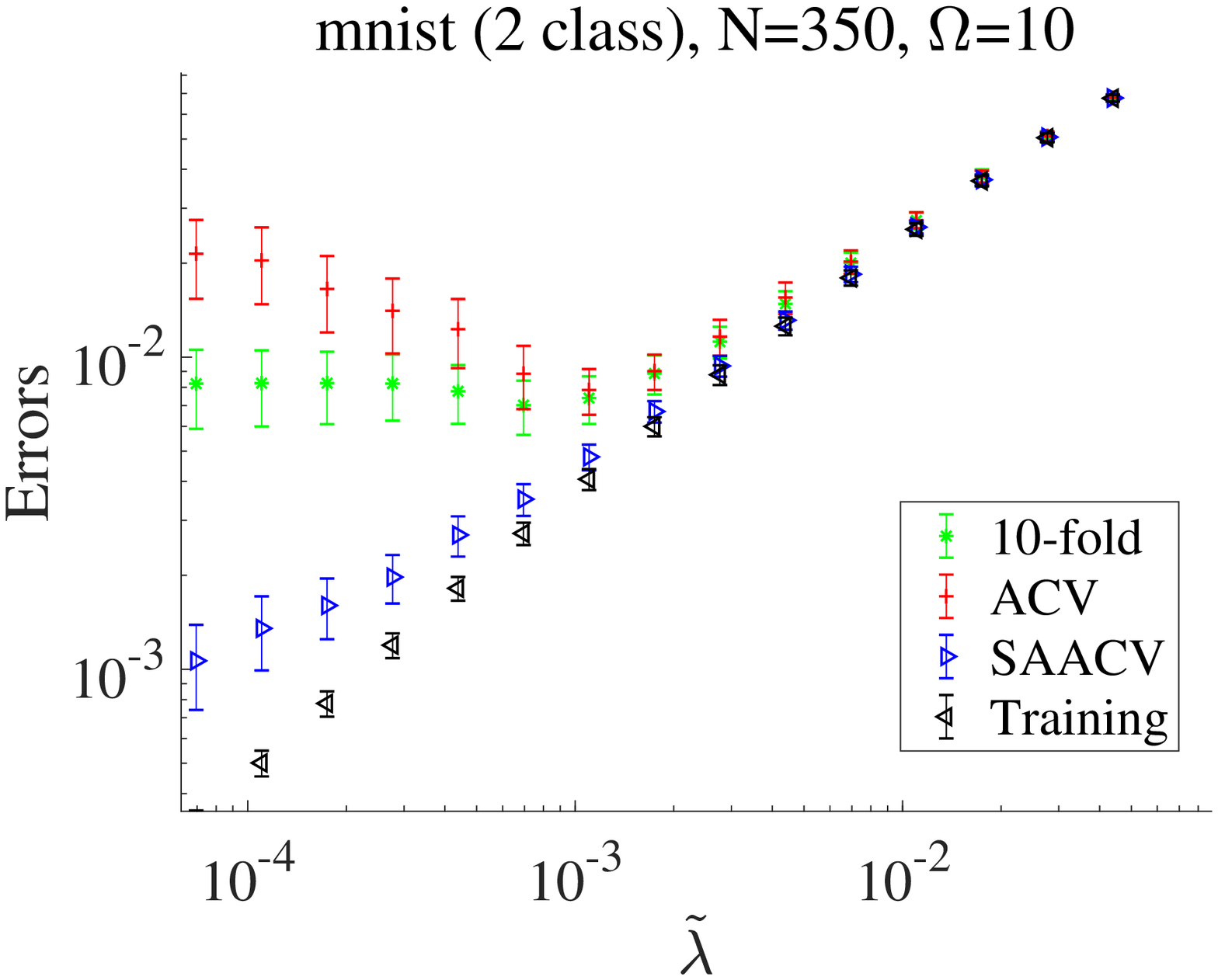}
 \includegraphics[width=0.45\columnwidth]{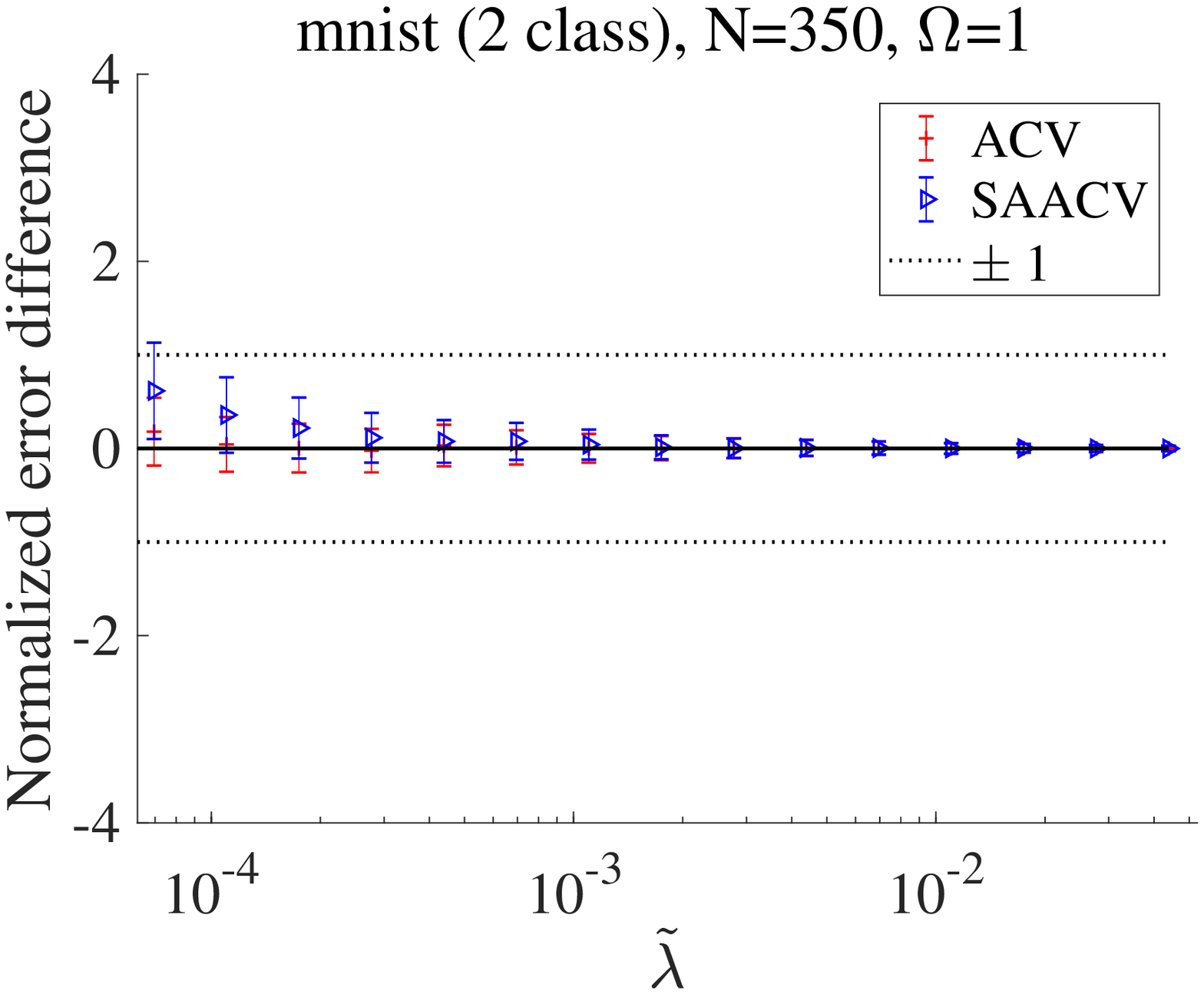}
 \includegraphics[width=0.45\columnwidth]{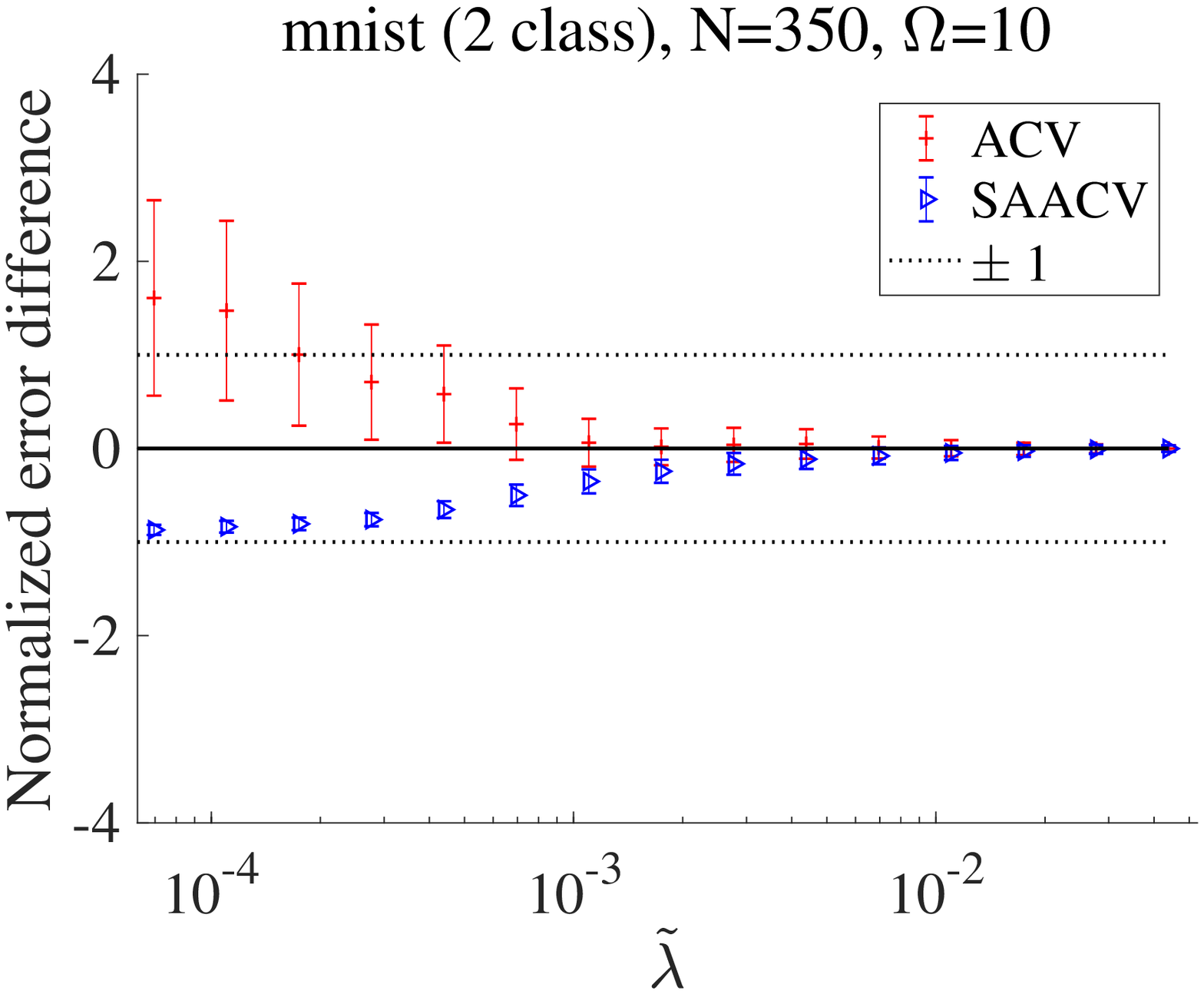}
\caption{(Upper) Log-log plots of the errors against $\tilde{\lambda}$ of mnist handwritten data with two digits $0$ and $1$. The amplifying factor is not applied (or $\Omega=1$) in the left panels while is applied ($\Omega=10$) in the right ones. The strong heterogeneity among the classes affect the performance of SAACV. (Lower) The normalized error difference corresponding to the upper panels.}
\Lfig{SAACV_fail-mnist}
\end{center}
\end{figure}
%%%%%%%%%%%%%%%%%%%%%
These two examples clearly show that ACV shows a consistency with the LOO CV behavior while SAACV does not, demonstrating that SAACV gives an inaccurate estimate of the CV error for datasets with strong heterogeneity. This kind of heterogeneity can naturally emerge in some applications: for example if we consider problems in medical statistics, a number of biological markers can give distinguishably large values for affected patients compared to unaffected ones, yielding larger values in norm for feature vectors of affected patients. This consideration suspects the efficiency of SAACV. We, however, stress that this kind of heterogeneity attributed to the belonging class can be absorbed by rescaling the weights as $\{ \V{w}_{a} \}_a \to \{ \Omega_a^{-1} \V{w}_{a} \}_a$, where $\Omega_a$ is chosen to homogenize the feature vector norm in different classes as $||\V{x}_{y_{\mu}}\Omega_a||_2\approx {\rm const}$. For the $\ell_1$ regularization case, this resultantly leads to the regularization coefficients which take different values adaptively to the belonging class as
\be
\lambda \sum_{a} ||\V{w}_a||_1 \to  \sum_{a} \lambda\Omega_a ||\V{w}_a||_1
= \sum_{a}  \lambda_a ||\V{w}_a||_1.
\ee
For this problem with adaptive regularization coefficients, our approximation formula can be applied in the completely same manner, which can be convinced by seeing \Rcodes{ACV}{SAACV} where the value of the regularization coefficient is not required as the argument. The $\ell_2$-norm can also be handled, though the codes should be extended to take into account the groupwise coefficients as arguments. We argue that this rescaling is a natural prescription to treat strong heterogeneity among different classes, and once employing this prescription the weak point of SAACV is naturally cured. 

As a noteworthy remark, we point out that the basic idea of SAACV is closely related to Wahba's generalized cross-validation (GCV) for linear regression~\citep{golub1979generalized}. In GCV, the heterogeneity in coefficient corresponding to $C_{\mu}^{\bs \mu}$ in SAACV is also neglected, and hence it shares the same weak point as SAACV, when it is regarded as an approximation of the CV estimator to generalization errors. We stress that this kind of approximation reducing the computational cost is again needed because the data size and the model dimensionality are increasing rapidly in recent years.  

%%%%%%%%%%%%%%%%%%%%%%%%%%%%%%%%%%%%%%%%%%%%%%%%%%%%%
%%%%%%%%%%%%%%%%%%%%%%%%%%%%%%%%%%%%%%%%%%%%%%%%%%%%%
%%%%%%%%%%%%%%%%%%%%%%%%%%%%%%%%%%%%%%%%%%%%%%%%%%%%%
\section{Conclusion}\Lsec{Conclusion}
In this paper, we have developed an approximate formula for the CV estimator of the predictive likelihood of the multinomial logistic regression regularized by the $\ell_1$-norm. An extension to the elastic net regularization has been also stated. We have demonstrated their advantages and disadvantages in numerical experiments using simulated and real-world datasets. Two versions of the approximation have been defined based on the developed formula. The first version, abbreviated as ACV, has a better performance, in terms of computational time, for middle size problems. It will eventually become worse than the literal $k$-fold CV with moderate $k$s as the problem size grows, because its computational time is scaled as a third-order polynomial of the feature dimensionality and data size, $N$ and $M$, though such a tendency has not been observed in the investigated range of $N$. We have also defined the second version based on ACV, the computational time of which is just scaled linearly with respect to $N$ and $M$. This second approximation is called SAACV, and it has been demonstrated that SAACV is slow for small size problems but has a great advantage for large size problems. Hence, we suggest leveraging the literal CV for small, ACV for middle, and SAACV for large size problems. 

Our derivation is based on the perturbation which assumes that there is a small difference between the full and leave-one-out solutions. This assumption will not be satisfied for some specific cases. Even with this restriction, we expect the range of application of our formula is wide enough and we would like to encourage readers to leverage it in their own work.
We have implemented MATLAB and python codes and they are available in~\citep{obuchi:acv_mlr_matlab,obuchi:acv_mlr_python}.

The perturbative approach employed here is fairly general and can be applied to a wide class of generalized linear models with convex regularizations. The development of practical formulas for these cases will be of great assistance, given that we are living in the Big Data era. 

%%%%%%%%%%%%%%%%%%%%%%%%%%%%%%%%%%%%%%%%%%%%%%%%%%%%%
%%%%%%%%%%%%%%%%%%%%%%%%%%%%%%%%%%%%%%%%%%%%%%%%%%%%%
%%%%%%%%%%%%%%%%%%%%%%%%%%%%%%%%%%%%%%%%%%%%%%%%%%%%%
%\section*{Acknowledgement}
\acks
This work was supported by JSPS KAKENHI Nos. 18K11463 (TO), 25120013 and 17H00764 (YK). TO is also supported by a Grant for Basic Science Research Projects from the Sumitomo Foundation. The authors are grateful to Takashi Takahashi for implementation of the approximation formula in python.

\appendix
%%%%%%%%%%%%%%%%%%%%%%%%%%%%%%%%%%%%%%%%%%%%%%%%%%%%%
%%%%%%%%%%%%%%%%%%%%%%%%%%%%%%%%%%%%%%%%%%%%%%%%%%%%%
%%%%%%%%%%%%%%%%%%%%%%%%%%%%%%%%%%%%%%%%%%%%%%%%%%%%%
\section{The SA approximation}\Lsec{The SA approximation}
Let us derive \Req{chi-SA} in the SA approximation. We work on a framework called a cavity method in statistical physics or belief propagation (BP) in computer science. We start from defining the so-called Boltzmann distribution:
\be
&&
P\lb \lbb \V{w}_{a} \rbb_{a=1}^{L} \Bigr| D^M,\lambda \rb=\frac{1}{Z\lb D^M,\lambda \rb}e^{-\beta \mathcal{H}\lb \lbb \V{w}_{a} \rbb_{a=1}^{L} \Bigr| D^M,\lambda \rb}
\no \\ &&
=\frac{ e^{-\beta \sum_{a} \lb \lambda_1||\V{w}_a ||_1+\frac{\lambda_2}{2}||\V{w}_a ||_2^2\rb} }{Z\lb D^M,\lambda \rb} 
\prod_{\mu=1}^{M}\phi^{\beta}\lb y_{\mu}|\{u_{\mu a}\}_a \rb.
\Leq{Boltzmann}
\ee
In the $\beta \to \infty$ limit, this distribution converges to a point-wise measure of the solution of \Req{w-full} and hence, it is useful for analyzing \Req{Boltzmann}. We note that the BP is usually applied to graphical models having sparse tree-like structures, but is also applicable to ones with densely connected structures. In such applications, the BP can be regarded as a systematic implementation of the Thouless-Anderson-Palmer (TAP) approach~\citep{thouless1977solution} in statistical physics, which can yield a set of self-consistent equations of the first and second moments of variables in statistical models when the models are of densely connected types. This approach has been applied to many different models in machine learning, which continuously yields evidences of its effectiveness~\citep{opper1996mean,opper1997mean,opper2000gaussian1,opper2000gaussian2}. When applied to densely connected models, certain correlations between variables have to be neglected to make the computation tractable; for that reason this approach, or the associated algorithm derived from it, is recently called approximate message passing (AMP)~\citep{kabashima2003cdma,donoho2009message}. Basically, the AMP assumes that ``interactions'' between the variables are weak: in the present problem this implies $(1/M)\sum_{\mu=1}^{M}x_{\mu_i}x_{\mu_j}
-\lb (1/M)\sum_{\mu=1}^{M}x_{\mu_i}\rb \lb (1/M)\sum_{\mu=1}^{M}x_{\mu_j}\rb
\approx 0~(i \neq j)$. This treatment can be justified if each feature vector $\V{x}_{\mu}$ is i.i.d.. Rigorous proofs of this fact are available for linear models and their some variants~\citep{bayati2011dynamics,barbier2017mutual}. We implicitly assume this in the following derivation.

In this appendix, we introduce a new vector representation summarizing class variables: $\V{w}_i=(w_{ai})_a$. Note that this is different from the notation used in the main body of this paper, $\V{w}_a=(w_{ai})_i$ in which the feature components are summarized. 

By regarding $\V{w}_i$  as a single variable node, the BP decomposes \Req{Boltzmann} into two types of messages as follows:
\be
&&
\tilde{M}_{\mu \to i}(\V{w}_{i})=\int \prod_{j(\neq i)}d\V{w}_{j}
~\phi^{\beta}(\V{u}_{\mu})\prod_{j(\neq i)} M_{ j \to \mu }(\V{w}_j), 
\Leq{BP1}
\\ &&
M_{i \to \mu}(\V{w}_{i})=e^{-\beta \lb \lambda_1||\V{w}_i ||_1+\frac{\lambda_2}{2}||\V{w}_i ||_2^2\rb}\prod_{\nu (\neq \mu)}\tilde{M}_{ \nu \to i }(\V{w}_i), 
\Leq{BP2}
\ee
where $\V{u}_{\mu}=(u_{\mu a})_a$. A crucial observation to assess \Reqs{BP1}{BP2} is that the argument of the potential function $\phi(\V{u}_\mu)$ has a sum of an extensive number of random variables; the central limit theorem thus justifies treating it as a Gaussian variable with the appropriate mean and variance. Hence, according to \Req{BP1} where $\V{w}_{i}$ is special, we can divide the extensive sum as follows: 
\be
u_{\mu a}=\sum_{j}x_{\mu j}w_{a j}
=
x_{\mu i}w_{a i}+\sum_{j(\neq i)}x_{\mu j}w_{aj}
\approx
x_{\mu i}w_{a i}
+\sum_{j(\neq i)}x_{\mu j}\Ave{w_{aj}}^{\bs \mu}
+t_a,
\Leq{central}
\ee
where the second term on the right-hand side represents the mean of $\sum_{j(\neq i)}x_{\mu j}w_{aj}$, the symbol $\Ave{\cdot}^{\bs \mu}$ denotes the average over the Boltzmann distribution without the $\mu$th potential function, and $\V{t}^{\mu}=(t^{\mu}_a)_a$ denotes the zero-mean Gaussian variables whose covariance is set to be that of $\lb \sum_{j(\neq i)}x_{\mu j}w_{aj} \rb_a$. This expression allows us to replace the integration $\int \prod_{j(\neq i)}d\V{w}_{j}$ by that over $\V{t}^{\mu}$ in \Req{BP1}. This significantly simplifies the computation and yields: 
\be
&&
\tilde{M}_{\mu \to i}(\V{w}_{i})
\approx
%\sqrt{\frac{\beta^{L}\det{(C_{\mu}^{\bs \mu})^{-1}}}{(2\pi)^{L} }}
\int d\V{t}
~e^{\beta \lb -\frac{1}{2}\V{t}^{\top} (C_{\mu}^{\bs \mu})^{-1} \V{t} -\NLL{\mu}(\V{w}_i,\V{t})\rb}
\equiv
%\sqrt{\frac{\beta^{L}\det{(C_{\mu}^{\bs \mu})^{-1}}}{(2\pi)^{L} }}
\int d\V{t}~e^{\beta f^{\mu}(\V{w}_i,\V{t})}
\Leq{Mhat}
\ee
where $\NLL{\mu}(\V{w}_{i},\V{t})$ is the negative log-likelihood whose argument $u_{\mu a}$ is approximated by \Req{central} and $C_{\mu}^{\bs \mu}$ is the rescaled covariance of $\sum_{j(\neq i)}x_{\mu j}w_{aj}$ defined as
\be
&&
\chi^{\bs \mu}_{(ai)(bj)}
\equiv 
\beta
\lb 
\Ave{w_{ai} w_{bj}}^{\bs \mu}-\Ave{w_{ai}}^{\bs \mu}\Ave{w_{bj}}^{\bs \mu}
\rb,
\no \\ &&
\lb C_{\mu}^{\bs \mu} \rb_{ab}
\equiv 
\sum_{i,j}x_{\mu i}x_{\mu j}\chi^{\bs \mu}_{(ai)(bj)}.
\Leq{Cdef}
\ee
In the second equation we added the contribution from $i$ for simplicity. It does not affect the following result because the $i$th term contribution is small enough. Let us focus on the limit $\beta \to \infty$. This limit allows us to use the saddle-point method, or Laplace's method, with respect to $\V{t}^{\mu}$. The associated saddle-point equation is:  
\be
\hat{\V{t}}^{\mu}=-C_{\mu}^{\bs \mu} \V{b}^{\mu}(\V{w}_i,\hat{\V{t}}^{\mu}),
\ee
where $\V{b}^{\mu}(\V{w}_i,\hat{\V{t}}^{\mu})$ is the gradient of $q_{\mu}$ defined at \Req{b^mu} but the argument $u_{\mu a}$ is approximated by \Req{central}. Now, let us expand the exponent $f^{\mu}(\V{w}_i,\V{t})$ in \Req{Mhat} with respect to the dynamical variables $\V{w}_i$ up to the second order. Putting $z_a=\sum_{i}x_{\mu i}w_{a i}$, we can define the derivatives as:
\be
&&
\Part{\hat{\V{t}}^{\mu}}{z_a}{}=-(I_{L}+C_{\mu}^{\bs \mu}F^{\mu})^{-1}C_{\mu}^{\bs \mu}F^{\mu}_{\WC a},~
\\ &&
\Part{f^{\mu}(\V{w}_i,\hat{\V{t}}^{\mu})}{z_a}{}=-b^{\mu}_{a}(\V{w}_i,\hat{\V{t}}^{\mu}),~
\\ &&
\frac{\partial^2 f^{\mu}(\V{w}_i,\hat{\V{t}}^{\mu})}{\partial z_a\partial z_b}
=-F^{\mu}_{ab}-\lb \Part{\hat{\V{t}}^{\mu}}{z_a}{} \rb^{\top} F^{\mu}_{\WC b}
=-\lb \lb I_{L}+F^{\mu}C_{\mu}^{\bs \mu}\rb^{-1}F^{\mu} \rb_{ab}.
\ee
Hence,
\be
\tilde{M}_{\mu \to i}(\V{w}_i)
\propto
e^{\beta \lb \lb \V{h}^{\mu}_i\rb^{\top} \V{w}_i-\frac{1}{2}\V{w}_i^{\top}\Gamma^{\mu}_i\V{w}_i \rb}
\ee
where
\be
&&
\V{h}^{\mu}_i=-x_{\mu i}\V{b}^{\mu},
\no \\ &&
\Gamma^{\mu}_i=
x_{\mu i}^2
(I_{L}+F^{\mu}C_{\mu}^{\bs \mu})^{-1}F^{\mu}
.
\Leq{Gamma}
\ee
Note that this second order expansion is justified in the limit $\beta \to \infty$. 

Collecting all the messages except for $\mu$, we can construct the LOO marginal distribution of $\V{w}_i$ as:
\be
&&
P^{\bs \mu}(\V{w}_{i})\propto 
e^{-\beta \lb \lambda_1||\V{w}_i ||_1+\frac{\lambda_2}{2}||\V{w}_i ||_2^2\rb}
\prod_{\nu(\neq \mu)} \tilde{M}_{\nu \to i}(\V{w}_{i})
\no \\ &&
\propto 
e^{
\beta 
\lb
(\sum_{\nu(\neq \mu)}\V{h}^{\mu}_i)^{\top} \V{w}_i
-\frac{1}{2}\V{w}_i^{\top}\lb \lambda_2 I_{L}+ \sum_{\nu(\neq \mu)}\Gamma^{\mu}_i \rb \V{w}_i 
-  \lambda_1 ||\V{w}_{i}||_1 
\rb
}.
\Leq{LOO-marginal}
\ee
Now, we can close the equation for the rescaled variance $\lb \chi^{\bs \mu}_{i} \rb_{ab}\equiv \chi^{\bs \mu}_{(ai),(bi)}$, because we can compute the variance of $\V{w}_{i}$ from \Req{LOO-marginal}. By considering the scaling, we can recognize that the variances vanish in the speed of $O(\beta^{-2})$ if one of the two components or both are inactive. The active-active components of the variance are scaled by $O(\beta^{-1})$ and remain in the rescaled variance. Focusing on the limit $\beta \to \infty$, we thus obtain:  
\be
\lb \chi^{\bs \mu}_{i} \rb_{\hat{A}_i\hat{A}_i}
= \lb \lambda_2 I_{|\hat{A}_i|}+\lb\sum_{\nu(\neq \mu)} \Gamma^{\nu}_i\rb_{\hat{A}_i\hat{A}_i} \rb^{-1}
\approx \lb \lambda_2 I_{|\hat{A}_i|}+\lb\sum_{\nu} \Gamma^{\nu}_i\rb_{\hat{A}_i\hat{A}_i} \rb^{-1}.
\Leq{selfconsistent-SA}
\ee
At the last step, the $\mu$th term is added since its contribution is expected to be small enough in the summation. This manifests that the $\mu$-dependence of $\chi^{\bs \mu}$ can be neglected and we rewrite it as $\chi^{\bs \mu}=\chi$ hereafter. By considering the meaning of the Hessian, it is easy to understand that $G^{\bs \mu}$ is identified with $\lb \lambda_2 I_{L}+\sum_{\nu(\neq \mu)} \Gamma^{\nu}_i\rb$. This yields \Req{SAkey}. 

By assuming the vanishing correlation between $\V{w}_i$ and $\V{w}_j$ for $i\neq j$, we can write
\be
\chi_{(ai)(bj)}\approx 
\delta_{ij}\left\{
\begin{array}{cc}
\lb \chi_i\rb_{ab}  &  (a,b\in \hat{A}_i ) \\ 
0  &  ({\rm otherwise})
\end{array}
\right..
\Leq{SAassumption}
\ee 
These leads to: 
\be
\lb C_{\mu}^{\bs \mu} \rb_{ab}=\sum_{ij}x_{\mu i}x_{\mu j}\chi^{\bs \mu}_{(ai)(bj)}
\approx 
\sum_{i}x_{\mu i}^2 \lb \chi_i \rb_{ab}
\approx 
\sigma_{x}^2\sum_{i}\lb \chi_i \rb_{ab}\equiv \lb C_{\rm SA} \rb_{ab}
\Leq{Capp}
\ee
The $\mu$-dependence through $x^2_{\mu i}$ is neglected at the last step, because the sum $\sum_{i}$ would mask such a weak $\mu$-dependence as long as strong heterogeneity in $\{x_{\mu i}\}_{\mu}$ is absent. Similarly, we may write the sum inside the parentheses of the righthand side of \Req{selfconsistent-SA} as:
\be
\sum_{\nu} \Gamma^{\nu}_i \approx  \sigma_{x}^2 \sum_{\nu} (I_{L}+F^{\nu}C_{\rm SA})^{-1}F^{\nu}.
\Leq{Gamma-app}
\ee
Inserting \Reqss{SAassumption}{Gamma-app} into \Req{selfconsistent-SA}, we obtain \Req{chi-SA}. 

Careful readers may be concerned about the neglected $\mu$-dependence of $\chi^{\bs \mu}$, as well as that of $G^{\bs \mu}$. If this can be neglected, may we replace $G^{\bs \mu}$ with $G$ from the beginning at \Req{u-LOO-pre}? The answer is of course no. The reason is that the difference between $G^{\bs \mu}$ and $G$ is not negligible if they are ``projected'' onto $X^{\mu}$ as in \Req{u-LOO-pre}. If they are projected onto other directions perpendicular to $X^{\mu}$, the difference is actually tiny and can be neglected, but for computing the factor $C_{\mu}^{\bs \mu}$ we need to take into account this difference appropriately. This results in the additional factor $(I-F_{\mu}C_{\mu})^{-1}$ in \Req{u-LOO}. In the SA approximation, the factor $C$ is computed based on neglecting the difference between $G$ and $G^{\bs \mu}$. As a result we cannot discriminate the two factors $C_{\mu}$ and $C_{\mu}^{\bs \mu}$. This consideration implies that our SA estimation of $C$, $C_{\rm SA}$, should be applied to $C_{\mu}^{\bs \mu}$ in \Req{u-LOO-pre} and should NOT be applied to $C_{\mu}$ in \Req{u-LOO}, because the latter formula formally takes into account the difference in advance.

%%%%%%%%%%%%%%%%%%%%%%%%%%%%%%%%%%%%%%%%%%%%%%%%%%%%%
%%%%%%%%%%%%%%%%%%%%%%%%%%%%%%%%%%%%%%%%%%%%%%%%%%%%%
%%%%%%%%%%%%%%%%%%%%%%%%%%%%%%%%%%%%%%%%%%%%%%%%%%%%%
\bibliography{obuchi_2nd}

\end{document}